\documentclass[a4paper,10pt]{article}
\usepackage[margin=1in]{geometry}

\usepackage{amsthm}
\usepackage[utf8]{inputenc}
\usepackage[T1]{fontenc}   
\usepackage{textgreek}
\usepackage{comment}
\usepackage{hyperref}       
\usepackage{url}            
\usepackage{nicefrac}       
\usepackage{microtype}     
\usepackage{fullpage}

\usepackage[dvipsnames]{xcolor}
\usepackage{framed}
\colorlet{shadecolor}{pink} 
\usepackage{authblk}
\usepackage{adjustbox}
\usepackage{comment}
\usepackage{multirow}
\usepackage{natbib}
\usepackage{makecell}   
	
\usepackage{color, colortbl}
\usepackage{graphicx}
\usepackage{soul}
\usepackage{threeparttable}
\usepackage{booktabs}
\usepackage{tablefootnote}

\usepackage{amsmath,amsthm,amssymb,amsfonts}
\usepackage{enumerate}
\usepackage{cleveref}
\usepackage{bm}
\usepackage{pifont}

\theoremstyle{plain} 
\newcommand{\vertiii}[1]{{\left\vert\kern-0.25ex\left\vert\kern-0.25ex\left\vert #1 
\right\vert\kern-0.25ex\right\vert\kern-0.25ex\right\vert}}
  
\definecolor{burgundy}{rgb}{0.5, 0.0, 0.13}

\usepackage{hyperref}

\usepackage{algorithm}
\usepackage{algpseudocode}

\hypersetup{colorlinks=true,citecolor=burgundy,linkcolor=blue,filecolor=magenta,urlcolor=cyan,}

\definecolor{antiquewhite}{rgb}{0.98, 0.92, 0.84} 
\definecolor{blizzardblue}{rgb}{0.67, 0.9, 0.93}

\usepackage{amsmath,amsfonts,bm}

\def\eqref#1{equation~\ref{#1}}

\def\1{\bm{1}}

\DeclareMathAlphabet{\mathsfit}{\encodingdefault}{\sfdefault}{m}{sl}
\SetMathAlphabet{\mathsfit}{bold}{\encodingdefault}{\sfdefault}{bx}{n}

\newcommand{\E}{\mathbb{E}}

\newcommand{\KL}{D_{\mathrm{KL}}}
\newcommand{\Var}{\mathrm{Var}}

\DeclareMathOperator*{\argmax}{arg\,max}

\usepackage[colorinlistoftodos,bordercolor=orange,backgroundcolor=orange!20,linecolor=orange,textsize=scriptsize]{todonotes}

\newtheorem{theorem}{Theorem}
\newtheorem{definition}{Definition}
\newtheorem{lemma}{Lemma}
\newtheorem{proposition}{Proposition}

\newtheorem{remark}{Remark}
\newcommand{\SAMerging}{\texttt{SAMerging}\xspace}
\usepackage{xspace}

\title{\bf Model Merging via Multi-Teacher Knowledge Distillation}
\date{}

\author{Seyed Arshan Dalili \qquad Mehrdad Mahdavi \vspace*{.2em} \\ 
 \quad  The Pennsylvania State University \vspace*{.2em} \\  \{\texttt{sbd5760,mzm616\}@psu.edu}}
\begin{document}

\maketitle

\begin{abstract}
Model merging has emerged as a lightweight alternative to joint multi-task learning (MTL), yet the generalization properties of merged models remain largely unexplored. Establishing such theoretical guarantees is non-trivial, as the merging process typically forbids access to the original training data and involves combining fine-tuned models trained on fundamentally heterogeneous data distributions. Without a principled understanding of these dynamics, current methods often rely on heuristics to approximate the optimal combination of parameters. This dependence is most critical in coefficient scaling, the weighting factors that modulate the magnitude of each fine-tuned model's contribution to the shared parameter. However, without a principled objective to guide their selection, these methods lead to brittle performance and are highly sensitive to scaling initialization. We address this gap by (i) establishing a novel flatness-aware PAC-Bayes generalization bound specifically for the model merging setting. This analysis introduces a ``cross-task heterogeneity'' term that formally captures the mismatch between diverse fine-tuned model priors and the target multi-task distributions. Guided by this theoretical insight, (ii) we frame model merging as multi-teacher knowledge distillation on scarce, unlabeled data. We formally demonstrate that minimizing the student-teacher Kullback-Leibler divergence directly tightens the upper bound on the merged model's excess risk. Guided by the flatness-aware bound derived, (iii) we operationalize this objective via \SAMerging, a method that employs Sharpness-Aware Minimization (SAM) to find flat minima. Empirically, SAMerging establishes a new state of the art across vision and NLP benchmarks, achieving remarkable performance with orders-of-magnitude greater data efficiency, requiring as few as 16 examples per task, while incurring no additional inference or memory overhead. The code is available at \href{https://github.com/arshandalili/SAMerging}{https://github.com/arshandalili/SAMerging}.
\end{abstract}

\section{Introduction}
\label{sec:introduction}

\begin{figure}[t]
\centering
\includegraphics[width=0.5\linewidth]{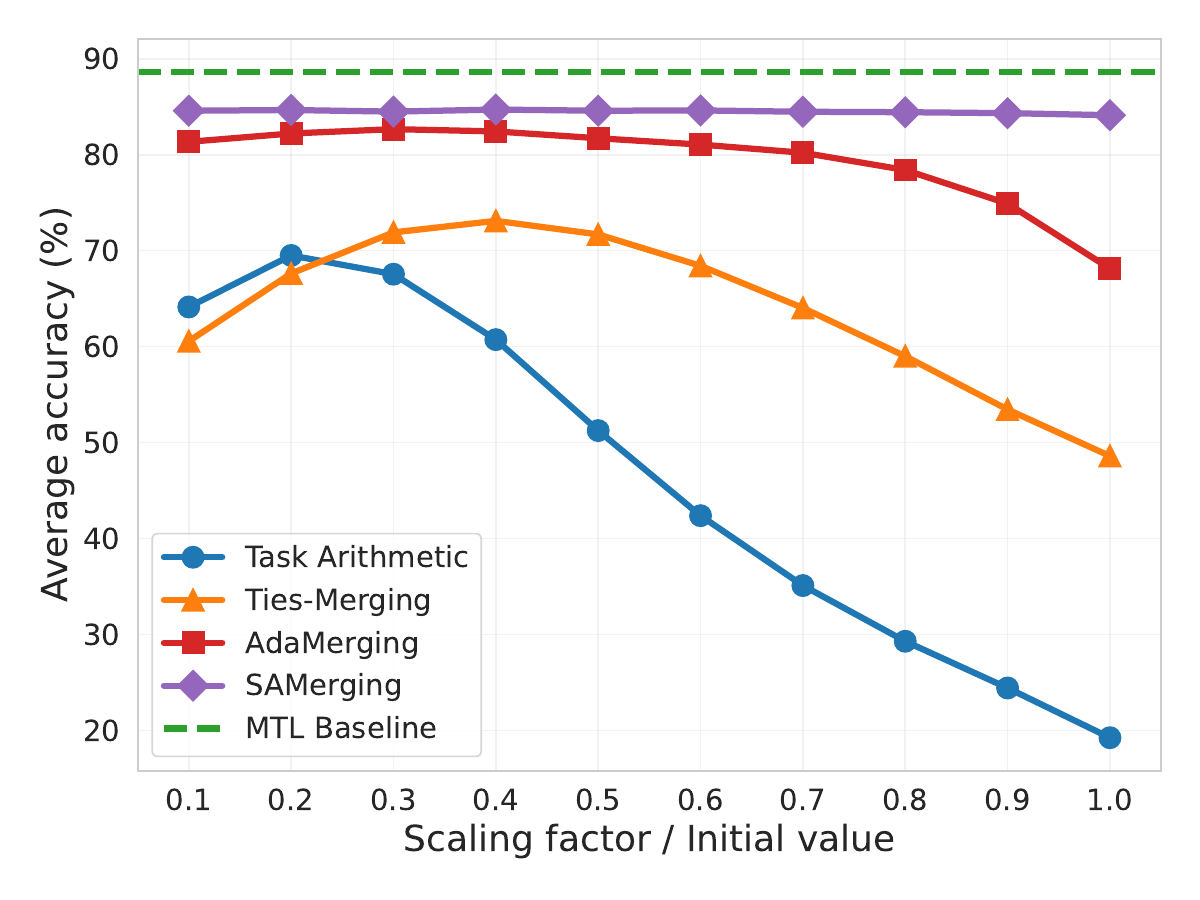}
\caption{\textbf{Sensitivity to merge scaling and initialization.} On TA-8, we compare merge scaling (TA/TIES) and initialization (AdaMerging/\SAMerging). While designed to learn coefficients, \emph{AdaMerging}'s performance is sensitive to initialization, suggesting its objective/optimizer is a bottleneck. In contrast, \SAMerging attains higher and more stable accuracy across the range.}
\label{fig:lambda_init_acc}
\end{figure}

The pretrain-fine-tune paradigm has become the dominant approach for obtaining models that can solve various tasks in fields like Natural Language Processing (NLP) and Computer Vision (CV). Recently, due to the increasing use of these models in resource-limited devices, there has been growing interest in developing models that can handle multiple tasks simultaneously. One line of research, namely model merging, leverages the existing fine-tuned models to achieve the multi-task model's parameters \citep{Ilharco_Ribeiro_Wortsman_Gururangan_Schmidt_Hajishirzi_Farhadi_2023, Matena_Raffel_2022, Wortsman_Ilharco_Gadre_others_2022}. Model merging seeks to combine fine-tuned models into a single model that retains the specialized capabilities of each task-specific fine-tuned model~\citep{Breiman_1996, Chen_Guestrin_2016, Ganaie_Hu_Malik_Tanveer_Suganthan_2022}, without the need to run multiple constituent models, so both inference cost and memory cost will be in $\mathcal{O}(1)$ instead of $\mathcal{O}(n)$ for $n$ tasks \citep{Yang_Shen_Guo_Wang_Cao_Zhang_Tao_2024}. Moreover, due to limited data access, privacy concerns, and high fine-tuning costs, model merging is gaining interest, especially in privacy-preserving settings like federated learning~\citep{Tao_Mason_Kulkarni_Boix_2025, Liu_Liu_Ye_Shen_Li_Jiang_Li_2024, Chen_Zhou_Long_Jiang_Zhang_2025, Salami_Buzzega_Mosconi_Bonato_Sabetta_Calderara_2025, Tsouvalas_Ozcelebi_Meratnia_2025}.

 One line of methods in model merging is based on the notion of ``task arithmetic'' \citep{Ilharco_Ribeiro_Wortsman_Gururangan_Schmidt_Hajishirzi_Farhadi_2023}, which treats each fine-tuned model's offset from the pretrained weights as a task vector. Scaling and summing these vectors across tasks can yield a multi-task model with performance comparable to a fine-tuned model. This insight has inspired numerous merging methods~\citep{Ortiz-Jimenez_Favero_Frossard_2023, Yadav_Tam_Choshen_Raffel_Bansal_2023, Yang_Wang_Shen_Liu_Guo_Wang_Tao_2023} and theory on when it succeeds \citep{Li_Zhang_Zhang_Wang_Liu_Chen_2025, Zhou_Chen_Chen_Zhang_Yan_2024, Wang_Wang_2024}. However, performance of these methods is highly sensitive to the scaling coefficients (see Fig.~\ref{fig:lambda_init_acc}), even for approaches like AdaMerging that aim to learn the merging coefficients~\citep{Yang_Wang_Shen_Liu_Guo_Wang_Tao_2023}. This motivates more principled ways to learn these coefficients for stronger generalization.

Flatter loss landscapes have long been associated with better generalization~\citep{Hochreiter_Schmidhuber_1997, Neyshabur_Bhojanapalli_McAllester_Srebro_2017, Petzka_Kamp_Adilova_Sminchisescu_Boley_2021, Andriushchenko_Croce_Müller_Hein_Flammarion_2023, Yue_Jiang_Ye_Gao_Liu_Zhang_2023, Haddouche_Viallard_Simsekli_Guedj_2025}, inspiring metrics and algorithms  such as Sharpness-Aware Minimization (SAM), which explicitly seeks wider minima to improve generalization \citep{Foret_Kleiner_Mobahi_Neyshabur_2021}. In multi-task learning (MTL), classical theory attributes gains to shared representations and inductive biases \citep{Caruana_1997, Baxter_2000, Argyriou_Evgeniou_Pontil_2006, Maurer_Pontil_Romera-Paredes_2016, Zhang_Yang_2021, Zakerinia_Ghobadi_Lampert_2025}, yet heterogeneous tasks often suffer negative transfer and interference \citep{Tsouvalas_Ozcelebi_Meratnia_2025, Zakerinia_Lampert_2025}. These observations suggest that solution geometry, favoring flatter minima, can mitigate cross-task interference by reducing sensitivity to parameter perturbations and stabilizing shared features, thereby improving MTL generalization~\citep{Dai_Zhu_2020, Dinh_Pascanu_Bengio_Bengio_2017, Andriushchenko_Croce_Müller_Hein_Flammarion_2023}. It is thus natural to explore flatness-aware optimization (e.g., SAM) for MTL fine-tuning and merging; \citet{Lee_Jung_Baik_2025} takes a step in this direction by modifying the fine-tuning stage to encourage flatter optima for individual tasks, thereby reducing sensitivity to perturbations introduced during merging. Alternatively, \citep{Zhang_Theus_Teney_Orvieto_Pang_Mauw_2025} leverages optimizer-induced implicit bias to control the effective noise scale to shape the merging landscape geometry for effective merging. Although effective, these approaches treat tasks uniformly and do not explicitly account for the heterogeneity across tasks.

Motivated by the limitations of the abovementioned approaches that either learn or estimate the merging coefficients, or modify the fine-tuning procedure in isolation, we introduce a new perspective on model merging grounded in generalization theory. Specifically, we derive a PAC-Bayes generalization bound that explicitly links the generalization performance of the merged model to the flatness of its loss basin. Inspired by the emerging links between flatness \citep{Haddouche_Viallard_Simsekli_Guedj_2025}, multi-task learning~\citep{Zakerinia_Lampert_2025}, and AdaMerging \citep{Yang_Wang_Shen_Liu_Guo_Wang_Tao_2023}, we propose \SAMerging, a novel method that merges fine-tuned models while explicitly seeking flatter solutions. Unlike previous approaches, \SAMerging is designed to retain task-specific performance while promoting generalization through geometry-aware merging. It achieves state-of-the-art results across diverse tasks in CV and NLP and remains robust under variations in data size and task count. In summary, our contributions are:
\begin{itemize}
    \item The central contribution of the present work is a new PAC-Bayes generalization bound for multi-task model merging, which formally captures the gap between two paradigms: merging independently fine-tuned models (\emph{a mixture of models}) versus jointly training on \emph{a mixture of data}. This bound introduces a novel discrepancy term that quantifies how merging diverges from standard multi-task learning, offering a principled lens through which to analyze and improve merging strategies.
    \item Building on this insight, we propose a new merging objective inspired by multi-teacher knowledge distillation. Specifically, we minimize the Kullback-Leibler (KL) divergence between the merged model and each task fine-tuned model to align the merged model's behavior with each fine-tuned model on its respective task. This approach not only aligns with our theoretical analysis but also outperforms entropy-based merging criteria such as those used in AdaMerging~\citep{Yang_Wang_Shen_Liu_Guo_Wang_Tao_2023}.
    \item Finally, motivated by the PAC-Bayes connection between generalization and flatness, we incorporate sharpness-aware minimization into the merging process. This promotes flatter solutions during optimization, leading to further gains in performance. Our ablation studies show that the KL-based objective and SAM contribute complementary benefits, with their combination yielding state-of-the-art results across both vision and language tasks.
\end{itemize}


\section{Related Work}
\label{sec:related-work}\vspace{-2mm}

\paragraph{Joint training for multi-task learning (MTL)}
\label{subsec:JointForMTL}
Joint training for MTL aggregates data from different tasks to learn them jointly. This method enables knowledge transfer with inductive bias and shared representations \citep{Caruana_1997, Baxter_2000, Wu_Wang_Ge_Lu_Zhou_Shan_Luo_2023}. Prior work tries to tackle this problem by working either on the (i) architecture of the models or (ii) training and optimization regime. On the architecture, work includes refining cross-task coupling \citep{Misra_Shrivastava_Gupta_Hebert_2016}, selective sharing \citep{Sun_Panda_Feris_Saenko_2020}, and mixture-of-experts to learn which experts to share per task \citep{Hazimeh_Zhao_Chowdhery_Sathiamoorthy_Chen_Mazumder_Hong_Chi_2021, Tang_Liu_Zhao_Gong_2020}. On the training and optimization, it focuses on mitigating gradient conflicts \citep{Yu_Kumar_Gupta_Levine_Hausman_Finn_2020, Liu_Liu_Jin_Stone_Liu_2024, Quinton_Rey_2025}, adjusting training weights \citep{Kendall_Gal_Cipolla_2018, Chen_Badrinarayanan_Lee_Rabinovich_2018}, or formulating MTL as multi-objective optimization to seek Pareto-optimal trade-offs with convergence guarantees and controllable preferences \citep{Lin_Zhen_Li_Zhang_Kwong_2019, Shamsian_Navon_Glazer_Kawaguchi_Chechik_Fetaya_2023}. Joint MTL can boost efficiency and generalization, but in the foundation-model era, it is often impractical to pool raw data and update large backbones due to compute and privacy constraints. Meanwhile, the rise of fine-tuned models on platforms like Hugging Face motivates post-hoc \emph{model merging}, shifting heterogeneity and task-interference challenges from data to models.

\paragraph{Model merging for MTL}
\label{subsec:MMforMTL}
In contrast to joint MTL, \emph{model merging} fuses multiple task-specific fine-tuned models into one MTL model \citep{Yang_Shen_Guo_Wang_Cao_Zhang_Tao_2024}. Data-free approaches include simple weight averaging/model soups \citep{Wortsman_Ilharco_Gadre_others_2022} and task arithmetic (TA) \citep{Ilharco_Ribeiro_Wortsman_Gururangan_Schmidt_Hajishirzi_Farhadi_2023}. TA has led to numerous new methods for merging data-free approaches, such as TIES-Merging \citep{Yadav_Tam_Choshen_Raffel_Bansal_2023}, DARE \citep{Yu_Yu_Yu_Huang_Li_2024}, PCBMerging \citep{Du_Lee_Li_Jiang_Guo_Yu_Liu_Goh_Tang_He}, and Isotropic Merging \citep{Marczak_Magistri_Cygert_Twardowski_Bagdanov_Weijer_2025}. Beyond data-free merging, data-dependent methods use unlabeled samples per task for a one-time \emph{offline} calibration. The deployed model remains a single network with no extra parameters. Concretely, Fisher Merging estimates Fisher information from gradients on unlabeled data to make the Fisher Information Matrix (FIM) \citep{Matena_Raffel_2022} and uses that as the weight for merging; RegMean/RegMean++ compute feature inner-product Gram to regularize averaging \citep{Jin_Ren_Preotiuc-Pietro_Cheng_2025, Nguyen_Huu-Tien_Suzuki_Nguyen_2025}, and AdaMerging learns (layer-/task-wise) merge coefficients by minimizing entropy \citep{Yang_Wang_Shen_Liu_Guo_Wang_Tao_2023}. By contrast, methods that learn per-task heads/masks/adapters introduce inference-time compute and memory overhead \citep{Yang_Shen_Wang_Guo_Chen_Wang_Tao_2024} and are not the focus of this work. In deployed settings, models already see inputs from target domains, so calibrating on a handful of unlabeled samples is a far weaker assumption than joint MTL's requirement for pooled training data. We thus trade minimal label-free calibration for \emph{zero} inference overhead. We achieve this by demonstrating that \SAMerging will reach state-of-the-art performance by utilizing as few as one batch of size $16$ for each task. After calibration, one backbone serves all tasks with $\Theta(1)$ memory and latency, whereas ensembles, adapters, or per-task heads/masks incur runtime and memory costs \citep{Yang_Shen_Wang_Guo_Chen_Wang_Tao_2024}.

\paragraph{Fine-tuning for mergeability vs.\ post-hoc merging.}
Beyond the choice of merging method, prior work differs in whether it (i) modifies the fine-tuning stage to produce more \emph{mergeable fine-tuned models} for tasks, or (ii) assumes fine-tuned models for tasks are already trained and then performs \emph{post-hoc merging}. The former explicitly shapes the fine-tuning trajectory so that task-specific checkpoints fall into a shared, broad low-loss basin, often evidenced by almost-linear low-loss paths between independently obtained solutions, thereby improving the reliability of subsequent weight-space merging \citep{Ainsworth_Hayase_Srinivasa_2023, Zhou_Chen_Chen_Zhang_Yan_2024}. Concretely, this can be done via sharpness-aware fine-tuning \citep{Lee_Jung_Baik_2025}, Jacobian/NTK-based regularization \citep{Yoshida_Naraki_Horie_Yamaki_Shimizu_Saito_McAuley_Naganuma_2024} along with linearized fine-tuning to promote weight disentanglement \citep{Ortiz-Jimenez_Favero_Frossard_2023}, or by controlling optimizer-induced implicit bias that shapes the merging landscape geometry \citep{Zhang_Theus_Teney_Orvieto_Pang_Mauw_2025}. In this paper, we focus on the post-hoc setting where only the final fine-tuned model checkpoints are available. As formalized by Theorem~\ref{thm:mtl-merged-pb}, our PAC-Bayes generalization bound makes fine-tuned models' \emph{mergeability} a principled criterion for predicting when post-hoc merging should succeed.

\paragraph{Knowledge distillation}
\label{subsec:KD}
We cast model merging as multi-teacher knowledge distillation \citep{Hinton_Vinyals_Dean_2015, Yang_Wang_Shen_Liu_Guo_Wang_Tao_2023, Xu_Li_Zhang_2025}: compress an \emph{ensemble of experts} into a single student by minimizing the KL divergence between the ensemble's soft predictive distribution and the student's outputs on unlabeled data, as in \citep{Hinton_Vinyals_Dean_2015}. This function-space target is robust to weight misalignment and permutations. In contrast to element-wise, feature-level merging \citep{Xu_Li_Zhang_2025}, we analyze the merged model's predictive distribution, which enables explicit excess-risk guarantees.

\paragraph{MTL Flatness and Generalization}
\label{subsec:FlatGeneralization}
Extensive evidence links flatter minima to better generalization and robustness, and PAC-Bayesian analysis formalizes the link such that when weight posterior concentrates in a broad low-loss region, complexity terms shrink, yielding non-vacuous bounds \citep{Neyshabur_Bhojanapalli_McAllester_Srebro_2017, Petzka_Kamp_Adilova_Sminchisescu_Boley_2021, Dziugaite_Roy_2017}.  Sharpness-aware Minimization (SAM) \citep{Foret_Kleiner_Mobahi_Neyshabur_2021} achieves flatter minima by penalizing the worst-case loss in a neighborhood, thereby improving generalization across architectures and tasks while demonstrating robustness to label noise \citep{Baek_Kolter_Raghunathan_2024} and quantization \citep{Na_Mehta_Strubell_2022}, making it an ideal choice for our merging settings where we don't have access to training data or labels. In MTL settings, as we demonstrate, favoring flat basins reduces cross-task sensitivity, allowing the merged model to generalize better.
\section{Methodology}
\label{sec:methodology}
We develop \SAMerging in three steps. First,  we analyze the generalization of a merged model through a PAC-Bayes lens, which reveals a \emph{cross-task heterogeneity} term that precisely captures fine-tuned model and task mismatch. The bound suggests that better generalization may be achieved in flatter loss basins. Second, we cast model merging as \emph{multi-teacher knowledge distillation} on a small, unlabeled calibration set; as shown, minimizing the KL divergence of the merged and fine-tuned model directly tightens an upper bound on the merged model's excess MTL risk. Third, we operationalize this objective with \emph{Sharpness-Aware Minimization} (SAM), which seeks flatter basins that enable the merged model to generalize better across tasks. Together, these pieces yield a data-efficient, label-free procedure that avoids inference overhead.
All proofs for the lemmas, propositions, and theorems are deferred to Appendix~\ref{apx:proofs}.
\subsection{Generalization of the merged model}
\label{subsec:generalization}

We start by establishing the generalization of the \emph{post-hoc} merged model through a PAC-Bayes analysis.

\noindent\textbf{Notation and setup.}
Let $[T]=\{1,\dots,T\}$. Each task $t$ is associated with a distribution $\mathcal D_t$ over
$\mathcal Z=\mathcal X\times\mathcal Y$ and an i.i.d.\ sample
$\mathcal S_t=\{(x_i^{(t)},y_i^{(t)})\}_{i=1}^{n_t}\sim\mathcal D_t$.
Let $f_\theta:\mathcal X\to\widehat{\mathcal Y}$ be a model with parameters $\theta\in\Theta$, and let
$\ell:\widehat{\mathcal Y}\times\mathcal Y\to[0,1]$ be $\gamma$-smooth and convex in the model scores.
For task $t$, define the population and empirical risks
\[
\mathcal L_{\mathcal D_t}(\theta)=\E_{(x,y)\sim\mathcal D_t}\big[\ell(f_\theta(x),y)\big],\qquad
\widehat{\mathcal L}_{\mathcal S_t}(\theta)=\frac{1}{n_t}\sum_{i=1}^{n_t}\ell\!\big(f_\theta(x_i^{(t)}),y_i^{(t)}\big).
\]

\noindent\textbf{Evaluation mixture.}
Let $\bm\alpha\in\Delta^{T-1}$ denote the weights of the evaluation mixture across tasks. We write
\[
\mathcal L_{\bm\alpha}(\theta)=\sum_{t=1}^T \alpha_t\,\mathcal L_{\mathcal D_t}(\theta),\qquad
\widehat{\mathcal L}_{\bm\alpha}(\theta)=\sum_{t=1}^T \alpha_t\,\widehat{\mathcal L}_{\mathcal S_t}(\theta).
\]
Joint multi-task learning would directly minimize $\widehat{\mathcal L}_{\bm\alpha}(\theta)$ using pooled data.
In contrast, in \emph{model merging} we assume only a pretrained checkpoint $\theta_0$ and task-specific
fine-tuned checkpoints $\{\theta_t\}_{t=1}^T$ are available.

\noindent\textbf{Post-hoc merging objective.}
Define the task vector $\tau_t=\theta_t-\theta_0$. Following task arithmetic, we parameterize a merged model
via layer-wise coefficients $\lambda=\{\lambda_t^l\}_{t,l}$:
\[
\theta_\lambda^{\,l}=\theta_0^{\,l}+\sum_{t=1}^T \lambda_t^{\,l}\,\tau_t^{\,l},\qquad
\theta_{\mathrm{merge}}=\theta_{\lambda^\star},\quad
\lambda^\star\in\arg\min_{\lambda}\ \widehat{\mathcal L}_{\bm\alpha}(\theta_\lambda).
\]

\noindent\textbf{PAC-Bayes view of merging.}
To study generalization, we work with distributions over parameters. Let
$P=\mathcal N(\mu_P,\Sigma_P)$ be a data-free prior on $\Theta$, and for each task $t$ let
$Q_t=\mathcal N(\mu_t,\Sigma_t)$ be a task-dependent posterior. For any posterior $Q$, define
$\mathcal L_{\mathcal D_t}(Q)=\E_{\theta\sim Q}[\mathcal L_{\mathcal D_t}(\theta)]$
(and similarly $\widehat{\mathcal L}_{\mathcal S_t}(Q)$).
We analyze a \emph{mixture posterior}
\[
Q_{\mathrm{merge}}:=\sum_{t=1}^T \beta_t Q_t,\qquad \bm\beta\in\Delta^{T-1},
\]
which can be interpreted as a randomized predictor that samples $Q_t$ with probability $\beta_t$.
Importantly, $\bm\beta$ is introduced only for analysis; it may be distinct from the algorithmic merging
coefficients $\lambda$ used to construct $\theta_{\mathrm{merge}}$.

\noindent\textbf{Flatness proxy.}
To make the bound sensitive to basin geometry, we use a squared-gradient proxy for flatness. For task $t$,
\[
\mathcal G_{\mathcal D_t}(\theta)
= \E_{(x,y)\sim\mathcal D_t}\!\left[\left\|\nabla_\theta \ell\!\big(f_\theta(x),y\big)\right\|_2^{2}\right],\qquad
\widehat{\mathcal G}_{\mathcal S_t}(\theta)
= \frac{1}{n_t}\sum_{i=1}^{n_t}\!\left\|\nabla_\theta \ell\!\big(f_\theta(x_i^{(t)}),y_i^{(t)})\big)\right\|_2^{2},
\]
and for a posterior $Q$ over $\Theta$,
\[
\mathcal G_{\mathcal D_t}(Q)
= \E_{(x,y)\sim\mathcal D_t}\!\left[\E_{h\sim Q}\left\|\nabla_h \ell\!\big(f_h(x),y\big)\right\|_2^{2}\right],\qquad
\widehat{\mathcal G}_{\mathcal S_t}(Q)
= \frac{1}{n_t}\sum_{i=1}^{n_t}\E_{h\sim Q}\!\left\|\nabla_h \ell\!\big(f_h(x_i^{(t)}),y_i^{(t)})\big)\right\|_2^{2}.
\]

\noindent
We begin with a basic but useful observation: the risk of a mixture posterior is the corresponding
mixture of risks.

\begin{lemma}
\label{lem:linear-in-q}
For any task $t$ and posteriors $\{Q_j\}_{j=1}^T$, if $Q_{\mathrm{merge}}=\sum_{j}\beta_j Q_j$, then
$$
\mathcal L_{\mathcal D_t}(Q_{\mathrm{merge}})=\sum_{j=1}^T \beta_j\,\mathcal L_{\mathcal D_t}(Q_j).
$$
\end{lemma}

\begin{proposition}
\label{prop:decomposition}
The multi-task risk $\mathcal L_{\bm \alpha}(Q_{\mathrm{merge}})$ can be decomposed as
\begin{equation*}
\mathcal L_{\bm\alpha}(Q_{\mathrm{merge}}) = \sum_{j=1}^T \beta_j\,\mathcal L_{\mathcal D_j}(Q_j) +\underbrace{\sum_{i=1}^T\sum_{j=1}^T \alpha_i\beta_j\,\big(\mathcal L_{\mathcal D_i}(Q_j)-\mathcal L_{\mathcal D_j}(Q_j)\big)}_{:=\mathcal{H}_{Q}(\bm \alpha, \bm \beta)}
\end{equation*}
\end{proposition}
The cross-task heterogeneity term $\mathcal{H}_{Q}(\bm \alpha, \bm \beta)$ measures how much worse fine-tuned model $Q_j$ performs on $\mathcal{D}_i$ compared to its own domain $\mathcal{D}_j$. It vanishes if $\mathcal{D}_i$ coincide as in the single-task setting or $Q_j\equiv Q$ for all $j$ as in a joint-trained MTL model. Now, we bound the \(\sum_{j=1}^T \beta_j\,\mathcal L_{\mathcal D_j}(Q_j)\) using the following theorem. 
\begin{theorem}\label{thm:mtl-pb}
Fix nonnegative $\{\delta_t\}_{t=1}^T$ such that $\delta = \sum_{t=1}^T \delta_t \le 1$. For any $\eta_t \in(0,2)$ for each task $t$, any data-free prior $P=\mathcal{N}(\mu_P, \Sigma_P)$, any loss $\ell:\widehat{\mathcal Y}\times\mathcal Y\to[0,1]$, with probability at least $1-\delta$ over $\{S_t\}_{t=1}^T$ from $\{\mathcal D_t\}_{t=1}^T$ with $\vert S_t \vert = n_t $, for all $Q_t:= \mathcal{N}({\theta_t}, \Sigma_t)$,
\begin{align*}
\mathcal L_{\bm\alpha}(Q_\mathrm{merge})\;\le\;
\sum_{t=1}^T \beta_t\,&\Bigg[
\frac{1}{1-\frac{\eta_t}{2}}
\left(
\hat{\mathcal L}_{S_t}(Q_t)
+\frac{\KL(Q_t\Vert P)+\log(\frac{1}{\delta_t})}{\eta_t n_t}
\!\right)+\frac{\eta_t}{2-\eta_t }\|\Sigma_t\|\,\mathcal G_{\mathcal D_t}(Q_t)\Bigg]
\\&+\sum_{i=1}^T\sum_{j=1}^T \alpha_i\beta_j\,\big(\mathcal L_{\mathcal D_i}(Q_j)-\mathcal L_{\mathcal D_j}(Q_j)\big).
\end{align*}
\end{theorem}
The bound on $\mathcal L_{\bm\alpha}(Q_{\mathrm{merge}})$ decomposes into three parts: 
(i) \emph{per-task PAC-Bayes terms} that require each fine-tuned model $Q_t$ to generalize on its own domain, 
(ii) a \emph{flatness penalty} $\mathcal G_{\mathcal D_t}(Q_t)$, which is small when the loss landscape is flat, and 
(iii) the \emph{cross-task heterogeneity}  
which captures transfer mismatch across tasks. 
Hence, excess risk is controlled when fine-tuned models are accurate in their own domains. Fine-tuned models occupy flat basins with methods that encourage flatter minima and thus reduce $\mathcal G_{\mathcal D_t}(Q_t)$, and a small cross-task heterogeneity term.

\noindent\textbf{Going from posterior bound to a single model.}~To pass from posterior-level guarantees to a single model in a non-convex landscape, we linearize the network at the pretrained point $\theta_0$ (NTK approximation as done in \citet{Jacot_Gabriel_Hongler_2020, Ortiz-Jimenez_Favero_Frossard_2023}). Let
\begin{equation*}
\Phi(x):=\nabla_\theta f_{\theta_0}(x), \qquad
f_\theta(x)=f_{\theta_0}(x)+\Phi(x)^\top(\theta-\theta_0),
\end{equation*}
for $\theta$ in a \emph{neighborhood} of $\theta_0$. This makes the score affine in $\theta$ and induces the task kernels
\begin{equation*}
\mathcal K_{\mathcal D_t}:=\E_{(x,y)\sim\mathcal D_t}\!\big[\Phi(x)\Phi(x)^\top\big],\qquad
\widehat{\mathcal K}_{\mathcal S_t}:=\frac{1}{n_t}\sum_{i=1}^{n_t}\Phi\!\big(x_i^{(t)}\big)\Phi\!\big(x_i^{(t)}\big)^\top.
\end{equation*}
Within this local model, convexity and $\gamma$-smoothness in the score allows us to relate Gaussian posteriors $Q_t=\mathcal N(\mu_t,\Sigma_t)$ to their means via trace terms in $\mathcal K$, enabling a single-model bound for the merged parameter:
\[
\theta_{\mathrm{merge}}:= \E \left[Q_\mathrm{merge}\right] = \E\left[\sum_{j=1}^T \beta_j\,Q_j\right] = \sum_{j=1}^T \beta_j\,\E \left[Q_j\right] = \sum_{j=1}^T \beta_j\,\mu_j,\qquad
\Delta_j:=\mu_j-\theta_{\mathrm{merge}}.
\]
Note that posterior-level bounds themselves do not rely on NTK; the NTK is used only to pass from posterior-level to a single model. We next establish bounds on the loss and flatness of the posterior, which will be leveraged in the subsequent derivation.

\begin{lemma}
\label{lem:L-by-mu}
Assume the NTK linearization around $\theta_0$. Let $\ell$ be convex and $\gamma$-smooth in the score. For $Q=\mathcal N(\mu,\Sigma)$ and distribution $\mathcal D$,
\[
\mathcal L_{\mathcal D}(\mu)
\;\le\;
\mathcal L_{\mathcal D}(Q)
\;\le\;
\mathcal L_{\mathcal D}(\mu)+\frac{\gamma}{2}\,\mathrm{tr}\!\big(\Sigma\,\mathcal K_{\mathcal D}\big).
\]
The empirical statement follows verbatim with $\mathcal D\to\mathcal S$ and $\mathcal K_{\mathcal D}\to\widehat{\mathcal K}_{\mathcal S}$.
\end{lemma}

\begin{lemma}
\label{lem:G_by_mu}
Let $\ell$ be convex and $\gamma$-smooth in the score, and let $Q=\mathcal N(\mu,\Sigma)$. Then, for $\mathcal D$,
\[
\mathcal G_{\mathcal D}(Q)
\;\le\;
\Big(\sqrt{\mathcal G_{\mathcal D}(\mu)}+\gamma\,\sqrt{\mathrm{tr}\!\big(\Sigma\,\mathcal K_{\mathcal D}^2\big)}\Big)^2.
\]
The empirical statement follows verbatim with $\mathcal D\to\mathcal S$ and $\mathcal K_{\mathcal D}\to\widehat{\mathcal K}_{\mathcal S}$.
\end{lemma}
Now, under NTK, we bound the heterogeneity term with loss and flatness at $\theta_\mathrm{merge}$.

\begin{lemma}
\label{lem:det-hetero-term-bound}
Under NTK, with loss $\ell$ being convex and $\gamma$-smooth in score, let $\mathcal K_\alpha=\sum_{t=1}^T\alpha_t\mathcal K_{\mathcal D_t}$, $\mathcal K_\beta=\sum_{j=1}^T\beta_j\mathcal K_{\mathcal D_j}$, and $\Delta_j = \mu_j - \theta_{\mathrm{merge}}$.
Then, for $\theta_{\mathrm{merge}}$,
\begin{align*}
\mathcal{H}_{Q}(\bm \alpha, \bm \beta)
\;\le\;&
\left(\mathcal L_{\bm\alpha}(\theta_{\mathrm{merge}})-\mathcal L_{\bm\beta}(\theta_{\mathrm{merge}})\right)
\\&+\sqrt{2\!\left(\sum_{t=1}^T \alpha_t \mathcal G_{\mathcal D_t}(\theta_{\mathrm{merge}})+\sum_{j=1}^T \beta_j \mathcal G_{\mathcal D_j}(\theta_{\mathrm{merge}})\right)}
\;\sqrt{\sum_{j=1}^T \beta_j \|\Delta_j\|_2^2}\\
& \quad+\frac{\gamma}{2}\sum_{j=1}^T \beta_j\!\left[\Delta_j^\top(\mathcal K_\alpha+\mathcal K_\beta)\Delta_j + \mathrm{tr}\left(\Sigma_j\mathcal{K}_{\bm\alpha}\right)\right]. 
\end{align*}
\end{lemma}
With these, we are ready to state our main PAC-Bayes bound for the merged model.

\begin{theorem}\label{thm:mtl-merged-pb}
Assume the NTK regime. Fix nonnegative $\{\delta_t\}_{t=1}^T$ with $\delta = \sum_{t=1}^T \delta_t \le 1$, task weights $\bm\alpha,\bm\beta\in\Delta^{T-1}$, constants $\eta_t\in(0,2)$ for each task $t$, a data-free prior $P\in\mathcal M(\mathcal H)$, and a $\gamma$-smooth loss $\ell:\widehat{\mathcal Y}\times\mathcal Y\to[0,1]$ convex in its score argument. Over $\{S_t\}_{t=1}^T$ from $\{\mathcal D_t\}_{t=1}^T$ with $\vert S_t \vert = n_t $, with probability at least $1-\delta$, the following holds  for all Gaussian posteriors $Q_t=\mathcal N(\mu_t,\Sigma_t)$ and for the merged parameter 
$\theta_{\mathrm{merge}}:=\sum_{j=1}^T \beta_j\,\mu_j$ with $\Delta_j:=\mu_j-\theta_{\mathrm{merge}}$:

\begin{align*}
\mathcal L_{\bm\alpha}\left(\theta_\mathrm{merge}\right)\;\le\;
&\sum_{t=1}^T \beta_t\,\Bigg[
\frac{1}{1-\frac{\eta_t}{2}}
\left(
\hat{\mathcal L}_{S_t}\left(\mu_t\right) + \frac{\gamma}{2}\operatorname{tr}\left(\Sigma_t\mathcal{K}_{\mathcal{D}_t}\right)
+\frac{\KL\left(Q_t\Vert P\right)+\log\left(\frac{1}{\delta_t}\right)}{\eta_t n_t}\right)
\\&+\frac{\eta_t}{2-\eta_t }\|\Sigma_t\|\,\left(\sqrt{\mathcal G_{\mathcal D}(\mu_t)} + \gamma \sqrt{\operatorname{tr}\left(\Sigma_t\mathcal{K}_{\mathcal{D}_t}^2\right)}\right)^2\Bigg]+\left[\mathcal L_{\bm\alpha}(\theta_{\mathrm{merge}})-\mathcal L_{\bm\beta}(\theta_{\mathrm{merge}})\right]
\\&
+\sqrt{2\!\left(\sum_{t=1}^T \alpha_t \mathcal G_{\mathcal D_t}(\theta_{\mathrm{merge}})+\sum_{j=1}^T \beta_j \mathcal G_{\mathcal D_j}(\theta_{\mathrm{merge}})\right)}
\;\sqrt{\sum_{j=1}^T \beta_j \|\Delta_j\|_2^2}\\
&\qquad+\frac{\gamma}{2}\sum_{j=1}^T \beta_j\!\left[\Delta_j^\top(\mathcal K_\alpha+\mathcal K_\beta)\Delta_j + \mathrm{tr}\left(\Sigma_j\mathcal{K}_{\bm\alpha}\right)\right]
\end{align*}
where $\mathcal{K}_{\bm\alpha} = \sum_{t=1}^T \alpha_t \mathcal{K}_{\mathcal{D}_t}$ and $\mathcal{K}_{\bm\beta} = \sum_{t=1}^T \beta_t \mathcal{K}_{\mathcal{D}_t}$.
\end{theorem}

Theorem~\ref{thm:mtl-merged-pb} implies that the risk \(\mathcal L_{\bm\alpha}(\theta_{\mathrm{merge}})\) is controlled by: \emph{(i) Per-task PAC-Bayes contributions} computed at each fine-tuned model \(Q_t=\mathcal N(\mu_t,\Sigma_t)\) that combine the fine-tuned model's empirical loss with its flatness on its own domain via \(\mathcal G_{\mathcal D_t}\). This component directly explains why using flatter fine-tuned models improves merging, as in \citet{Lee_Jung_Baik_2025}, since flatter fine-tuned models tighten these terms. \emph{(ii) A cross-task heterogeneity contribution} that, under the NTK assumption, is further bounded by the mixture gap \(\mathcal L_{\bm\alpha}(\theta_{\mathrm{merge}})-\mathcal L_{\bm\beta}(\theta_{\mathrm{merge}})\), flatness of the merged model  \(\sum_t \alpha_t\mathcal G_{\mathcal D_t}(\theta_{\mathrm{merge}})\), the dispersion \(\sum_j \beta_j \|\Delta_j\|_2^2\),  and the quadratics \(\sum_j \beta_j\,\Delta_j^\top(\mathcal K_{\bm\alpha}+\mathcal K_{\bm\beta})\Delta_j\).

\emph{Operationally}, the bound suggests desirable design choices for any label-free merging routine: (i) favoring flatter basins (e.g., via sharpness-aware perturbations) to directly shrink the flatness terms \(\mathcal G_{\mathcal D_t}(\theta_{\mathrm{merge}})\), (ii) selecting coefficients \(\bm\beta\) that pull \(\theta_{\mathrm{merge}}\) to reduce dispersion and the kernel-weighted penalties, and (iii) aligning \(\bm\beta\) with the evaluation mixture \(\bm\alpha\) to minimize the gap \(\mathcal L_{\bm\alpha}(\theta_{\mathrm{merge}})-\mathcal L_{\bm\beta}(\theta_{\mathrm{merge}})\). Furthermore, we know that NTK approximation is best within a limited distance, so we encourage the merged model to be \emph{around the pretrained} and not get too far. Taken together, these choices would tighten the bound. The bound also clarifies failure modes: fine-tuned models that are simultaneously sharp and far from consensus increase the heterogeneity terms, and any principled algorithm should accordingly assign them smaller coefficients in \(\bm\beta\).

\begin{remark}
    We distinguish our approach from prior work that modifies the fine-tuning stage by training task-specific fine-tuned models with sharpness-aware minimization to encourage flatter minima for individual fine-tuned models, and subsequently merges these fine-tuned models to obtain a model that is robust to parameter perturbations to better preserve individual task performance. In contrast, our method operates on already fine-tuned models and does not interfere with or alter the fine-tuning procedure of the individual tasks. Our theoretical analysis reveals that fine-tuning models with SAM and applying SAM during the merging stage address two distinct terms in the generalization bound. Specifically, fine-tuning models using SAM as in \cite{Lee_Jung_Baik_2025} shrinks the term \(\sqrt{\mathcal G_{\mathcal D}(\mu_t)}\), while \SAMerging  pushes the \emph{merged model} itself toward flatter basins by shrinking the \(\sqrt{2\!\left(\sum_{t=1}^T \alpha_t \mathcal G_{\mathcal D_t}(\theta_{\mathrm{merge}})+\sum_{j=1}^T \beta_j \mathcal G_{\mathcal D_j}(\theta_{\mathrm{merge}})\right)}\) term.
\end{remark}

\subsection{Model Merging as Multi-Teacher Knowledge Distillation}
We now turn to rigorously justify estimating the merging coefficients $\lambda$ via multi-teacher knowledge distillation: minimizing the KL divergence between the student's (merged model) and the teachers' (fine-tuned models) distributions tightens a provable upper bound on the merged model's classification error. We first fix notation and losses, then recall standard links between distributional distances and $0$–$1$ risk that underlie the single-task bound and its multi-task extension.

\begin{definition}
\label{def:losses}
Let $\mathcal{Y}$ be a finite label set and let $(x, y)\sim\mathcal D$. 
For each $x$, denote by $y(\cdot\mid x) \in \mathbb{R}^{\vert \mathcal{Y}\vert}$ the \emph{true} conditional label distribution, 
by $p(\cdot\mid x)$ a (possibly misspecified) teacher/fine-tuned model, and by $q(\cdot\mid x)$ a student/merged model. 
Let $h_q(x)\in\argmax_{y\in\mathcal Y} q(y\mid x)$ be the deterministic classifier induced by $q$, 
and define the $0$-$1$ risk under a conditional distribution $s(\cdot\mid x)$ by 
\[ 
\mathcal{L}^{0-1}_{s}(h_q):=\E_{(x, y) \sim\mathcal D}\big[\,1-s(h_q(x)\mid x)\,\big],
\]
and the Bayes optimal risk by
\[
\mathcal{L}^{0-1,\star}_{s}:=\E_{(x, y) \sim\mathcal D}\big[\,1-\max_{y} s(y\mid x)\,\big]. 
\]
\end{definition}

Now we propose the excess risk bound for a single task, which serves as the foundation for our main multi-task result.

\begin{lemma}[Single-Task Excess Risk Bound]
\label{lem:single_task_bound}
For any task $t$, let $y_t(\cdot|x)$ be the true data distribution, $p_t(\cdot|x)$ be a teacher, and $q_\lambda(\cdot|x)$ be the student. Let $h_\lambda$ be the classifier induced by the student.  The student's excess $0$-$1$ risk is bounded by:
\begin{equation*}
\mathcal{L}^{0-1}_{y_t}(h_\lambda)
\;\le\;
\sqrt{2\,\E_{x \sim \mathcal{D}_t}\,\KL\big(p_t(\cdot|x) \;\Vert\; q_\lambda(\cdot|x)\big)}
+
\sqrt{2\,\E_{x \sim \mathcal{D}_t}\,\KL\big(y_t(\cdot|x) \;\Vert\; p_t(\cdot|x)\big)}.
\end{equation*}
\end{lemma}
Extending this result, we arrive at the main theorem in this section, which bounds the average excess risk of the merged model across all tasks.

\begin{theorem}[Multi-Task Excess Risk Bound]
\label{thm:mtl_excess_bound}
Let there be $T$ tasks. For each task $t$, let $y_t(\cdot|x)$ be the true distribution, $p_t(\cdot|x)$ be the teacher (fine-tuned model) for task $t$, and $q_\lambda(\cdot|x)$ be the student (merged model). Let $h_\lambda$ be the classifier induced by the student. For evaluation weights $\bm \alpha \in \Delta^{T-1}$, the student's average excess risk is bounded by:
\begin{align}
\label{eq:mtl_excess_bound}
\sum_{t=1}^T \alpha_t \mathcal{L}^{0-1}_{y_t}(h_\lambda)
\;\le\; & \sqrt{2 \sum_{t=1}^T \alpha_t \E_{x \sim \mathcal{D}_t} \KL\big(p_t(\cdot|x) \;\Vert\; q_\lambda(\cdot|x)\big)} \\
& + \sqrt{2 \sum_{t=1}^T \alpha_t \E_{x \sim \mathcal{D}_t} \KL\big(y_t(\cdot|x) \;\Vert\; p_t(\cdot|x)\big)}. \nonumber
\end{align}
\end{theorem} 

Theorem~\ref{thm:mtl_excess_bound} decomposes the average excess risk into (i) an optimizable \emph{student–teacher fit} term, given by the KL divergence $\KL(p_t\,\Vert\,q_\lambda)$, and (ii) a fixed \emph{teacher error} term that depends only on the fine-tuned models. Since the latter is independent of $\lambda$, tightening the bound reduces to minimizing the fit term. Our objective achieves this by minimizing the student–teacher KL divergence on calibration data, thereby directly tightening the proven risk bound for the merged model. In contrast, methods such as AdaMerging \citep{Yang_Wang_Shen_Liu_Guo_Wang_Tao_2023} use entropy minimization without an explicit excess-risk guarantee.

\subsection{\SAMerging Objective and Optimization}
By Theorem~\ref{thm:mtl_excess_bound}, the average excess risk is controlled by the student–teacher fit term $\KL(p_t\Vert q_\lambda)$. In other words, the discrepancy between the true task distribution $p_t$ and the model distribution $q_\lambda$ directly governs the tightness of the bound. Motivated by this and by the flatness terms in our PAC-Bayes bound in Theorem~\ref{thm:mtl-merged-pb}, we minimize the multi-teacher KD loss and search for \emph{flat} solutions via SAM \citep{Foret_Kleiner_Mobahi_Neyshabur_2021}:
\begin{equation*}
\mathcal{L}_{\mathrm{KD}}(\lambda)
\;=\;
\sum_{t=1}^T \alpha_t \;
\E_{x\in \mathcal{B}_t}\!\left[
\mathrm{KL}\!\left( p_t(\cdot\mid x)\,\Vert\, q_\lambda(\cdot\mid x) \right)
\right]
\end{equation*}
where $\mathcal{B}_t$ is a batch of unlabeled data for task $t$.
The SAM-enhanced problem that we minimize is
\begin{equation*}
\label{eq:samerging-minmax}
\min_{\lambda}\;
\max_{\|\varepsilon\|_2 \le \rho}\;
\mathcal{L}_{\mathrm{KD}}(\lambda+\varepsilon),
\end{equation*}
where $\rho>0$ controls the SAM neighborhood. In practice, we usually set $\alpha_t=\frac{1}{T}$.  We can also initialize the coefficient from $0$ or add a norm penalty to encourage the remaining in the NTK-faithful neighborhood of $\theta_0$, which helps reduce the dispersion and kernel-weighted penalties highlighted by Theorem~\ref{thm:mtl-merged-pb}. The pseudo-code of \SAMerging is provided in Appendix~\ref{apx:samerging-alg}.

\begin{remark}
While both \SAMerging and AdaMerging leverage unlabeled data to learn merging coefficients, they fundamentally differ in their objectives and theoretical grounding. AdaMerging minimizes predictive entropy, a heuristic that encourages the model to be confident (producing ``peaky'' distributions) but does not necessarily ensure correctness or alignment with the fine-tuned model models. Consequently, it lacks explicit guarantees regarding generalization error. In contrast, \SAMerging casts merging as multi-teacher distillation, as in Theorem~\ref{thm:mtl_excess_bound}, this objective directly tightens an upper bound on the merged model's excess risk. Furthermore, while AdaMerging relies on standard optimization, rendering it brittle to initialization, \SAMerging, based onthe  bound in Theorem~\ref{thm:mtl-merged-pb}, integrates SAM to explicitly navigate toward flat loss basins.
\end{remark}
\section{Experiments}
\vspace{-1mm}We first introduce our experimental setup in Section~\ref{subsec:exp_setup} and then report the results in Section~\ref{subsec:results}. Full tasks and baseline descriptions (\ref{apx:exp-baseline}), experimental setups and ablations (\ref{apx:exp-setup}) and tables (\ref{apx:exp-res}) are provided in the appendix.

\subsection{Experimental Setup}\label{subsec:exp_setup}

\noindent\textbf{Tasks and data.}~We evaluate generalization across increasing interference regimes on four suites following \citet{Ilharco_Ribeiro_Wortsman_Gururangan_Schmidt_Hajishirzi_Farhadi_2023, Wang_Dimitriadis_Ortiz-Jimenez_Fleuret_Frossard_2024}: (i) \textbf{TA-8} (8 image classification tasks), (ii) \textbf{TALL-14}  (TA-8 + six more tasks), (iii) \textbf{TALL-20} (TALL-14 + six more tasks), and (iv) \textbf{GLUE} (7 NLP tasks). Vision backbones are CLIP ViT-B/32 and ViT-L/14. For NLP GLUE tasks, we use GPT-2, fine-tuned per task, to obtain task vectors. The setup is similar to \citet{Wang_Dimitriadis_Ortiz-Jimenez_Fleuret_Frossard_2024}.


\noindent\textbf{Baselines and metric.}~We compare against standard and state-of-the-art merging baselines. \textbf{Data-free}: Simple Averaging (SA) \citep{Wortsman_Ilharco_Gadre_others_2022}, Task Arithmetic (TA) \citep{Ilharco_Ribeiro_Wortsman_Gururangan_Schmidt_Hajishirzi_Farhadi_2023}, TIES-Merging (TIES) \citep{Yadav_Tam_Choshen_Raffel_Bansal_2023}, and Isotropic Merging \citep{Marczak_Magistri_Cygert_Twardowski_Bagdanov_Weijer_2025}. \textbf{Data-dependent}: Fisher Merging \citep{Matena_Raffel_2022}, RegMean \citep{Jin_Ren_Preotiuc-Pietro_Cheng_2025} / RegMean++ \citep{Nguyen_Huu-Tien_Suzuki_Nguyen_2025}, and AdaMerging \citep{Yang_Wang_Shen_Liu_Guo_Wang_Tao_2023} (we use the \emph{layer-wise} variant, which consistently outperforms the task-wise version). We report average multi-task accuracy ($\mathrm{Acc.}\,\%$) and normalized accuracy relative to the mean accuracy of the individual fine-tuned model ($\mathrm{Norm.}\,\%$).

\newcommand{\NA}{\multicolumn{1}{c}{\textemdash}}
\newcommand{\te}[1]{\multicolumn{1}{c}{#1}}
\newcommand{\sota}[1]{{\bfseries #1{$^{\color{ForestGreen}\text{\tiny +0.1}\uparrow}$}}}

\begin{table*}[t]
\centering
\begin{threeparttable}
\setlength{\tabcolsep}{3.5pt}
\scriptsize

\begin{tabular*}{\textwidth}{@{\extracolsep{\fill}} l *{12}{c}}
\toprule
\multirow{3}{*}{Method} &
\multicolumn{6}{c}{\textbf{CLIP ViT-B/32}} &
\multicolumn{6}{c}{\textbf{CLIP ViT-L/14}} \\
\cmidrule(lr){2-7} \cmidrule(lr){8-13}
& \multicolumn{2}{c}{TA-8} & \multicolumn{2}{c}{TALL-14} & \multicolumn{2}{c}{TALL-20}
& \multicolumn{2}{c}{TA-8} & \multicolumn{2}{c}{TALL-14} & \multicolumn{2}{c}{TALL-20} \\
\cmidrule(lr){2-3} \cmidrule(lr){4-5} \cmidrule(lr){6-7}
\cmidrule(lr){8-9} \cmidrule(lr){10-11} \cmidrule(lr){12-13}
& {Acc.} & {Norm.} & {Acc.} & {Norm.} & {Acc.} & {Norm.}
& {Acc.} & {Norm.} & {Acc.} & {Norm.} & {Acc.} & {Norm.} \\
\midrule
\multicolumn{13}{l}{\it Bases} \\
Pretrained
  & \te{48.2} & \te{53.4} & \te{56.9} & \te{64.3} & \te{55.6} & \te{61.9}
  & \te{64.6} & \te{68.5} & \te{69.1} & \te{74.0} & \te{65.6} & \te{70.2} \\
Fine-tuned
  & \te{90.3} & \te{100.0} & \te{88.5} & \te{100.0} & \te{89.8} & \te{100.0}
  & \te{94.3} & \te{100.0} & \te{93.4} & \te{100.0} & \te{93.5} & \te{100.0} \\
MTL
& \te{88.5} & \te{98.0} & \te{87.7} & \te{99.1} & \te{88.9} & \te{99.0}
& \te{92.3} & \te{98.0} & \te{91.6} & \te{98.1} & \te{91.8} & \te{98.2} \\
\midrule
\multicolumn{13}{l}{\it Data-free} \\
Simple Averaging
  & \te{66.3} & \te{73.4} & \te{65.4} & \te{73.4} & \te{61.1} & \te{68.0}
  & \te{79.9} & \te{86.5} & \te{77.5} & \te{83.0} & \te{71.1} & \te{76.0} \\
Task Arithmetic
  & \te{67.5} & \te{74.8} & \te{66.5} & \te{75.1} & \te{61.1} & \te{68.0}
  & \te{82.1} & \te{87.1} & \te{77.9} & \te{83.4} & \te{71.1} & \te{76.0} \\
TIES-Merging
  & \te{71.9} & \te{79.6} & \te{67.6} & \te{76.4} & \te{62.7} & \te{69.8}
  & \te{83.8} & \te{88.9} & \te{77.8} & \te{83.3} & \te{72.3} & \te{77.3} \\
Isotropic Merging
  & \te{78.8} & \te{87.3} & \te{78.8} & \te{89.0} & \te{73.5} & \te{81.8}
  & \te{90.3} & \te{95.8} & \te{\underline{89.8}} & \te{\underline{96.1}} & \te{\underline{84.8}} & \te{\underline{90.7}} \\
{ PCB Merging}
  & \te{ 75.4} & \te{ 83.5} & \te{ 70.3} & \te{ 79.4} & \te{ 64.1} & \te{ 71.4}
  & \te{ 84.2} & \te{ 89.3} & \te{ 80.4} & \te{ 86.1} & \te{ 72.6} & \te{ 77.6} \\
\midrule
\multicolumn{13}{l}{\it Data-dependent} \\
Fisher ($k{=}1600$)
  & \te{70.5} & \te{78.1} & \te{67.1} & \te{75.8} & \te{62.2} & \te{69.3}
  & \te{73.3} & \te{77.7} & \te{75.4} & \te{80.7} & \te{70.4} & \te{75.3} \\
RegMean ($k{=}1600$)
  & \te{80.5} & \te{89.1} & \te{76.1} & \te{86.0} & \te{70.0} & \te{78.0}
  & \te{89.0} & \te{94.4} & \te{86.3} & \te{92.4} & \te{78.8} & \te{87.8} \\
RegMean++ ($k{=}1600$)
  & \te{\underline {84.2}} & \te{\underline{93.2}} & \te{\underline{79.8}} & \te{\underline{90.2}} & \te{\underline{74.0}} & \te{\underline{82.4}}
  & \te{88.3} & \te{93.6} & \te{87.9} & \te{94.1} & \te{82.5} & \te{88.2} \\
AdaMerging LW ($k{=}1600$)
  & \te{73.7} & \te{81.6} & \te{71.1} & \te{80.3} & \te{61.5} & \te{68.5}
  & \te{85.1} & \te{90.2} & \te{81.9} & \te{87.7} & \te{71.5} & \te{76.5} \\
AdaMerging LW ($k{=}16000$)
  & \te{82.6} & \te{91.5} & \te{77.7} & \te{87.8} & \te{69.4} & \te{77.3}
  & \te{\underline{91.0}} & \te{\underline{96.5}} & \te{87.2} & \te{93.4} & \te{79.0} & \te{84.5} \\
\multicolumn{13}{l}{\bf \it Ours} \\
\bf \SAMerging ($k{=}1600$)
  & \te{\bf 87.1} & \te{\bf 96.5} & \te{\bf 83.7} & \te{\bf 94.6} & \te{\bf 81.1} & \te{\bf 90.3}
  & \te{\bf 92.6} & \te{\bf 98.2} & \te{\bf 90.7} & \te{\bf 97.1} & \te{\bf 89.9} & \te{\bf 96.1} \\
\bottomrule
\end{tabular*}
\caption{Average results across CLIP ViT backbones on TA-8, TALL-14, and TALL-20. Best result is \textbf{bold} and second best result is \underline{underlined}. Accuracy (Acc., \%) and normalized accuracy vs.\ Avg.\ fine-tuning (Norm., \%; Avg.\ FT is 100\%). Data-dependent methods use a maximum of $k$ unlabeled samples per task for adaptation.}
\label{tab:overall_merging_clip}
\end{threeparttable}
\end{table*}
\begin{table*}[!h]
\centering
\label{tab:samerging}
{
\begin{threeparttable}
\scriptsize
\begin{tabular}{l *{12}{c}}
\toprule
\multirow{3}{*}{Method} &
\multicolumn{6}{c}{\textbf{CLIP ViT-B/32}} &
\multicolumn{6}{c}{\textbf{CLIP ViT-L/14}} \\
\cmidrule(lr){2-7} \cmidrule(lr){8-13}
& \multicolumn{2}{c}{TA-8} & \multicolumn{2}{c}{TALL-14} & \multicolumn{2}{c}{TALL-20}
& \multicolumn{2}{c}{TA-8} & \multicolumn{2}{c}{TALL-14} & \multicolumn{2}{c}{TALL-20} \\
\cmidrule(lr){2-3} \cmidrule(lr){4-5} \cmidrule(lr){6-7}
\cmidrule(lr){8-9} \cmidrule(lr){10-11} \cmidrule(lr){12-13}
& {Acc.} & {Norm.} & {Acc.} & {Norm.} & {Acc.} & {Norm.}
& {Acc.} & {Norm.} & {Acc.} & {Norm.} & {Acc.} & {Norm.} \\
\midrule
Fine-tuned
  & \te{90.3} & \te{100.0} & \te{88.5} & \te{100.0} & \te{89.8} & \te{100.0}
  & \te{94.3} & \te{100.0} & \te{93.4} & \te{100.0} & \te{93.5} & \te{100.0} \\
MTL
  & \te{88.5} & \te{98.0} & \te{87.7} & \te{99.1} & \te{88.9} & \te{99.0}
  & \te{92.3} & \te{97.9} & \te{91.6} & \te{98.1} & \te{91.8} & \te{98.2} \\
ProDistill ($k{=}16$)
  & \te{81.1} & \te{89.8} & \te{80.5} & \te{91.0} & \te{77.8} & \te{86.6}
  & \te{87.2} & \te{92.5} & \te{89.0} & \te{95.3} & \te{\bf 86.8} & \te{\bf 92.8} \\
\bf SAMerging ($k{=}16$)
  & \te{\bf 83.8} & \te{\bf 92.8} & \te{\bf 81.2} & \te{\bf 91.8} & \te{\bf 77.9} & \te{\bf 85.7}
  & \te{\bf 91.2} & \te{\bf 96.7} & \te{\bf 89.1} & \te{\bf 95.4} & \te{85.8} & \te{91.8} \\
\bottomrule
\end{tabular}
\caption{{ Comparison of \SAMerging and ProDistill on CLIP ViT backbones on TA-8, TALL-14, and TALL-20. Best result is \textbf{bold}.}}
\label{tab:prodistill}
\end{threeparttable}
}
\end{table*}
\subsection{Results}
\label{subsec:results}
\textbf{Overall performance.}
Table~\ref{tab:overall_merging_clip} summarizes multi-task results across CLIP backbones. \SAMerging achieves the best average accuracy and normalized accuracy across all settings, outperforming both data-free and data-dependent baselines. Notably, while AdaMerging (layer-wise) requires $k{=}16\mathrm{K}$ unlabeled samples per task for adaptation, \SAMerging uses only $k{=}1.6\mathrm{K}$ and yields higher accuracy, indicating better data efficiency. Relative to strong pruning- or arithmetic-based methods, \SAMerging consistently outperforms them and closes the gap to the single-task fine-tuned model, while preserving the advantages of post-hoc merging (i.e., no joint training). Additionally, we compare our approach against ProDistill, another few-shot distillation method. As shown in Table~\ref{tab:prodistill}, SAMerging consistently achieves higher accuracy than ProDistill across all benchmarks and backbones.\\
\textbf{Scaling with the number of tasks.}
{ Table~\ref{tab:overall_merging_clip}} shows that the absolute gains of \SAMerging widen as the number of merged tasks increases. This trend supports our design objective: by explicitly stabilizing adaptation with sharpness-aware updates and a $\KL$ guidance term, \SAMerging mitigates cross-task interference that typically accumulates with more fine-tuned models being merged.
\newcommand{\drop}[1]{\scriptsize${\color{Mahogany}\text{\tiny -#1}\downarrow}$}

\begin{figure}[h!]
    \centering
    \includegraphics[width=1\linewidth]{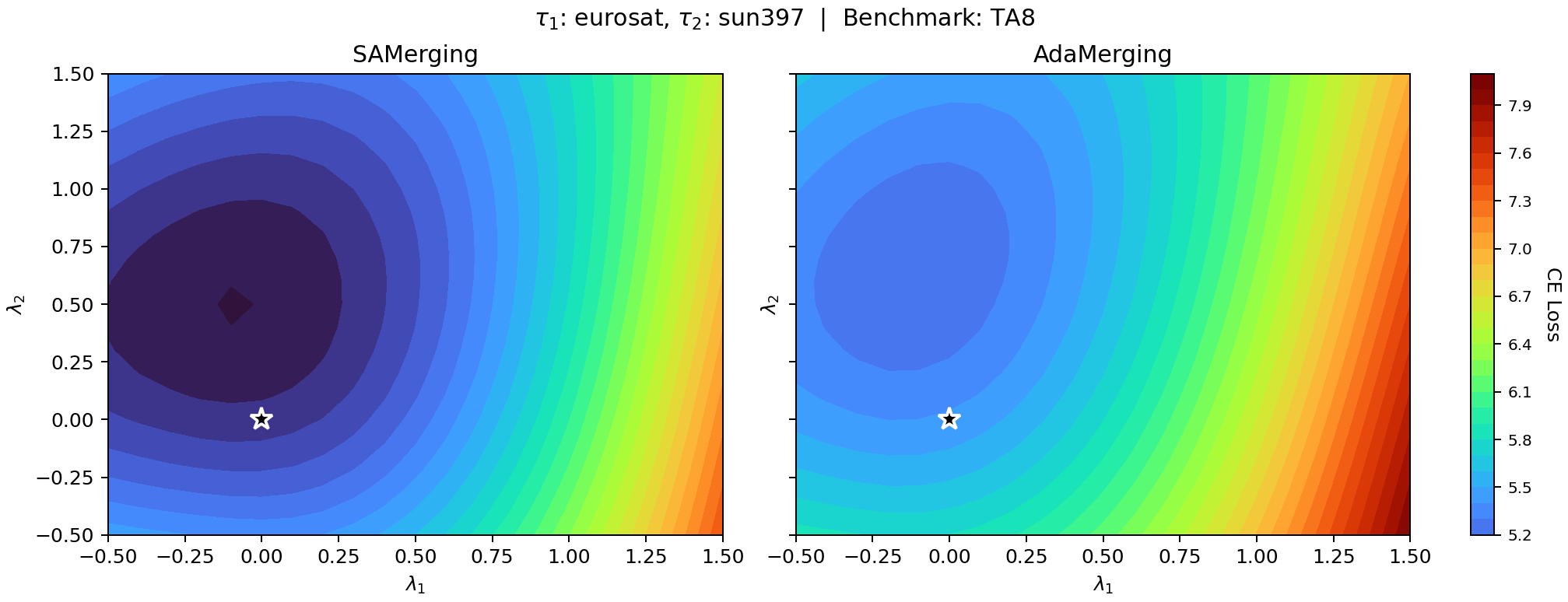}
    \caption{ The loss landscape around the merged model of \SAMerging and AdaMerging on TA-8 with perturbing along EuroSAT and SUN397.}
    \label{fig:eurosat_sun_heatmap}
\end{figure}

\begin{figure}[h!]
    \centering
    \includegraphics[width=1\linewidth]{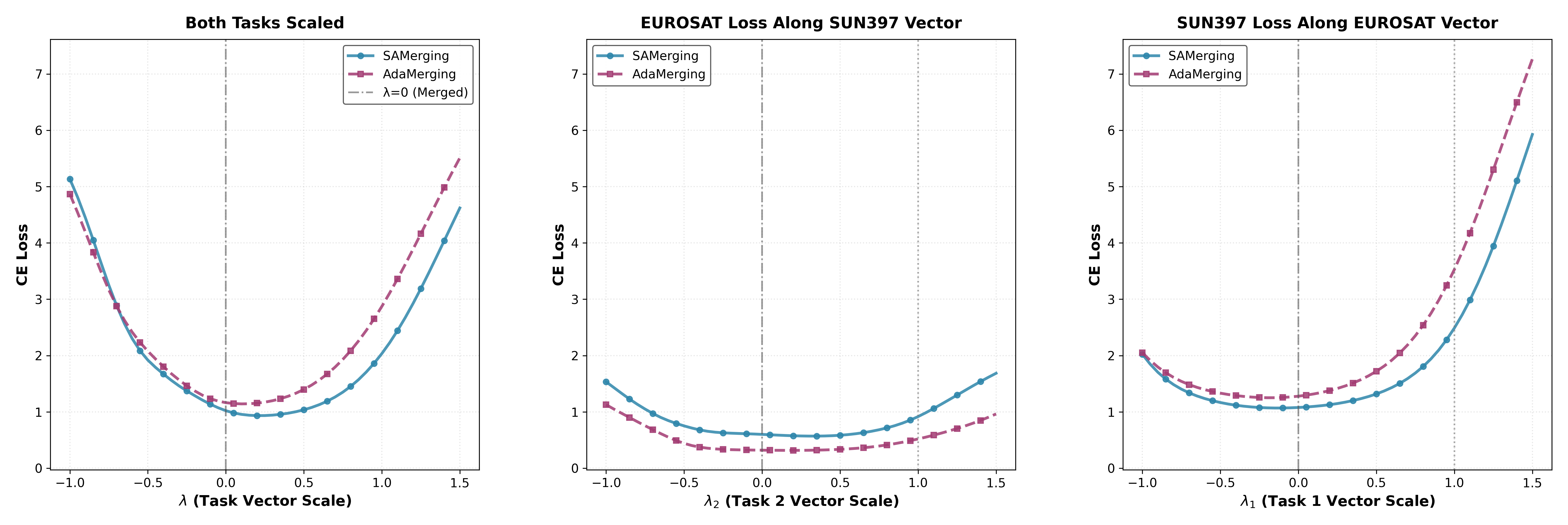}
    \caption{ The loss behavior around the merged model of \SAMerging and AdaMerging on TA-8 by perturbing along task vectors of EuroSAT and SUN397. }
    \label{fig:eurosat_sun_comparison}
\end{figure}

\begin{figure}[h!]
    \centering
    \includegraphics[width=0.4\linewidth]{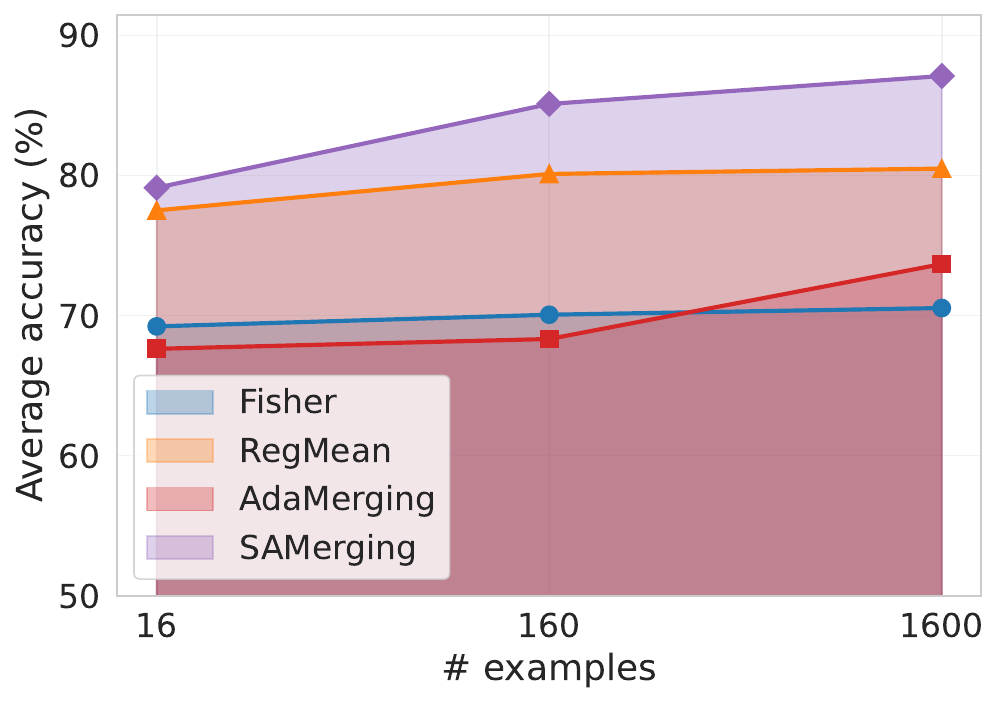}
    \caption{Data-dependent methods gain in performance with increasing number of calibration data on TA-8 using ViT-B/32.}
    \label{fig:data-effect}
\end{figure}
\begin{table}[h!]
    \centering
    \small
    \begin{tabular}{@{}l c@{}}
        \toprule
        Method & Avg. \\
        \midrule
        Fine-tuned (STL) & 82.0 \\
        Simple Average & 56.1 \\
        Task Arithmetic ($\lambda{=}0.5$) & \underline{70.0} \\
        TIES-Merging ($\lambda{=}0.6$) & \underline{70.0} \\
        Fisher Merging & 58.7 \\
        RegMean & 68.8 \\
        AdaMerging & 68.8 \\
        \textbf{\SAMerging} & \textbf{74.9} \\
        \bottomrule
    \end{tabular}
    \caption{Merging methods performance on GLUE using GPT-2. Best result is \textbf{bold} and second best is \underline{underlined}.}
    \label{tab:glue-summary}
\end{table}

\begin{table}[h!]
    \centering
    \small
    \begin{tabular}{@{}l c c c@{}}
        \toprule
        Method & MNLI & IMDb & Avg. \\
        \midrule
        Pre-trained & 33.0 & 50.0 & 41.5 \\
        Finetuned & 91.7 & 97.1 & 94.4 \\
        Task Arithmetic ($\lambda{=}0.5$) & 72.5 & 54.4 & 63.5 \\
        Task Arithmetic ($\lambda{=}1.0$) & \textbf{91.2} & 65.9 & 78.6 \\
        TIES-Merging & 88.7 & \underline{95.9} & 92.3 \\
        AdaMerging ($k{=}1600$) & 72.5 & 54.4 & 63.5 \\
        SAMerging ($k{=}1600$) & \underline{90.5} & \textbf{96.4} & \textbf{93.5} \\
        \bottomrule
    \end{tabular}
    \caption{{ Merging methods performance on MNLI and IMDb using DeBERTa-V2-XXL (1.5B parameters). Best result is \textbf{bold} and second best is \underline{underlined}.}}
    \label{tab:deberta-summary}
\end{table}

\begin{table}[h!]
    \centering
    \small
    \setlength{\tabcolsep}{12pt} 
    \renewcommand{\arraystretch}{1.05}
    \begin{tabular}{@{}l@{}c@{}}
        \toprule
        Variant & \multicolumn{1}{c}{Avg. Acc. (\%)} \\
        \midrule
        \SAMerging & \textbf{85.2} \\
        \addlinespace[1pt]
        \quad \text{--} KL & 84.7\, \drop{0.5\%} \\
        \quad \text{--} SAM & 84.2\, \drop{1.0\%} \\
        \quad \makecell[l]{\text{--} KL \& \text{--} SAM} & 83.5\drop{1.7\%} \\
        \bottomrule
    \end{tabular}
    \caption{Ablation of using $\KL$ as objective and SAM as optimizer on TA-8 using ViT-B/32. Average Acc.\% and drop vs.\ \SAMerging.}
    \label{tab:ablation-kl-sam}
\end{table}
\noindent\textbf{Language tasks.}
As shown in Table~\ref{tab:glue-summary}, \SAMerging achieves the highest average performance on GLUE with GPT-2 fine-tuned models, surpassing TIES-Merging and Task Arithmetic; the same objective transfers beyond vision backbones to autoregressive language models and remains effective under heterogeneous task difficulty. Interestingly, data-dependent baselines (e.g., AdaMerging) underperform data-free ones (e.g., Task Arithmetic, TIES), underscoring the brittleness of entropy minimization.\\
\noindent\textbf{Loss Landscape Geometry.}~ To validate our theoretical motivation, we analyze the geometry of the solution space found by our method. Figure~\ref{fig:eurosat_sun_heatmap} visualizes the 2D loss landscape of the merged model when perturbed along the task vectors of EuroSAT and SUN397 tasks. \SAMerging converges to a significantly broader low-loss basin, indicated by the expansive blue region, compared to the narrower minimum found by AdaMerging, which exhibits sharper transitions to high-loss red areas. This observation is further quantified in Figure~\ref{fig:eurosat_sun_comparison}, which presents 1D loss along these specific task vectors. As shown, the loss curve for SAMerging is visibly flatter around the merged, meaning the cross-entropy loss increases more slowly as parameters deviate towards the other from the optimal point compared to AdaMerging. For additional visualizations of loss landscapes across other task pairs and benchmarks, please refer to Appendix~\ref{apx:vis}.\\
\noindent\textbf{Ablation Study.}~As shown in Table~\ref{tab:ablation-kl-sam}, we do an ablation study to find out the additive gain of performance by changing the optimizer and objective, and the KL objective and SAM each provide measurable gains, and their combination (KL+SAM) yields the strongest improvements. We also conduct ablation experiments to determine the effect of the number of examples on the data-dependent method, as shown in Figure~\ref{fig:data-effect}. The results show that \SAMerging can achieve near SOTA performance with only a handful of data (e.g., $16$ here).

\vspace{-1mm}\section{Conclusion} We propose \SAMerging, a data-efficient, label-free merger that learns layer-wise coefficients by explicitly seeking flat minima. We derive a PAC-Bayes bound for multi-task merging with a heterogeneity term that clarifies when merging succeeds, and cast coefficient learning as multi-teacher distillation, where minimizing student–teacher KL tightens the excess-risk bound of the merged model. Coupling this KL objective with SAM yields a procedure that generalizes well. Empirically, \SAMerging achieves state-of-the-art results on TA-8 and TALL-14/20 with CLIP ViT and GLUE, consistently outperforming data-free and data-dependent baselines, without additional inference parameters or latency.

\paragraph{Limitations and future work} Even with more tasks, regimes with strong task interference or heavy domain shift (e.g., conflicting label spaces or multi-label settings) remain underexplored. The analysis assumes a local NTK-style linearization, so behavior far from this regime is uncertain. Our evaluation focuses solely on classification; extending it to generative tasks is left for future work. Finally, SAM adds calibration-time cost; lighter flatness proxies may reduce this overhead.

\section*{Acknowledgement}

This work was partially supported by   NSF CAREER Award \#2239374.

\bibliography{arxiv/references}

\begin{thebibliography}{89}
\providecommand{\natexlab}[1]{#1}
\providecommand{\url}[1]{\texttt{#1}}
\expandafter\ifx\csname urlstyle\endcsname\relax
  \providecommand{\doi}[1]{doi: #1}\else
  \providecommand{\doi}{doi: \begingroup \urlstyle{rm}\Url}\fi

\bibitem[Ilharco et~al.(2023)Ilharco, Ribeiro, Wortsman, Gururangan, Schmidt, Hajishirzi, and Farhadi]{Ilharco_Ribeiro_Wortsman_Gururangan_Schmidt_Hajishirzi_Farhadi_2023}
Gabriel Ilharco, Marco~Tulio Ribeiro, Mitchell Wortsman, Suchin Gururangan, Ludwig Schmidt, Hannaneh Hajishirzi, and Ali Farhadi.
\newblock Editing models with task arithmetic.
\newblock \penalty0 (arXiv:2212.04089), March 2023.
\newblock \doi{10.48550/arXiv.2212.04089}.
\newblock URL \url{http://arxiv.org/abs/2212.04089}.
\newblock arXiv:2212.04089.

\bibitem[Matena and Raffel(2022)]{Matena_Raffel_2022}
Michael Matena and Colin Raffel.
\newblock Merging models with fisher-weighted averaging.
\newblock In \emph{Proceedings of the 36th International Conference on Neural Information Processing Systems}, NIPS ’22, Red Hook, NY, USA, 2022. Curran Associates Inc.
\newblock ISBN 9781713871088.
\newblock event-place: New Orleans, LA, USA.

\bibitem[Wortsman et~al.(2022)Wortsman, Ilharco, Gadre, Roelofs, Gontijo-Lopes, Morcos, Namkoong, Farhadi, Carmon, Kornblith, and Schmidt]{Wortsman_Ilharco_Gadre_others_2022}
Mitchell Wortsman, Gabriel Ilharco, Samir~Yitzhak Gadre, Rebecca Roelofs, Raphael Gontijo-Lopes, Ari~S. Morcos, Hongseok Namkoong, Ali Farhadi, Yair Carmon, Simon Kornblith, and Ludwig Schmidt.
\newblock Model soups: averaging weights of multiple fine-tuned models improves accuracy without increasing inference time.
\newblock \penalty0 (arXiv:2203.05482), July 2022.
\newblock \doi{10.48550/arXiv.2203.05482}.
\newblock URL \url{http://arxiv.org/abs/2203.05482}.
\newblock arXiv:2203.05482.

\bibitem[Breiman(1996)]{Breiman_1996}
Leo Breiman.
\newblock Bagging predictors.
\newblock \emph{Machine Learning}, 24\penalty0 (2):\penalty0 123–140, August 1996.
\newblock ISSN 1573-0565.
\newblock \doi{10.1007/BF00058655}.
\newblock URL \url{https://doi.org/10.1007/BF00058655}.

\bibitem[Chen and Guestrin(2016)]{Chen_Guestrin_2016}
Tianqi Chen and Carlos Guestrin.
\newblock Xgboost: A scalable tree boosting system.
\newblock \penalty0 (arXiv:1603.02754), June 2016.
\newblock \doi{10.48550/arXiv.1603.02754}.
\newblock URL \url{http://arxiv.org/abs/1603.02754}.
\newblock arXiv:1603.02754.

\bibitem[Ganaie et~al.(2022)Ganaie, Hu, Malik, Tanveer, and Suganthan]{Ganaie_Hu_Malik_Tanveer_Suganthan_2022}
M.~A. Ganaie, Minghui Hu, A.~K. Malik, M.~Tanveer, and P.~N. Suganthan.
\newblock Ensemble deep learning: A review.
\newblock \penalty0 (arXiv:2104.02395), August 2022.
\newblock \doi{10.48550/arXiv.2104.02395}.
\newblock URL \url{http://arxiv.org/abs/2104.02395}.
\newblock arXiv:2104.02395.

\bibitem[Yang et~al.(2024{\natexlab{a}})Yang, Shen, Guo, Wang, Cao, Zhang, and Tao]{Yang_Shen_Guo_Wang_Cao_Zhang_Tao_2024}
Enneng Yang, Li~Shen, Guibing Guo, Xingwei Wang, Xiaochun Cao, Jie Zhang, and Dacheng Tao.
\newblock Model merging in llms, mllms, and beyond: Methods, theories, applications and opportunities.
\newblock \penalty0 (arXiv:2408.07666), September 2024{\natexlab{a}}.
\newblock \doi{10.48550/arXiv.2408.07666}.
\newblock URL \url{http://arxiv.org/abs/2408.07666}.
\newblock arXiv:2408.07666.

\bibitem[Tao et~al.(2025)Tao, Mason, Kulkarni, and Boix]{Tao_Mason_Kulkarni_Boix_2025}
Zhixu~Silvia Tao, Ian Mason, Sanjeev Kulkarni, and Xavier Boix.
\newblock Task arithmetic through the lens of one-shot federated learning.
\newblock \penalty0 (arXiv:2411.18607), July 2025.
\newblock \doi{10.48550/arXiv.2411.18607}.
\newblock URL \url{http://arxiv.org/abs/2411.18607}.
\newblock arXiv:2411.18607.

\bibitem[Liu et~al.(2024{\natexlab{a}})Liu, Liu, Ye, Shen, Li, Jiang, and Li]{Liu_Liu_Ye_Shen_Li_Jiang_Li_2024}
Xiang Liu, Liangxi Liu, Feiyang Ye, Yunheng Shen, Xia Li, Linshan Jiang, and Jialin Li.
\newblock Fedlpa: One-shot federated learning with layer-wise posterior aggregation.
\newblock \penalty0 (arXiv:2310.00339), October 2024{\natexlab{a}}.
\newblock \doi{10.48550/arXiv.2310.00339}.
\newblock URL \url{http://arxiv.org/abs/2310.00339}.
\newblock arXiv:2310.00339.

\bibitem[Chen et~al.(2025)Chen, Zhou, Long, Jiang, and Zhang]{Chen_Zhou_Long_Jiang_Zhang_2025}
Shutong Chen, Tianyi Zhou, Guodong Long, Jing Jiang, and Chengqi Zhang.
\newblock Fedmerge: Federated personalization via model merging.
\newblock \penalty0 (arXiv:2504.06768), April 2025.
\newblock \doi{10.48550/arXiv.2504.06768}.
\newblock URL \url{http://arxiv.org/abs/2504.06768}.
\newblock arXiv:2504.06768.

\bibitem[Salami et~al.(2025)Salami, Buzzega, Mosconi, Bonato, Sabetta, and Calderara]{Salami_Buzzega_Mosconi_Bonato_Sabetta_Calderara_2025}
Riccardo Salami, Pietro Buzzega, Matteo Mosconi, Jacopo Bonato, Luigi Sabetta, and Simone Calderara.
\newblock Closed-form merging of parameter-efficient modules for federated continual learning.
\newblock \penalty0 (arXiv:2410.17961), March 2025.
\newblock \doi{10.48550/arXiv.2410.17961}.
\newblock URL \url{http://arxiv.org/abs/2410.17961}.
\newblock arXiv:2410.17961.

\bibitem[Tsouvalas et~al.(2025)Tsouvalas, Ozcelebi, and Meratnia]{Tsouvalas_Ozcelebi_Meratnia_2025}
Vasileios Tsouvalas, Tanir Ozcelebi, and Nirvana Meratnia.
\newblock Many-task federated fine-tuning via unified task vectors.
\newblock \emph{arXiv preprint arXiv:2502.06376}, 2025.

\bibitem[Ortiz-Jimenez et~al.(2023)Ortiz-Jimenez, Favero, and Frossard]{Ortiz-Jimenez_Favero_Frossard_2023}
Guillermo Ortiz-Jimenez, Alessandro Favero, and Pascal Frossard.
\newblock Task arithmetic in the tangent space: Improved editing of pre-trained models.
\newblock \emph{Advances in Neural Information Processing Systems}, 36:\penalty0 66727–66754, December 2023.
\newblock URL \url{https://proceedings.neurips.cc/paper_files/paper/2023/hash/d28077e5ff52034cd35b4aa15320caea-Abstract-Conference.html}.

\bibitem[Yadav et~al.(2023)Yadav, Tam, Choshen, Raffel, and Bansal]{Yadav_Tam_Choshen_Raffel_Bansal_2023}
Prateek Yadav, Derek Tam, Leshem Choshen, Colin Raffel, and Mohit Bansal.
\newblock Ties-merging: Resolving interference when merging models.
\newblock \penalty0 (arXiv:2306.01708), October 2023.
\newblock \doi{10.48550/arXiv.2306.01708}.
\newblock URL \url{http://arxiv.org/abs/2306.01708}.
\newblock arXiv:2306.01708.

\bibitem[Yang et~al.(2023)Yang, Wang, Shen, Liu, Guo, Wang, and Tao]{Yang_Wang_Shen_Liu_Guo_Wang_Tao_2023}
Enneng Yang, Zhenyi Wang, Li~Shen, Shiwei Liu, Guibing Guo, Xingwei Wang, and Dacheng Tao.
\newblock Adamerging: Adaptive model merging for multi-task learning.
\newblock October 2023.
\newblock URL \url{https://openreview.net/forum?id=nZP6NgD3QY}.

\bibitem[Li et~al.(2025)Li, Zhang, Zhang, Wang, Liu, and Chen]{Li_Zhang_Zhang_Wang_Liu_Chen_2025}
Hongkang Li, Yihua Zhang, Shuai Zhang, Meng Wang, Sijia Liu, and Pin-Yu Chen.
\newblock When is task vector provably effective for model editing? a generalization analysis of nonlinear transformers.
\newblock \penalty0 (arXiv:2504.10957), May 2025.
\newblock \doi{10.48550/arXiv.2504.10957}.
\newblock URL \url{http://arxiv.org/abs/2504.10957}.
\newblock arXiv:2504.10957.

\bibitem[Zhou et~al.(2024)Zhou, Chen, Chen, Zhang, and Yan]{Zhou_Chen_Chen_Zhang_Yan_2024}
Zhanpeng Zhou, Zijun Chen, Yilan Chen, Bo~Zhang, and Junchi Yan.
\newblock On the emergence of cross-task linearity in the pretraining-finetuning paradigm.
\newblock \penalty0 (arXiv:2402.03660), May 2024.
\newblock \doi{10.48550/arXiv.2402.03660}.
\newblock URL \url{http://arxiv.org/abs/2402.03660}.
\newblock arXiv:2402.03660.

\bibitem[Wang and Wang(2024)]{Wang_Wang_2024}
Yibin Wang and Haifeng Wang.
\newblock Distributionally robust unsupervised domain adaptation.
\newblock \emph{Journal of Computational and Applied Mathematics}, 436:\penalty0 115369, January 2024.
\newblock ISSN 0377-0427.
\newblock \doi{10.1016/j.cam.2023.115369}.
\newblock URL \url{https://www.sciencedirect.com/science/article/pii/S0377042723003138}.

\bibitem[Hochreiter and Schmidhuber(1997)]{Hochreiter_Schmidhuber_1997}
Sepp Hochreiter and Jürgen Schmidhuber.
\newblock Flat minima.
\newblock \emph{Neural computation}, 9\penalty0 (1):\penalty0 1–42, 1997.

\bibitem[Neyshabur et~al.(2017)Neyshabur, Bhojanapalli, McAllester, and Srebro]{Neyshabur_Bhojanapalli_McAllester_Srebro_2017}
Behnam Neyshabur, Srinadh Bhojanapalli, David McAllester, and Nati Srebro.
\newblock Exploring generalization in deep learning.
\newblock \emph{Advances in neural information processing systems}, 30, 2017.

\bibitem[Petzka et~al.(2021)Petzka, Kamp, Adilova, Sminchisescu, and Boley]{Petzka_Kamp_Adilova_Sminchisescu_Boley_2021}
Henning Petzka, Michael Kamp, Linara Adilova, Cristian Sminchisescu, and Mario Boley.
\newblock Relative flatness and generalization.
\newblock \emph{Advances in neural information processing systems}, 34:\penalty0 18420–18432, 2021.

\bibitem[Andriushchenko et~al.(2023)Andriushchenko, Croce, Müller, Hein, and Flammarion]{Andriushchenko_Croce_Müller_Hein_Flammarion_2023}
Maksym Andriushchenko, Francesco Croce, Maximilian Müller, Matthias Hein, and Nicolas Flammarion.
\newblock A modern look at the relationship between sharpness and generalization.
\newblock \emph{arXiv preprint arXiv:2302.07011}, 2023.

\bibitem[Yue et~al.(2023)Yue, Jiang, Ye, Gao, Liu, and Zhang]{Yue_Jiang_Ye_Gao_Liu_Zhang_2023}
Yun Yue, Jiadi Jiang, Zhiling Ye, Ning Gao, Yongchao Liu, and Ke~Zhang.
\newblock Sharpness-aware minimization revisited: Weighted sharpness as a regularization term.
\newblock In \emph{Proceedings of the 29th ACM SIGKDD Conference on Knowledge Discovery and Data Mining}, KDD ’23, page 3185–3194, New York, NY, USA, 2023. Association for Computing Machinery.
\newblock ISBN 9798400701030.
\newblock \doi{10.1145/3580305.3599501}.
\newblock URL \url{https://doi.org/10.1145/3580305.3599501}.
\newblock event-place: Long Beach, CA, USA.

\bibitem[Haddouche et~al.(2025)Haddouche, Viallard, Simsekli, and Guedj]{Haddouche_Viallard_Simsekli_Guedj_2025}
Maxime Haddouche, Paul Viallard, Umut Simsekli, and Benjamin Guedj.
\newblock A pac-bayesian link between generalisation and flat minima.
\newblock \penalty0 (arXiv:2402.08508), February 2025.
\newblock \doi{10.48550/arXiv.2402.08508}.
\newblock URL \url{http://arxiv.org/abs/2402.08508}.
\newblock arXiv:2402.08508.

\bibitem[Foret et~al.(2021)Foret, Kleiner, Mobahi, and Neyshabur]{Foret_Kleiner_Mobahi_Neyshabur_2021}
Pierre Foret, Ariel Kleiner, Hossein Mobahi, and Behnam Neyshabur.
\newblock Sharpness-aware minimization for efficiently improving generalization.
\newblock \penalty0 (arXiv:2010.01412), April 2021.
\newblock \doi{10.48550/arXiv.2010.01412}.
\newblock URL \url{http://arxiv.org/abs/2010.01412}.
\newblock arXiv:2010.01412.

\bibitem[Caruana(1997)]{Caruana_1997}
Rich Caruana.
\newblock Multitask learning.
\newblock \emph{Machine Learning}, 28\penalty0 (1):\penalty0 41–75, July 1997.
\newblock ISSN 1573-0565.
\newblock \doi{10.1023/A:1007379606734}.
\newblock URL \url{https://doi.org/10.1023/A:1007379606734}.

\bibitem[Baxter(2000)]{Baxter_2000}
Jonathan Baxter.
\newblock A model of inductive bias learning.
\newblock \emph{J. Artif. Int. Res.}, 12\penalty0 (1):\penalty0 149–198, March 2000.
\newblock ISSN 1076-9757.

\bibitem[Argyriou et~al.(2006)Argyriou, Evgeniou, and Pontil]{Argyriou_Evgeniou_Pontil_2006}
Andreas Argyriou, Theodoros Evgeniou, and Massimiliano Pontil.
\newblock Multi-task feature learning.
\newblock \emph{Advances in neural information processing systems}, 19, 2006.

\bibitem[Maurer et~al.(2016)Maurer, Pontil, and Romera-Paredes]{Maurer_Pontil_Romera-Paredes_2016}
Andreas Maurer, Massimiliano Pontil, and Bernardino Romera-Paredes.
\newblock The benefit of multitask representation learning.
\newblock \emph{Journal of Machine Learning Research}, 17\penalty0 (81):\penalty0 1–32, 2016.

\bibitem[Zhang and Yang(2021)]{Zhang_Yang_2021}
Yu~Zhang and Qiang Yang.
\newblock A survey on multi-task learning.
\newblock \emph{IEEE transactions on knowledge and data engineering}, 34\penalty0 (12):\penalty0 5586–5609, 2021.

\bibitem[Zakerinia et~al.(2025)Zakerinia, Ghobadi, and Lampert]{Zakerinia_Ghobadi_Lampert_2025}
Hossein Zakerinia, Dorsa Ghobadi, and Christoph~H. Lampert.
\newblock From low intrinsic dimensionality to non-vacuous generalization bounds in deep multi-task learning.
\newblock \penalty0 (arXiv:2501.19067), May 2025.
\newblock \doi{10.48550/arXiv.2501.19067}.
\newblock URL \url{http://arxiv.org/abs/2501.19067}.
\newblock arXiv:2501.19067.

\bibitem[Zakerinia and Lampert(2025)]{Zakerinia_Lampert_2025}
Hossein Zakerinia and Christoph~H. Lampert.
\newblock Fast rate bounds for multi-task and meta-learning with different sample sizes.
\newblock \penalty0 (arXiv:2505.15496), May 2025.
\newblock \doi{10.48550/arXiv.2505.15496}.
\newblock URL \url{http://arxiv.org/abs/2505.15496}.
\newblock arXiv:2505.15496.

\bibitem[Dai and Zhu(2020)]{Dai_Zhu_2020}
Xiaowu Dai and Yuhua Zhu.
\newblock On large batch training and sharp minima: a fokker–planck perspective.
\newblock \emph{Journal of Statistical Theory and Practice}, 14\penalty0 (3):\penalty0 53, 2020.

\bibitem[Dinh et~al.(2017)Dinh, Pascanu, Bengio, and Bengio]{Dinh_Pascanu_Bengio_Bengio_2017}
Laurent Dinh, Razvan Pascanu, Samy Bengio, and Yoshua Bengio.
\newblock Sharp minima can generalize for deep nets.
\newblock In \emph{International Conference on Machine Learning}, page 1019–1028. PMLR, 2017.

\bibitem[Lee et~al.(2025)Lee, Jung, and Baik]{Lee_Jung_Baik_2025}
Yeoreum Lee, Jinwook Jung, and Sungyong Baik.
\newblock Mitigating parameter interference in model merging via sharpness-aware fine-tuning.
\newblock \emph{arXiv preprint arXiv:2504.14662}, 2025.

\bibitem[Zhang et~al.(2025)Zhang, Theus, Teney, Orvieto, Pang, and Mauw]{Zhang_Theus_Teney_Orvieto_Pang_Mauw_2025}
Chenxiang Zhang, Alexander Theus, Damien Teney, Antonio Orvieto, Jun Pang, and Sjouke Mauw.
\newblock How does the optimizer implicitly bias the model merging loss landscape?
\newblock \penalty0 (arXiv:2510.04686), October 2025.
\newblock \doi{10.48550/arXiv.2510.04686}.
\newblock URL \url{http://arxiv.org/abs/2510.04686}.
\newblock arXiv:2510.04686.

\bibitem[Wu et~al.(2023)Wu, Wang, Ge, Lu, Zhou, Shan, and Luo]{Wu_Wang_Ge_Lu_Zhou_Shan_Luo_2023}
Chengyue Wu, Teng Wang, Yixiao Ge, Zeyu Lu, Ruisong Zhou, Ying Shan, and Ping Luo.
\newblock $π$-tuning: Transferring multimodal foundation models with optimal multi-task interpolation.
\newblock \penalty0 (arXiv:2304.14381), May 2023.
\newblock \doi{10.48550/arXiv.2304.14381}.
\newblock URL \url{http://arxiv.org/abs/2304.14381}.
\newblock arXiv:2304.14381.

\bibitem[Misra et~al.(2016)Misra, Shrivastava, Gupta, and Hebert]{Misra_Shrivastava_Gupta_Hebert_2016}
Ishan Misra, Abhinav Shrivastava, Abhinav Gupta, and Martial Hebert.
\newblock Cross-stitch networks for multi-task learning.
\newblock \penalty0 (arXiv:1604.03539), April 2016.
\newblock \doi{10.48550/arXiv.1604.03539}.
\newblock URL \url{http://arxiv.org/abs/1604.03539}.
\newblock arXiv:1604.03539.

\bibitem[Sun et~al.(2020)Sun, Panda, Feris, and Saenko]{Sun_Panda_Feris_Saenko_2020}
Ximeng Sun, Rameswar Panda, Rogerio Feris, and Kate Saenko.
\newblock Adashare: Learning what to share for efficient deep multi-task learning.
\newblock \penalty0 (arXiv:1911.12423), November 2020.
\newblock \doi{10.48550/arXiv.1911.12423}.
\newblock URL \url{http://arxiv.org/abs/1911.12423}.
\newblock arXiv:1911.12423.

\bibitem[Hazimeh et~al.(2021)Hazimeh, Zhao, Chowdhery, Sathiamoorthy, Chen, Mazumder, Hong, and Chi]{Hazimeh_Zhao_Chowdhery_Sathiamoorthy_Chen_Mazumder_Hong_Chi_2021}
Hussein Hazimeh, Zhe Zhao, Aakanksha Chowdhery, Maheswaran Sathiamoorthy, Yihua Chen, Rahul Mazumder, Lichan Hong, and Ed~Chi.
\newblock Dselect-k: Differentiable selection in the mixture of experts with applications to multi-task learning.
\newblock In \emph{Advances in Neural Information Processing Systems}, volume~34, page 29335–29347. Curran Associates, Inc., 2021.
\newblock URL \url{https://proceedings.neurips.cc/paper/2021/hash/f5ac21cd0ef1b88e9848571aeb53551a-Abstract.html}.

\bibitem[Tang et~al.(2020)Tang, Liu, Zhao, and Gong]{Tang_Liu_Zhao_Gong_2020}
Hongyan Tang, Junning Liu, Ming Zhao, and Xudong Gong.
\newblock Progressive layered extraction (ple): A novel multi-task learning (mtl) model for personalized recommendations.
\newblock In \emph{Fourteenth ACM Conference on Recommender Systems}, page 269–278, Virtual Event Brazil, September 2020. ACM.
\newblock ISBN 9781450375832.
\newblock \doi{10.1145/3383313.3412236}.
\newblock URL \url{https://dl.acm.org/doi/10.1145/3383313.3412236}.

\bibitem[Yu et~al.(2020)Yu, Kumar, Gupta, Levine, Hausman, and Finn]{Yu_Kumar_Gupta_Levine_Hausman_Finn_2020}
Tianhe Yu, Saurabh Kumar, Abhishek Gupta, Sergey Levine, Karol Hausman, and Chelsea Finn.
\newblock Gradient surgery for multi-task learning.
\newblock \penalty0 (arXiv:2001.06782), December 2020.
\newblock \doi{10.48550/arXiv.2001.06782}.
\newblock URL \url{http://arxiv.org/abs/2001.06782}.
\newblock arXiv:2001.06782.

\bibitem[Liu et~al.(2024{\natexlab{b}})Liu, Liu, Jin, Stone, and Liu]{Liu_Liu_Jin_Stone_Liu_2024}
Bo~Liu, Xingchao Liu, Xiaojie Jin, Peter Stone, and Qiang Liu.
\newblock Conflict-averse gradient descent for multi-task learning.
\newblock \penalty0 (arXiv:2110.14048), February 2024{\natexlab{b}}.
\newblock \doi{10.48550/arXiv.2110.14048}.
\newblock URL \url{http://arxiv.org/abs/2110.14048}.
\newblock arXiv:2110.14048.

\bibitem[Quinton and Rey(2025)]{Quinton_Rey_2025}
Pierre Quinton and Valérian Rey.
\newblock Jacobian descent for multi-objective optimization.
\newblock \penalty0 (arXiv:2406.16232), February 2025.
\newblock \doi{10.48550/arXiv.2406.16232}.
\newblock URL \url{http://arxiv.org/abs/2406.16232}.
\newblock arXiv:2406.16232.

\bibitem[Kendall et~al.(2018)Kendall, Gal, and Cipolla]{Kendall_Gal_Cipolla_2018}
Alex Kendall, Yarin Gal, and Roberto Cipolla.
\newblock Multi-task learning using uncertainty to weigh losses for scene geometry and semantics.
\newblock \penalty0 (arXiv:1705.07115), April 2018.
\newblock \doi{10.48550/arXiv.1705.07115}.
\newblock URL \url{http://arxiv.org/abs/1705.07115}.
\newblock arXiv:1705.07115.

\bibitem[Chen et~al.(2018)Chen, Badrinarayanan, Lee, and Rabinovich]{Chen_Badrinarayanan_Lee_Rabinovich_2018}
Zhao Chen, Vijay Badrinarayanan, Chen-Yu Lee, and Andrew Rabinovich.
\newblock Gradnorm: Gradient normalization for adaptive loss balancing in deep multitask networks.
\newblock \penalty0 (arXiv:1711.02257), June 2018.
\newblock \doi{10.48550/arXiv.1711.02257}.
\newblock URL \url{http://arxiv.org/abs/1711.02257}.
\newblock arXiv:1711.02257.

\bibitem[Lin et~al.(2019)Lin, Zhen, Li, Zhang, and Kwong]{Lin_Zhen_Li_Zhang_Kwong_2019}
Xi~Lin, Hui-Ling Zhen, Zhenhua Li, Qing-Fu Zhang, and Sam Kwong.
\newblock Pareto multi-task learning.
\newblock In \emph{Advances in Neural Information Processing Systems}, volume~32. Curran Associates, Inc., 2019.
\newblock URL \url{https://papers.nips.cc/paper_files/paper/2019/hash/685bfde03eb646c27ed565881917c71c-Abstract.html}.

\bibitem[Shamsian et~al.(2023)Shamsian, Navon, Glazer, Kawaguchi, Chechik, and Fetaya]{Shamsian_Navon_Glazer_Kawaguchi_Chechik_Fetaya_2023}
Aviv Shamsian, Aviv Navon, Neta Glazer, Kenji Kawaguchi, Gal Chechik, and Ethan Fetaya.
\newblock Auxiliary learning as an asymmetric bargaining game.
\newblock \penalty0 (arXiv:2301.13501), June 2023.
\newblock \doi{10.48550/arXiv.2301.13501}.
\newblock URL \url{http://arxiv.org/abs/2301.13501}.
\newblock arXiv:2301.13501.

\bibitem[Yu et~al.(2024)Yu, Yu, Yu, Huang, and Li]{Yu_Yu_Yu_Huang_Li_2024}
Le~Yu, Bowen Yu, Haiyang Yu, Fei Huang, and Yongbin Li.
\newblock Language models are super mario: Absorbing abilities from homologous models as a free lunch.
\newblock \penalty0 (arXiv:2311.03099), June 2024.
\newblock \doi{10.48550/arXiv.2311.03099}.
\newblock URL \url{http://arxiv.org/abs/2311.03099}.
\newblock arXiv:2311.03099.

\bibitem[Du et~al.(2024)Du, Lee, Li, Jiang, Guo, Yu, Liu, Goh, Tang, He, and Zhang]{Du_Lee_Li_Jiang_Guo_Yu_Liu_Goh_Tang_He}
Guodong Du, Junlin Lee, Jing Li, Runhua Jiang, Yifei Guo, Shuyang Yu, Hanting Liu, Sim~Kuan Goh, Ho-Kin Tang, Daojing He, and Min Zhang.
\newblock Parameter competition balancing for model merging.
\newblock \penalty0 (arXiv:2410.02396), October 2024.
\newblock \doi{10.48550/arXiv.2410.02396}.
\newblock URL \url{http://arxiv.org/abs/2410.02396}.
\newblock arXiv:2410.02396.

\bibitem[Marczak et~al.(2025)Marczak, Magistri, Cygert, Twardowski, Bagdanov, and Weijer]{Marczak_Magistri_Cygert_Twardowski_Bagdanov_Weijer_2025}
Daniel Marczak, Simone Magistri, Sebastian Cygert, Bartłomiej Twardowski, Andrew~D. Bagdanov, and Joost van~de Weijer.
\newblock No task left behind: Isotropic model merging with common and task-specific subspaces.
\newblock \penalty0 (arXiv:2502.04959), June 2025.
\newblock \doi{10.48550/arXiv.2502.04959}.
\newblock URL \url{http://arxiv.org/abs/2502.04959}.
\newblock arXiv:2502.04959.

\bibitem[Jin et~al.(2025)Jin, Ren, Preotiuc-Pietro, and Cheng]{Jin_Ren_Preotiuc-Pietro_Cheng_2025}
Xisen Jin, Xiang Ren, Daniel Preotiuc-Pietro, and Pengxiang Cheng.
\newblock Dataless knowledge fusion by merging weights of language models.
\newblock \penalty0 (arXiv:2212.09849), May 2025.
\newblock \doi{10.48550/arXiv.2212.09849}.
\newblock URL \url{http://arxiv.org/abs/2212.09849}.
\newblock arXiv:2212.09849.

\bibitem[Nguyen et~al.(2025)Nguyen, Huu-Tien, Suzuki, and Nguyen]{Nguyen_Huu-Tien_Suzuki_Nguyen_2025}
The-Hai Nguyen, Dang Huu-Tien, Takeshi Suzuki, and Le-Minh Nguyen.
\newblock Regmean++: Enhancing effectiveness and generalization of regression mean for model merging.
\newblock \penalty0 (arXiv:2508.03121), August 2025.
\newblock \doi{10.48550/arXiv.2508.03121}.
\newblock URL \url{http://arxiv.org/abs/2508.03121}.
\newblock arXiv:2508.03121.

\bibitem[Yang et~al.(2024{\natexlab{b}})Yang, Shen, Wang, Guo, Chen, Wang, and Tao]{Yang_Shen_Wang_Guo_Chen_Wang_Tao_2024}
Enneng Yang, Li~Shen, Zhenyi Wang, Guibing Guo, Xiaojun Chen, Xingwei Wang, and Dacheng Tao.
\newblock Representation surgery for multi-task model merging.
\newblock \penalty0 (arXiv:2402.02705), May 2024{\natexlab{b}}.
\newblock \doi{10.48550/arXiv.2402.02705}.
\newblock URL \url{http://arxiv.org/abs/2402.02705}.
\newblock arXiv:2402.02705.

\bibitem[Ainsworth et~al.(2023)Ainsworth, Hayase, and Srinivasa]{Ainsworth_Hayase_Srinivasa_2023}
Samuel~K. Ainsworth, Jonathan Hayase, and Siddhartha Srinivasa.
\newblock Git re-basin: Merging models modulo permutation symmetries.
\newblock \penalty0 (arXiv:2209.04836), March 2023.
\newblock \doi{10.48550/arXiv.2209.04836}.
\newblock URL \url{http://arxiv.org/abs/2209.04836}.
\newblock arXiv:2209.04836.

\bibitem[Yoshida et~al.(2024)Yoshida, Naraki, Horie, Yamaki, Shimizu, Saito, McAuley, and Naganuma]{Yoshida_Naraki_Horie_Yamaki_Shimizu_Saito_McAuley_Naganuma_2024}
Kotaro Yoshida, Yuji Naraki, Takafumi Horie, Ryosuke Yamaki, Ryotaro Shimizu, Yuki Saito, Julian McAuley, and Hiroki Naganuma.
\newblock Mastering task arithmetic: $\tau$jp as a key indicator for weight disentanglement.
\newblock October 2024.
\newblock URL \url{https://openreview.net/forum?id=1VwWi6zbxs}.

\bibitem[Hinton et~al.(2015)Hinton, Vinyals, and Dean]{Hinton_Vinyals_Dean_2015}
Geoffrey Hinton, Oriol Vinyals, and Jeff Dean.
\newblock Distilling the knowledge in a neural network.
\newblock \penalty0 (arXiv:1503.02531), March 2015.
\newblock \doi{10.48550/arXiv.1503.02531}.
\newblock URL \url{http://arxiv.org/abs/1503.02531}.
\newblock arXiv:1503.02531.

\bibitem[Xu et~al.(2025)Xu, Li, and Zhang]{Xu_Li_Zhang_2025}
Jing Xu, Jiazheng Li, and Jingzhao Zhang.
\newblock Scalable model merging with progressive layer-wise distillation.
\newblock \penalty0 (arXiv:2502.12706), May 2025.
\newblock \doi{10.48550/arXiv.2502.12706}.
\newblock URL \url{http://arxiv.org/abs/2502.12706}.
\newblock arXiv:2502.12706.

\bibitem[Dziugaite and Roy(2017)]{Dziugaite_Roy_2017}
Gintare~Karolina Dziugaite and Daniel~M. Roy.
\newblock Computing nonvacuous generalization bounds for deep (stochastic) neural networks with many more parameters than training data.
\newblock \penalty0 (arXiv:1703.11008), October 2017.
\newblock \doi{10.48550/arXiv.1703.11008}.
\newblock URL \url{http://arxiv.org/abs/1703.11008}.
\newblock arXiv:1703.11008.

\bibitem[Baek et~al.(2024)Baek, Kolter, and Raghunathan]{Baek_Kolter_Raghunathan_2024}
Christina Baek, Zico Kolter, and Aditi Raghunathan.
\newblock Why is sam robust to label noise?
\newblock \penalty0 (arXiv:2405.03676), May 2024.
\newblock \doi{10.48550/arXiv.2405.03676}.
\newblock URL \url{http://arxiv.org/abs/2405.03676}.
\newblock arXiv:2405.03676.

\bibitem[Na et~al.(2022)Na, Mehta, and Strubell]{Na_Mehta_Strubell_2022}
Clara Na, Sanket~Vaibhav Mehta, and Emma Strubell.
\newblock Train flat, then compress: Sharpness-aware minimization learns more compressible models.
\newblock In Yoav Goldberg, Zornitsa Kozareva, and Yue Zhang, editors, \emph{Findings of the Association for Computational Linguistics: EMNLP 2022}, page 4909–4936, Abu Dhabi, United Arab Emirates, December 2022. Association for Computational Linguistics.
\newblock \doi{10.18653/v1/2022.findings-emnlp.361}.
\newblock URL \url{https://aclanthology.org/2022.findings-emnlp.361/}.

\bibitem[Jacot et~al.(2020)Jacot, Gabriel, and Hongler]{Jacot_Gabriel_Hongler_2020}
Arthur Jacot, Franck Gabriel, and Clément Hongler.
\newblock Neural tangent kernel: Convergence and generalization in neural networks.
\newblock \penalty0 (arXiv:1806.07572), February 2020.
\newblock \doi{10.48550/arXiv.1806.07572}.
\newblock URL \url{http://arxiv.org/abs/1806.07572}.
\newblock arXiv:1806.07572.

\bibitem[Wang et~al.(2024)Wang, Dimitriadis, Ortiz-Jimenez, Fleuret, and Frossard]{Wang_Dimitriadis_Ortiz-Jimenez_Fleuret_Frossard_2024}
Ke~Wang, Nikolaos Dimitriadis, Guillermo Ortiz-Jimenez, François Fleuret, and Pascal Frossard.
\newblock Localizing task information for improved model merging and compression.
\newblock \penalty0 (arXiv:2405.07813), May 2024.
\newblock \doi{10.48550/arXiv.2405.07813}.
\newblock URL \url{http://arxiv.org/abs/2405.07813}.
\newblock arXiv:2405.07813.

\bibitem[Malrieu and Talay(2006)]{Malrieu_Talay_2006}
Florent Malrieu and Denis Talay.
\newblock Concentration inequalities for euler schemes.
\newblock In Harald Niederreiter and Denis Talay, editors, \emph{Monte Carlo and Quasi-Monte Carlo Methods 2004}, page 355–371, Berlin, Heidelberg, 2006. Springer.
\newblock ISBN 9783540311867.
\newblock \doi{10.1007/3-540-31186-6_21}.

\bibitem[Csiszár and Körner(2015)]{Csiszár_Körner_2015}
Imre Csiszár and János Körner.
\newblock \emph{Information Theory: Coding Theorems for Discrete Memoryless Systems}.
\newblock Cambridge University Press, USA, 2nd edition, 2015.
\newblock ISBN 1107565049.

\bibitem[Krause et~al.(2013)Krause, Stark, Deng, and Fei-Fei]{Krause_Stark_Deng_Fei-Fei_2013}
Jonathan Krause, Michael Stark, Jia Deng, and Li~Fei-Fei.
\newblock 3d object representations for fine-grained categorization.
\newblock In \emph{Proceedings of the IEEE international conference on computer vision workshops}, page 554–561, 2013.

\bibitem[Cimpoi et~al.(2014)Cimpoi, Maji, Kokkinos, Mohamed, and Vedaldi]{Cimpoi_Maji_Kokkinos_Mohamed_Vedaldi_2014}
Mircea Cimpoi, Subhransu Maji, Iasonas Kokkinos, Sammy Mohamed, and Andrea Vedaldi.
\newblock Describing textures in the wild.
\newblock In \emph{Proceedings of the IEEE conference on computer vision and pattern recognition}, page 3606–3613, 2014.

\bibitem[Helber et~al.(2019)Helber, Bischke, Dengel, and Borth]{Helber_Bischke_Dengel_Borth_2019}
Patrick Helber, Benjamin Bischke, Andreas Dengel, and Damian Borth.
\newblock Eurosat: A novel dataset and deep learning benchmark for land use and land cover classification.
\newblock \emph{IEEE Journal of Selected Topics in Applied Earth Observations and Remote Sensing}, 12\penalty0 (7):\penalty0 2217–2226, 2019.

\bibitem[Stallkamp et~al.(2011)Stallkamp, Schlipsing, Salmen, and Igel]{Stallkamp_Schlipsing_Salmen_Igel_2011}
Johannes Stallkamp, Marc Schlipsing, Jan Salmen, and Christian Igel.
\newblock The german traffic sign recognition benchmark: a multi-class classification competition.
\newblock In \emph{The 2011 international joint conference on neural networks}, page 1453–1460. IEEE, 2011.

\bibitem[LeCun(1998)]{LeCun_1998}
Yann LeCun.
\newblock The mnist database of handwritten digits.
\newblock \emph{http://yann. lecun. com/exdb/mnist/}, 1998.

\bibitem[Cheng et~al.(2017)Cheng, Han, and Lu]{Cheng_Han_Lu_2017}
Gong Cheng, Junwei Han, and Xiaoqiang Lu.
\newblock Remote sensing image scene classification: Benchmark and state of the art.
\newblock \emph{Proceedings of the IEEE}, 105\penalty0 (10):\penalty0 1865–1883, 2017.

\bibitem[Xiao et~al.(2016)Xiao, Ehinger, Hays, Torralba, and Oliva]{Xiao_Ehinger_Hays_Torralba_Oliva_2016}
Jianxiong Xiao, Krista~A Ehinger, James Hays, Antonio Torralba, and Aude Oliva.
\newblock Sun database: Exploring a large collection of scene categories.
\newblock \emph{International Journal of Computer Vision}, 119\penalty0 (1):\penalty0 3–22, 2016.

\bibitem[Netzer et~al.(2011)Netzer, Wang, Coates, Bissacco, Wu, Ng, et~al.]{Netzer_Wang_Coates_Bissacco_Wu_Ng_others_2011}
Yuval Netzer, Tao Wang, Adam Coates, Alessandro Bissacco, Baolin Wu, Andrew~Y Ng, et~al.
\newblock Reading digits in natural images with unsupervised feature learning.
\newblock In \emph{NIPS workshop on deep learning and unsupervised feature learning}, volume 2011, page~7. Granada, 2011.

\bibitem[Nilsback and Zisserman(2008)]{Nilsback_Zisserman_2008}
Maria-Elena Nilsback and Andrew Zisserman.
\newblock Automated flower classification over a large number of classes.
\newblock In \emph{2008 Sixth Indian conference on computer vision, graphics \& image processing}, page 722–729. IEEE, 2008.

\bibitem[Krizhevsky et~al.(2009)Krizhevsky, Hinton, et~al.]{Krizhevsky_Hinton_others_2009}
Alex Krizhevsky, Geoffrey Hinton, et~al.
\newblock Learning multiple layers of features from tiny images.
\newblock 2009.

\bibitem[Veeling et~al.(2018)Veeling, Linmans, Winkens, Cohen, and Welling]{Veeling_Linmans_Winkens_Cohen_Welling_2018}
Bastiaan~S Veeling, Jasper Linmans, Jim Winkens, Taco Cohen, and Max Welling.
\newblock Rotation equivariant cnns for digital pathology.
\newblock In \emph{International Conference on Medical image computing and computer-assisted intervention}, page 210–218. Springer, 2018.

\bibitem[Coates et~al.(2011)Coates, Ng, and Lee]{Coates_Ng_Lee_2011}
Adam Coates, Andrew Ng, and Honglak Lee.
\newblock An analysis of single-layer networks in unsupervised feature learning.
\newblock In \emph{Proceedings of the fourteenth international conference on artificial intelligence and statistics}, page 215–223. JMLR Workshop and Conference Proceedings, 2011.

\bibitem[Parkhi et~al.(2012)Parkhi, Vedaldi, Zisserman, and Jawahar]{Parkhi_Vedaldi_Zisserman_Jawahar_2012}
Omkar~M Parkhi, Andrea Vedaldi, Andrew Zisserman, and CV~Jawahar.
\newblock Cats and dogs.
\newblock In \emph{2012 IEEE conference on computer vision and pattern recognition}, page 3498–3505. IEEE, 2012.

\bibitem[Goodfellow et~al.(2013)Goodfellow, Erhan, Carrier, Courville, Mirza, Hamner, Cukierski, Tang, Thaler, Lee, et~al.]{Goodfellow_Erhan_Carrier_Courville_Mirza_Hamner_Cukierski_Tang_Thaler_Lee}
Ian~J Goodfellow, Dumitru Erhan, Pierre~Luc Carrier, Aaron Courville, Mehdi Mirza, Ben Hamner, Will Cukierski, Yichuan Tang, David Thaler, Dong-Hyun Lee, et~al.
\newblock Challenges in representation learning: A report on three machine learning contests.
\newblock In \emph{International conference on neural information processing}, page 117–124. Springer, 2013.

\bibitem[Cohen et~al.(2017)Cohen, Afshar, Tapson, and Van~Schaik]{Cohen_Afshar_Tapson_VanSchaik_2017}
Gregory Cohen, Saeed Afshar, Jonathan Tapson, and Andre Van~Schaik.
\newblock Emnist: Extending mnist to handwritten letters.
\newblock In \emph{2017 international joint conference on neural networks (IJCNN)}, page 2921–2926. IEEE, 2017.

\bibitem[Bossard et~al.(2014)Bossard, Guillaumin, and Van~Gool]{Bossard_Guillaumin_VanGool_2014}
Lukas Bossard, Matthieu Guillaumin, and Luc Van~Gool.
\newblock Food-101–mining discriminative components with random forests.
\newblock In \emph{European conference on computer vision}, page 446–461. Springer, 2014.

\bibitem[Clanuwat et~al.(2018)Clanuwat, Bober-Irizar, Kitamoto, Lamb, Yamamoto, and Ha]{Clanuwat_Bober-Irizar_Kitamoto_Lamb_Yamamoto_Ha_2018}
Tarin Clanuwat, Mikel Bober-Irizar, Asanobu Kitamoto, Alex Lamb, Kazuaki Yamamoto, and David Ha.
\newblock Deep learning for classical japanese literature.
\newblock \emph{arXiv preprint arXiv:1812.01718}, 2018.

\bibitem[Socher et~al.(2013)Socher, Perelygin, Wu, Chuang, Manning, Ng, and Potts]{Socher_Perelygin_Wu_Chuang_Manning_Ng_Potts_2013}
Richard Socher, Alex Perelygin, Jean Wu, Jason Chuang, Christopher~D Manning, Andrew~Y Ng, and Christopher Potts.
\newblock Recursive deep models for semantic compositionality over a sentiment treebank.
\newblock In \emph{Proceedings of the 2013 conference on empirical methods in natural language processing}, page 1631–1642, 2013.

\bibitem[Wang et~al.(2019)Wang, Singh, Michael, Hill, Levy, and Bowman]{Wang_Singh_Michael_Hill_Levy_Bowman_2019}
Alex Wang, Amanpreet Singh, Julian Michael, Felix Hill, Omer Levy, and Samuel~R. Bowman.
\newblock Glue: A multi-task benchmark and analysis platform for natural language understanding.
\newblock \penalty0 (arXiv:1804.07461), February 2019.
\newblock \doi{10.48550/arXiv.1804.07461}.
\newblock URL \url{http://arxiv.org/abs/1804.07461}.
\newblock arXiv:1804.07461.

\bibitem[Warstadt et~al.(2019)Warstadt, Singh, and Bowman]{Warstadt_Singh_Bowman_2019}
Alex Warstadt, Amanpreet Singh, and Samuel~R Bowman.
\newblock Neural network acceptability judgments.
\newblock \emph{Transactions of the Association for Computational Linguistics}, 7:\penalty0 625–641, 2019.

\bibitem[Dolan and Brockett(2005)]{Dolan_Brockett_2005}
Bill Dolan and Chris Brockett.
\newblock Automatically constructing a corpus of sentential paraphrases.
\newblock In \emph{Third international workshop on paraphrasing (IWP2005)}, 2005.

\bibitem[Williams et~al.(2017)Williams, Nangia, and Bowman]{Williams_Nangia_Bowman_2017}
Adina Williams, Nikita Nangia, and Samuel~R Bowman.
\newblock A broad-coverage challenge corpus for sentence understanding through inference.
\newblock \emph{arXiv preprint arXiv:1704.05426}, 2017.

\bibitem[Rajpurkar et~al.(2016)Rajpurkar, Zhang, Lopyrev, and Liang]{Rajpurkar_Zhang_Lopyrev_Liang_2016}
Pranav Rajpurkar, Jian Zhang, Konstantin Lopyrev, and Percy Liang.
\newblock Squad: 100,000+ questions for machine comprehension of text.
\newblock \emph{arXiv preprint arXiv:1606.05250}, 2016.

\bibitem[Tang et~al.(2024)Tang, Shen, Luo, Hu, Du, and Tao]{tang2024fusionbench}
Anke Tang, Li~Shen, Yong Luo, Han Hu, Bo~Du, and Dacheng Tao.
\newblock Fusionbench: A comprehensive benchmark of deep model fusion.
\newblock \emph{arXiv preprint arXiv:2406.03280}, 2024.

\end{thebibliography}
\bibliographystyle{unsrtnat}

\clearpage
\appendix
\section{Theoretical Definition}
\label{apx:def}
\begin{definition}[Poincar\'e inequality, \citet{Malrieu_Talay_2006, Haddouche_Viallard_Simsekli_Guedj_2025}]
\label{def:poincare}
A distribution $Q$ satisfies a Poincar\'e inequality with constant $c_P(Q)$ if for all sets of functions $f$ that are square-integrable, with their gradient’s norm also being square-integrable, we have
\begin{equation}
\Var_{h \sim Q}(f(h)) \le c_P(Q) \E_{h \sim Q}[\|\nabla_h f(h)\|^2],
\end{equation}
where $\mathrm{Var}_{h \sim Q}(f(h)) = \E_{h \sim Q}[f(h) - \E_{h \sim Q}[f(h)]]^2$ is the variance of $f$ with respect to $Q$.
We then say that $Q$ is Poincar\'e with constant $c_P(Q)$. A gaussian distribution $Q = \mathcal{N}(\mu, \Sigma)$ is Poincar\'e with constant $c_P(Q) = \|\Sigma\|$.
\end{definition}
\begin{definition}[Quadratically Self-Bounded, \citet{Haddouche_Viallard_Simsekli_Guedj_2025}]
\label{def:qsb}
    We say that $Q \in \mathcal{M}(\mathcal{H})$ is quadratically self-bounded 
with respect to the loss function $\ell \to \mathbb{R}_+$ 
and a constant $C > 0$ (namely $\mathrm{QSB}(\ell, C)$) if
\[
\E_{z\sim \mathcal{D}}
\left[
   \Big( \E_{h \sim Q}\,\ell(h, z) \Big)^2
\right]
\;\;\le\;\;
C \mathcal{L}_{\mathcal{D}}(Q)
\;=\;
C \; \E_{z\sim \mathcal{D}}
   \left[ \E_{h \sim Q}\,\ell(h, z) \right].
\]
Note that this is a relaxation of boundedness as if a loss is bounded $[0, C]$, the distribution $Q$ is $\mathrm{QSB}(\ell, C)$.
\end{definition}
\section{Lemmas and Theorems}
\label{apx:proofs}
\subsection{Proof of Lemma~\ref{lem:linear-in-q}}
\begin{proof}
\label{proof:linear-in-q}
We have $\mathcal L_{\mathcal D_t}(Q)=\int \mathcal{L}_{\mathcal D_t}(h)\,\mathrm dQ(h)$ and, by linearity of the integral, $\mathcal L_{\mathcal D_t}(Q_{\mathrm{merge}}) = \int \mathcal{L}_{\mathcal D_t}(h)\,\mathrm d\big(\sum_{j=1}^T\beta_j Q_j(h)\big)=\sum_{j=1}^T\beta_j\int \mathcal{L}_{\mathcal D_t}(h)\,\mathrm dQ_j(h)=\sum_{j=1}^T \beta_j\mathcal L_{\mathcal D_t}(Q_j)$.
\end{proof}
\subsection{Proof of Proposition~\ref{prop:decomposition}}
\begin{proof}
\label{proof:decomposition}
Using Lemma~\ref{lem:linear-in-q}, we have
\begin{align*}
\mathcal L_{\mathrm{mix}}(Q_{\mathrm{merge}}) &=\sum_{i=1}^T \alpha_i\,\mathcal L_{\mathcal D_i}(Q_{\mathrm{merge}}) =\sum_{i=1}^T\sum_{j=1}^T \alpha_i\beta_j\,\mathcal L_{\mathcal D_i}(Q_j).
\end{align*}
Then by adding and subtracting $\sum_{j=1}^T \beta_j\,\mathcal L_{\mathcal D_j}(Q_j)$ and rearranging, we have
\begin{align*}
\mathcal L_{\mathrm{mix}}(Q_{\mathrm{merge}}) &=\sum_{i=1}^T\sum_{j=1}^T \alpha_i\beta_j\,\mathcal L_{\mathcal D_i}(Q_j) +\Big[\sum_{j=1}^T \beta_j\,\mathcal L_{\mathcal D_j}(Q_j) - \sum_{j=1}^T \beta_j\,\mathcal L_{\mathcal D_j}(Q_j) \Big] \\&= \sum_{j=1}^T \beta_j\,\mathcal L_{\mathcal D_j}(Q_j) +\sum_{i=1}^T\sum_{j=1}^T \alpha_i\beta_j\,\big(\mathcal L_{\mathcal D_i}(Q_j)-\mathcal L_{\mathcal D_j}(Q_j)\big).
\end{align*}
\end{proof}

\begin{theorem}[Theorem 6, \citet{Haddouche_Viallard_Simsekli_Guedj_2025}]
\label{thm:thm6-h}
Let $C>0$, $\lambda\in\left(0,\frac{2}{C}\right)$, a prior $P$ over $\Theta$, a distribution $\mathcal{D}$, and a sample $\mathcal{S}_m\sim \mathcal{D}^m$. If a posterior $Q$ is Poincae\'e with constant $c_P(Q)$ (Definition~\ref{def:poincare}) and $\ell$ is $\mathrm{QSB}(\ell, C)$ under $Q$ (Definition~\ref{def:qsb}), then with probability at least $1-\delta$ over $\mathcal{S}_m$,
\begin{equation*}
\mathcal{L}_{\mathcal{D}}(Q)\;\le\;
\frac{1}{1-\frac{\lambda C}{2}}\left(
\widehat{\mathcal{L}}_{S_m}(Q)\;+\;\frac{\KL\!\left(Q\Vert P\right)+\log\!\left(\frac{1}{\delta}\right)}{\lambda m}
\right)
\;+\;\frac{\lambda}{2-\lambda C}\;c_P(Q)\;\mathbb{E}_{z\sim D}\,\mathbb{E}_{h\sim Q}\left\|\nabla_h \ell\!\left(h,z\right)\right\|_2^2.
\end{equation*}
\end{theorem}

\subsection{Proof of Theorem~\ref{thm:mtl-pb}}

\begin{proof}
    \label{proof:mtl-pb}
    We start from Proposition~\ref{prop:decomposition}. We have:
    \begin{align*}
        \mathcal L_{\mathrm{mix}}(Q_{\mathrm{merge}})= \sum_{j=1}^T \beta_j\,\mathcal L_{\mathcal D_j}(Q_j) +\sum_{i=1}^T\sum_{j=1}^T \alpha_i\beta_j\,\big(\mathcal L_{\mathcal D_i}(Q_j)-\mathcal L_{\mathcal D_j}(Q_j)\big).
    \end{align*}

    We now bound $\sum_{j=1}^T \beta_j\,\mathcal L_{\mathcal D_j}(Q_j)$ term. For each $t\in[T]$, apply Theorem~\ref{thm:thm6-h} with $m=n_t$, $\lambda=\eta_t$, $Q=Q_t=\mathcal{N}\!\left(\theta_t,\Sigma_t\right)$, and the same prior $P$; since $\ell\in\left[0,1\right]$, $\mathrm{QSB}(\ell, 1)$ holds \emph{(i.e., $C=1$)}, and for Gaussian $Q_t$, $c_P\left(Q_t\right)=\left\|\Sigma_t\right\|$. Hence, with probability at least $1-\delta_t$,
    \begin{equation*}
    \mathcal{L}_{D_t}\!\left(Q_t\right)\;\le\;
    \frac{1}{1-\frac{\eta_t}{2}}\left(
    \widehat{\mathcal{L}}_{S_t}\!\left(Q_t\right)\;+\;\frac{\KL\!\left(Q_t\Vert P\right)+\log\!\left(\frac{1}{\delta_t}\right)}{\eta_t n_t}
    \right)
    \;+\;\frac{\eta_t}{2-\eta_t}\,\left\|\Sigma_t\right\|\;\mathcal{G}_{D_t}\!\left(Q_t\right),
    \end{equation*}

Choose $\delta_t>0$ such that $\sum_{t=1}^{T}\delta_t\le \delta$. By a union bound over $t\in[T]$ and substituting the above inequality into the decomposition, with probability at least $1-\delta$,
\begin{align*}
\mathcal{L}_{\alpha}\!\left(Q_{\mathrm{merge}}\right)
\le & \sum_{t=1}^{T}\beta_t\left[
\frac{1}{1-\frac{\eta_t}{2}}\left(
\widehat{\mathcal{L}}_{S_t}\!\left(Q_t\right)\;+\;\frac{\KL\!\left(Q_t\Vert P\right)+\log\!\left(\frac{1}{\delta_t}\right)}{\eta_t n_t}
\right)
\;+\;\frac{\eta_t}{2-\eta_t}\,\left\|\Sigma_t\right\|_{\mathrm{op}}\;\mathcal{G}_{D_t}\!\left(Q_t\right)
\right]\;\\&+\sum_{i=1}^T\sum_{j=1}^T \alpha_i\beta_j\,\big(\mathcal L_{\mathcal D_i}(Q_j)-\mathcal L_{\mathcal D_j}(Q_j)\big).
\end{align*}
\end{proof}

\subsection{Proof of Lemma~\ref{lem:L-by-mu}}
\begin{proof}
\label{proof:L-by-mu}
Given the convexity of $\ell$ and Jensen's inequality, we have $\mathcal L_{\mathcal D_t}(\mu_t) = \mathcal L_{\mathcal D_t}(\E_{h \sim Q_t}(h)) \le\E_{h \sim Q_t}(\mathcal L_{\mathcal D_t}(h)) = \mathcal L_{\mathcal D_t}(Q_t).$
Fix $x, y$. Let $\Delta_h = f_h(x) - f_{\mu}(x) = \nabla_\theta f_{\theta_0} (x)^\top (h - \mu)$. With $\ell$ being convex, $\gamma$-smooth in scores
$$
\ell(f_{\mu}(x)+\Delta_h,y)\le \ell(f_{\mu}(x),y)+\langle \nabla_s\ell(f_{\mu}(x),y),\Delta_h\rangle+\frac{\gamma}{2}\|\Delta_h\|_2^2
$$

Now if we take expectations over $(x, y) \sim \mathcal D$, the linear term $\langle \nabla_s\ell(f_{\mu}(x),y),\Delta_h\rangle$ will vanish. Now, plugging $\E_{x\sim \mathcal D}\E_{h\sim Q}\ \|\Delta_h\|_2^2= \E_{x\sim \mathcal D}\E_{h\sim Q}\|\Phi(x)^\top(h-\mu)\|_2^2=\mathrm{tr}\big(\Sigma \mathcal K_{\mathcal D}\big)$ will give the bound. The proof for empirical follows the same procedure, but with the empirical observation of data.
\end{proof}

\subsection{Proof of Lemma~\ref{lem:G_by_mu}}

\begin{proof}
\label{proof:G-by-mu}
Based on the chain rule, we have,
\begin{equation*}
    \mathcal G_{\mathcal D}(Q) =  \E_{(x,\, y) \sim \mathcal{D}}\big[\E_{h\sim Q}\,\|\nabla_h\ell\!\left(f_h(x), y\right)\|_2^2\big] = \E_{(x,\, y) \sim \mathcal{D}}\big[\E_{h\sim Q}\,\|\Phi(x)^\top\nabla_s\ell\!\left(f_h(x), y\right)\|_2^2\big].
\end{equation*}
We then add and subtract $\nabla_s\ell\!\left(f_\mu(x), y\right)$,
\begin{equation*}
    \mathcal G_{\mathcal D}(Q) = \E_{(x,\, y) \sim \mathcal{D}}\big[\E_{h\sim Q}\,\|\Phi(x)^\top\!\left(\nabla_s\ell\!\left(f_\mu(x), y\right) + \nabla_s\ell\!\left(f_h(x), y\right) - \nabla_s\ell\!\left(f_\mu(x), y\right)\right)\|_2^2\big].
\end{equation*}

By Minkowski's inequality,
\begin{align*}
    \sqrt{\mathcal{G}_{\mathcal{D}}(Q)}
    &= \Big(\E_{(x,\, y) \sim \mathcal{D}}\E_{h\sim Q}\,\|\Phi(x)^\top(\nabla_s\ell(f_\mu(x),y)+\nabla_s\ell(f_h(x),y)-\nabla_s\ell(f_\mu(x),y))\|_2^2\Big)^{\!\frac{1}{2}}\\
    &\le \Big(\E_{(x,\, y) \sim \mathcal{D}}\E_{h\sim Q}\,\|\Phi(x)^\top\nabla_s\ell(f_\mu(x),y)\|_2^2\Big)^{\!\frac{1}{2}}
    \\&\quad+ \Big(\E_{(x,\, y) \sim \mathcal{D}}\E_{h\sim Q}\,\|\Phi(x)^\top(\nabla_s\ell(f_h(x),y)-\nabla_s\ell(f_\mu(x),y))\|_2^2\Big)^{\!\frac{1}{2}}\\
    &= \sqrt{\mathcal{G}_{\mathcal{D}}(\mu)} \;+\; \Big(\E_{(x,\, y) \sim \mathcal{D}}\E_{h\sim Q}\,\|\Phi(x)^\top(\nabla_s\ell(f_h(x),y)-\nabla_s\ell(f_\mu(x),y))\|_2^2\Big)^{\!\frac{1}{2}}.
\end{align*}
Now, given the convexity and $\gamma$-smoothness of $\ell$ with respect to scores,
\begin{align*}
    \|\nabla_s\ell(f_h(x),y)-\nabla_s\ell(f_\mu(x),y)\|_2
    \;\le\; \gamma\,|f_h(x)-f_\mu(x)|
    \;=\; \gamma\,|\Phi(x)^\top(h-\mu)|.
\end{align*}
Hence,
\begin{align*}
    \left(\E_{(x,\, y)}\E_{h}\,\|\Phi(x)^\top(\nabla_s\ell(f_h(x),y)-\nabla_s\ell(f_\mu(x),y))\|_2^2\right)^{\!\frac{1}{2}}
    &\le \gamma \left(\E_{x}\,\|\Phi(x)\|_2^2\,\E_{h}\big[(\Phi(x)^\top(h-\mu))^2\big]\right)^{\!\frac{1}{2}}\\
    &= \gamma \left(\E_{x}\,\|\Phi(x)\|_2^2\,\Phi(x)^\top\Sigma\,\Phi(x)\right)^{\!\frac{1}{2}}
    \;\le\; \gamma\,\sqrt{\operatorname{tr}\!\left(\Sigma\,\mathcal K_{\mathcal D}^2\right)}.
\end{align*}
Therefore,
\begin{equation*}
    \sqrt{\mathcal G_{\mathcal D}(Q)} \;\le\; \sqrt{\mathcal G_{\mathcal D}(\mu)} \;+\; \gamma\,\sqrt{\operatorname{tr}\!\left(\Sigma\,\mathcal K_{\mathcal D}^2\right)},
\end{equation*}
and squaring both sides gives
\begin{equation*}
    \mathcal G_{\mathcal D}(Q)
    \;\le\;
    \left(\sqrt{\mathcal G_{\mathcal D}(\mu)}+\gamma\,\sqrt{\mathrm{tr}\!\left(\Sigma\,\mathcal K_{\mathcal D}^2\right)}\right)^2.
\end{equation*}
The empirical version follows the same procedure with $\mathcal D\to\mathcal S$ and $\mathcal K_{\mathcal D}\to\widehat{\mathcal K}_{\mathcal S}$.
\end{proof}

\subsection{Lemma~\ref{lem:posterior-det-hetero-term-bound}}
Define the \emph{deterministic} heterogeneity term
\[\mathcal H_{\mu}(\bm \alpha, \bm \beta)\;:=\;\sum_{i=1}^T\sum_{j=1}^T \alpha_i\beta_j\left(\mathcal L_{\mathcal D_i}\left(\mu_j\right)-\mathcal L_{\mathcal D_j}\left(\mu_j\right)\right).\]
\begin{lemma}
\label{lem:posterior-det-hetero-term-bound}
Under NTK, with loss $\ell$ being convex and $\gamma$-smooth in score, let $\mathcal K_\alpha=\sum_{t=1}^T\alpha_t\mathcal K_{\mathcal D_t}$,
    \begin{align*}
    \mathcal{H}_Q(\bm \alpha, \bm \beta) \le \mathcal{H}_\mu(\bm \alpha, \bm \beta) + \frac{\gamma}{2}\sum_{j=1}^T\beta_j\mathrm{tr}\left(\Sigma_j\mathcal{K}_{\alpha}\right).
    \end{align*}
\end{lemma}
\begin{proof}
\label{proof:posterior-det-hetero-term-bound}
Under NTK, with loss $\ell$ being convex and $\gamma$-smooth in score, given Lemma~\ref{lem:L-by-mu}, we have:
\begin{align*}
    \mathcal{H}_Q(\bm \alpha, \bm \beta) &= \sum_{i=1}^T\sum_{j=1}^T \alpha_i\beta_j\!\left(\mathcal{L}_{\mathcal{D}_i}(Q_j)-\mathcal{L}_{\mathcal{D}_j}(Q_j)\right)\\
    &\le \sum_{i=1}^T\sum_{j=1}^T \alpha_i\beta_j\!\left(\mathcal{L}_{\mathcal{D}_i}(\mu_j) + \frac{\gamma}{2}\mathrm{tr}\left(\Sigma_j \mathcal{K}_{\mathcal{D}_i}\right) -\mathcal{L}_{\mathcal{D}_j}(\mu_j)\right)\\
    &=\sum_{i=1}^T\sum_{j=1}^T \alpha_i\beta_j\left(\mathcal L_{\mathcal D_i}\left(\mu_j\right)-\mathcal L_{\mathcal D_j}\left(\mu_j\right)\right) + \frac{\gamma}{2}\sum_{j=1}^T\beta_j\mathrm{tr}\left(\Sigma_j\sum_{i=1}^T \alpha_i\mathcal{K}_{\mathcal{D}_i}\right)\\
    & = \mathcal{H}_\mu(\bm \alpha, \bm \beta) + \frac{\gamma}{2}\sum_{j=1}^T\beta_j\mathrm{tr}\left(\Sigma_j\mathcal{K}_{\alpha}\right)
\end{align*}
\end{proof}

\subsection{Proof of Lemma~\ref{lem:det-hetero-term-bound}}
\begin{proof}
    \label{proof:det-hetero-term-bound}
    We have:
    \begin{align*}
        \mathcal{H}_\mu(\bm \alpha, \bm \beta) &= \sum_{i=1}^T\sum_{j=1}^T \alpha_i\beta_j\left(\mathcal L_{\mathcal D_i}\left(\mu_j\right)-\mathcal L_{\mathcal D_j}\left(\mu_j\right)\right) \\
        &= \sum_{j=1}^T\beta_j\left(\sum_{i=1}^T \alpha_i\left(\mathcal L_{\mathcal D_i}\left(\mu_j\right)\right)-\mathcal L_{\mathcal D_j}\left(\mu_j\right)\right) \\
        &= \sum_{j=1}^T\beta_j\left(\mathcal{L}_\alpha\left(\mu_j\right)-\mathcal L_{\mathcal D_j}\left(\mu_j\right)\right)
    \end{align*}
    Fix $\theta_{\mathrm{merge}}$ and for a data distribution $\mathcal{D}$, define $g_\mathcal{D}(\theta_\mathrm{merge}) =\E_{(x, y) \sim \mathcal{D}} \nabla_\theta \ell(f_\theta(x), y)$, for every $\mu_j$, using descent lemma, we have:
    \begin{align*}
        \mathcal{L}_\mathcal{D}(\theta_{\mathrm{merge}}) + \langle g_\mathcal{D}(\theta_{\mathrm{merge}}), \Delta_j \rangle - \frac{\gamma}{2} \Delta_j^\top \mathcal{K}_\mathcal{D} \Delta_j\le\mathcal{L}_\mathcal{D}(\mu_j)\le \mathcal{L}_\mathcal{D}(\theta_{\mathrm{merge}}) + \langle g_\mathcal{D}(\theta_{\mathrm{merge}}), \Delta_j \rangle + \frac{\gamma}{2} \Delta_j^\top \mathcal{K}_\mathcal{D} \Delta_j
    \end{align*}
    Next, we apply the first inequality to $\mathcal{L}_{\mathcal{D}} = \mathcal{L}_{\mathcal{D}_j}$ and the second inequality to $\mathcal{L}_{\mathcal{D}} = \mathcal{L}_{\alpha}$. We then have:
    \begin{align*}
        \mathcal{L}_{\alpha}(\mu_j) &\leq \mathcal{L}_{\alpha}(\theta_{\mathrm{merge}}) + \langle g_\alpha(\theta_{\mathrm{merge}}), \Delta_j \rangle + \frac{\gamma}{2} \Delta_j^\top \mathcal{K}_{\alpha} \Delta_j, \\
    \mathcal{L}_{\mathcal{D}_j}(\mu_j) &\geq \mathcal{L}_{\mathcal{D}_j}(\theta_{\mathrm{merge}}) + \langle g_j(\theta_{\mathrm{merge}}), \Delta_j \rangle - \frac{\gamma}{2} \Delta_j^\top \mathcal{K}_{\mathcal{D}_j} \Delta_j.
    \end{align*}
    Subtract and we have:
    \begin{align*}
        \mathcal{H}_\mu(\bm \alpha, \bm \beta) &= \sum_{j=1}^T\beta_j\left(\mathcal{L}_\alpha\left(\mu_j\right)-\mathcal L_{\mathcal D_j}\left(\mu_j\right)\right)\\
        &\le \sum_{j=1}^T\beta_j\left(\mathcal{L}_{\alpha}(\theta_{\mathrm{merge}}) -\mathcal{L}_{\mathcal{D}_j}(\theta_{\mathrm{merge}}) \right) + \sum_{j=1}^T\beta_j \langle g_\alpha(\theta_{\mathrm{merge}}) - g_j(\theta_{\mathrm{merge}}), \Delta_j \rangle +\\
        &\qquad \frac{\gamma}{2} \sum_{j=1}^T\beta_j \Delta_j^\top (\mathcal{K}_{\alpha} + \mathcal{K}_{\mathcal{D}_j}) \Delta_j
    \end{align*}
    We then bound the $\sum_{j=1}^T\beta_j \langle g_\alpha(\theta_{\mathrm{merge}}) - g_j(\theta_{\mathrm{merge}}), \Delta_j \rangle$ using Cauchy-Schwartz:
    \begin{align*}
        \sum_{j=1}^T\beta_j \langle g_\alpha(\theta_{\mathrm{merge}}) - g_j(\theta_{\mathrm{merge}}), \Delta_j \rangle &\le \sqrt{\sum_{j=1}^T\beta_j \|g_\alpha\left(\theta_{\mathrm{merge}}) - g_j(\theta_{\mathrm{merge}}\right)\|_2^2} \sqrt{\sum_{j=1}^T\beta_j \|\Delta_j\|_2^2}
    \end{align*}
    We now bound the first term:

    \begin{align*}
        \sum_{j=1}^T \beta_j \|g_\alpha\left(\theta_{\mathrm{merge}}\right) - g_j\left(\theta_{\mathrm{merge}}\right)\|_2^2 &= \sum_{j=1}^T \beta_j \left( \| g_\alpha\left(\theta_{\mathrm{merge}}\right) \|_2^2 - 2\langle g_\alpha\left(\theta_{\mathrm{merge}}\right), g_j\left(\theta_{\mathrm{merge}}\right) \rangle + \| g_j\left(\theta_{\mathrm{merge}}\right) \|_2^2  \right) \\
        &= \| g_\alpha\left(\theta_{\mathrm{merge}}\right) \|_2^2 - 2 \langle g_\alpha\left(\theta_{\mathrm{merge}}\right), g_\beta\left(\theta_{\mathrm{merge}}\right) \rangle + \sum_{j=1}^T \beta_j \| g_j\left(\theta_{\mathrm{merge}}\right) \|_2^2  \\
        &\leq 2\| g_\alpha\left(\theta_{\mathrm{merge}}\right) \|_2^2 + 2\sum_{j=1}^T \beta_j \| g_j\left(\theta_{\mathrm{merge}}\right) \|_2^2 \\
        &\leq 2\sum_{i=1}^T \alpha_i \| g_i\left(\theta_{\mathrm{merge}}\right) \|_2^2 + 2\sum_{j=1}^T \beta_j \| g_j\left(\theta_{\mathrm{merge}}\right) \|_2^2.
    \end{align*}
    Now, given Jensen's inequality, we have:
    \begin{align*}
        2\sum_{i=1}^T \alpha_i \|g_i\left(\theta_{\mathrm{merge}}\right) \|_2^2 + 2\sum_{j=1}^T \beta_j \| g_j\left(\theta_{\mathrm{merge}}\right) \|_2^2 \le 2\left(\sum_{i=1}^T \alpha_i \mathcal{G}_{\mathcal{D}_i}(\theta_{\mathrm{merge}}) + \sum_{j=1}^T \beta_j \mathcal{G}_{\mathcal{D}_j}(\theta_{\mathrm{merge}})\right)
    \end{align*}
    So we have:
    \begin{align*}
        \mathcal{H}_\mu(\bm \alpha, \bm \beta) &= \sum_{j=1}^T\beta_j\left(\mathcal{L}_\alpha\left(\mu_j\right)-\mathcal L_{\mathcal D_j}\left(\mu_j\right)\right)\\
        &\le \sum_{j=1}^T\beta_j\left(\mathcal{L}_{\alpha}(\theta_{\mathrm{merge}}) -\mathcal{L}_{\mathcal{D}_j}(\theta_{\mathrm{merge}}) \right) + \sqrt{2\left(\sum_{i=1}^T \alpha_i \mathcal{G}_{\mathcal{D}_i}(\theta_{\mathrm{merge}}) + \sum_{j=1}^T \beta_j \mathcal{G}_{\mathcal{D}_j}(\theta_{\mathrm{merge}})\right)} \sqrt{\sum_{j=1}^T\beta_j \|\Delta_j\|_2^2}\\
        &\qquad \frac{\gamma}{2} \sum_{j=1}^T\beta_j \Delta_j^\top (\mathcal{K}_{\alpha} + \mathcal{K}_{\mathcal{D}_j}) \Delta_j.
    \end{align*}
    Now, apply ~\ref{lem:posterior-det-hetero-term-bound} to get the following,
    \begin{align*}
    \mathcal{H}_{Q}(\bm \alpha, \bm \beta)
    \;\le\;&
    \left(\mathcal L_{\bm\alpha}(\theta_{\mathrm{merge}})-\mathcal L_{\bm\beta}(\theta_{\mathrm{merge}})\right)
    +\sqrt{2\!\left(\sum_{t=1}^T \alpha_t \mathcal G_{\mathcal D_t}(\theta_{\mathrm{merge}})+\sum_{j=1}^T \beta_j \mathcal G_{\mathcal D_j}(\theta_{\mathrm{merge}})\right)}
    \;\sqrt{\sum_{j=1}^T \beta_j \|\Delta_j\|_2^2}\\
    &\qquad+\frac{\gamma}{2}\sum_{j=1}^T \beta_j\!\left[\Delta_j^\top(\mathcal K_\alpha+\mathcal K_\beta)\Delta_j + \mathrm{tr}\left(\Sigma_j\mathcal{K}_{\bm\alpha}\right)\right]. 
    \end{align*}
    which concludes the proof.
    
\end{proof}
\subsection{Proof of Theorem~\ref{thm:mtl-merged-pb}}
\begin{proof}
    \label{proof:mtl-merged-pb}
    From Theorem~\ref{thm:mtl-pb}, we have:
    \begin{align*}
    \mathcal L_{\bm\alpha}(Q_\mathrm{merge})\;\le\;
    \sum_{t=1}^T \beta_t\,&\Bigg[
    \frac{1}{1-\frac{\eta_t}{2}}
    \left(
    \hat{\mathcal L}_{S_t}(Q_t)
    +\frac{\KL(Q_t\Vert P)+\log(\frac{1}{\delta_t})}{\eta_t n_t}
    \!\right)+\frac{\eta_t}{2-\eta_t }\|\Sigma_t\|\,\mathcal G_{\mathcal D_t}(Q_t)\Bigg]
    \\&+\sum_{i=1}^T\sum_{j=1}^T \alpha_i\beta_j\,\big(\mathcal L_{\mathcal D_i}(Q_j)-\mathcal L_{\mathcal D_j}(Q_j)\big).
    \end{align*}
    Then we apply Lemma~\ref{lem:L-by-mu} to $\widehat{\mathcal{L}}_{\mathcal{S}_t}(Q_t)$ and Lemma~\ref{lem:G_by_mu} to $\mathcal{G}_{\mathcal{D}_t}(Q_t)$ for every $t \in $. Then we have:
    \begin{align*}
    \mathcal L_{\bm\alpha}\left(Q_\mathrm{merge}\right)\;\le\;
    &\sum_{t=1}^T \beta_t\,\Bigg[
    \frac{1}{1-\frac{\eta_t}{2}}
    \left(
    \hat{\mathcal L}_{S_t}\left(\mu_t\right) + \frac{\gamma}{2}\operatorname{tr}\left(\Sigma_t\mathcal{K}_{\mathcal{D}_t}\right)
    +\frac{\KL\left(Q_t\Vert P\right)+\log\left(\frac{1}{\delta_t}\right)}{\eta_t n_t}\right)
    \\&+\frac{\eta_t}{2-\eta_t }\|\Sigma_t\|\,\left(\sqrt{\mathcal G_{\mathcal D}(\mu_t)} + \gamma \sqrt{\operatorname{tr}\left(\Sigma_t\mathcal{K}_{\mathcal{D}_t}^2\right)}\right)^2\Bigg] + \mathcal{H}_{Q}(\bm\alpha, \bm\beta)
    \end{align*}
    Now, we apply Lemma~\ref{lem:det-hetero-term-bound}. Then we have:

    \begin{align*}
    \mathcal L_{\bm\alpha}\left(Q_{\mathrm{merge}}\right)\;\le\;
    &\sum_{t=1}^T \beta_t\,\Bigg[
    \frac{1}{1-\frac{\eta_t}{2}}
    \left(
    \hat{\mathcal L}_{S_t}\left(\mu_t\right) + \frac{\gamma}{2}\operatorname{tr}\left(\Sigma_t\mathcal{K}_{\mathcal{D}_t}\right)
    +\frac{\KL\left(Q_t\Vert P\right)+\log\left(\frac{1}{\delta_t}\right)}{\eta_t n_t}\right)
    \\&+\frac{\eta_t}{2-\eta_t }\|\Sigma_t\|\,\left(\sqrt{\mathcal G_{\mathcal D}(\mu_t)} + \gamma \sqrt{\operatorname{tr}\left(\Sigma_t\mathcal{K}_{\mathcal{D}_t}^2\right)}\right)^2\Bigg]+\left[\mathcal L_{\bm\alpha}(\theta_{\mathrm{merge}})-\mathcal L_{\bm\beta}(\theta_{\mathrm{merge}})\right]
    \\&
    +\sqrt{2\!\left(\sum_{t=1}^T \alpha_t \mathcal G_{\mathcal D_t}(\theta_{\mathrm{merge}})+\sum_{j=1}^T \beta_j \mathcal G_{\mathcal D_j}(\theta_{\mathrm{merge}})\right)}
    \;\sqrt{\sum_{j=1}^T \beta_j \|\Delta_j\|_2^2}\\
    &\qquad+\frac{\gamma}{2}\sum_{j=1}^T \beta_j\!\left[\Delta_j^\top(\mathcal K_\alpha+\mathcal K_\beta)\Delta_j + \mathrm{tr}\left(\Sigma_j\mathcal{K}_{\bm\alpha}\right)\right]
    \end{align*}
    In the end, we use Lemma~\ref{lem:L-by-mu}'s left inequality:
    \begin{align*}
    \mathcal L_{\bm\alpha}\left(\theta_{\mathrm{merge}}\right) &= \sum_{t=1}^T \alpha_t\mathcal{L}_{\mathcal{D}_t}(\theta_{\mathrm{merge}})\;\le\;\sum_{t=1}^T \alpha_t\mathcal{L}_{\mathcal{D}_t}(\theta_{\mathrm{merge}}) =\mathcal L_{\bm\alpha}\left(Q_{\mathrm{merge}}\right)\;\le\;
    \\&\sum_{t=1}^T \beta_t\,\Bigg[
    \frac{1}{1-\frac{\eta_t}{2}}
    \left(
    \hat{\mathcal L}_{S_t}\left(\mu_t\right) + \frac{\gamma}{2}\operatorname{tr}\left(\Sigma_t\mathcal{K}_{\mathcal{D}_t}\right)
    +\frac{\KL\left(Q_t\Vert P\right)+\log\left(\frac{1}{\delta_t}\right)}{\eta_t n_t}\right)
    \\&+\frac{\eta_t}{2-\eta_t }\|\Sigma_t\|\,\left(\sqrt{\mathcal G_{\mathcal D}(\mu_t)} + \gamma \sqrt{\operatorname{tr}\left(\Sigma_t\mathcal{K}_{\mathcal{D}_t}^2\right)}\right)^2\Bigg]+\left[\mathcal L_{\bm\alpha}(\theta_{\mathrm{merge}})-\mathcal L_{\bm\beta}(\theta_{\mathrm{merge}})\right]
    \\&
    +\sqrt{2\!\left(\sum_{t=1}^T \alpha_t \mathcal G_{\mathcal D_t}(\theta_{\mathrm{merge}})+\sum_{j=1}^T \beta_j \mathcal G_{\mathcal D_j}(\theta_{\mathrm{merge}})\right)}
    \;\sqrt{\sum_{j=1}^T \beta_j \|\Delta_j\|_2^2}\\
    &\qquad+\frac{\gamma}{2}\sum_{j=1}^T \beta_j\!\left[\Delta_j^\top(\mathcal K_\alpha+\mathcal K_\beta)\Delta_j + \mathrm{tr}\left(\Sigma_j\mathcal{K}_{\bm\alpha}\right)\right]
    \end{align*}

    which concludes the proof.
\end{proof}

\subsection{Lemma~\ref{lem:excess01_l1}}

\begin{lemma}
\label{lem:excess01_l1}
For a fixed input $x$, let $p(\cdot \mid x)$ and $q(\cdot \mid x)$ be two conditional probability distributions. Let $y^\star \in \arg\max_y p(y \mid x)$ and $\hat{y} \in \arg\max_y q(y \mid x)$ be the optimal predictions under these distributions. Then, $p(y^\star \mid x) - p(\hat{y} \mid x) \le \|p(\cdot \mid x) - q(\cdot \mid x)\|_1$.
\end{lemma}
\begin{proof}
\label{proof:excess01_l1}
We can write the difference as $p(y^\star) - p(\hat{y}) = (p(y^\star) - q(y^\star)) + (q(y^\star) - q(\hat{y})) + (q(\hat{y}) - p(\hat{y}))$. By definition, $q(\hat{y}) \ge q(y^\star)$, so the middle term is non-positive. The remaining two terms are bounded by their absolute values, the sum of which is at most $\sum_{y \in \mathcal{Y}} |p(y) - q(y)|$.
\end{proof}
\subsection{Lemma~\ref{lem:risk_shift_tv}}
\begin{lemma}
\label{lem:risk_shift_tv}
For any two conditional distributions $s_1$ and $s_2$ over data distribution $\mathcal{D}$ and any classifier $h$, we have $|\mathcal{L}^{0-1}_{s_1}(h) - \mathcal{L}^{0-1}_{s_2}(h)| \le \E_{x \sim \mathcal{D}}[\mathrm{TV}(s_1, s_2)]$ and $|\mathcal{L}^{0-1, \star}_{s_1} - \mathcal{L}^{0-1, \star}_{s_2}| \le \E_{x \sim \mathcal{D}}[\mathrm{TV}(s_1, s_2)]$.
\end{lemma}
\begin{proof}
\label{proof:risk_shift_tv}
For any fixed $x$ and label $y\in \mathcal{Y}$, we have $|s_1(y|x) - s_2(y|x)| \le \mathrm{TV}(s_1, s_2)$. Taking $y=h(x)$ and averaging over $x$ yields the first inequality. The second follows because of Lemma~\ref{lem:risk_shift_tv}.
\end{proof}
\subsection{Lemma~\ref{lem:pinsker}}
\begin{lemma}[Pinsker's Inequality, \citet{Csiszár_Körner_2015}]
\label{lem:pinsker}
For any two distributions $u,v$ on a finite set, their total variation is bounded by the KL divergence: $\mathrm{TV}(u,v) \le \sqrt{\frac{1}{2}\,\KL(u\Vert v)}$.
\end{lemma}

\subsection{Proof of Lemma~\ref{lem:single_task_bound}}
\begin{proof}
\label{proof:single_task_bound}
For a single task, we drop the subscript $t$ for clarity. The student's excess risk under the true distribution $y$ can be decomposed by adding and subtracting terms related to the teacher distribution $p$: (Note that $\mathcal{L}^{0-1, \star}_{y} = 0$ by its definition)
\begin{equation*}
    \mathcal{L}^{0-1}_{y}(h_\lambda)-\mathcal{L}^{0-1, \star}_y = \left(\mathcal{L}^{0-1}_{p}(h_\lambda)-\mathcal{L}^{0-1, \star}_p\right) + \left[ \left(\mathcal{L}^{0-1}_{y}(h_\lambda)-\mathcal{L}^{0-1}_{p}(h_\lambda)\right) + \left(\mathcal{L}^{0-1, \star}_p-\mathcal{L}^{0-1, \star}_y\right) \right].
\end{equation*}
The first component is the \emph{student-teacher fit}, representing the student's excess risk relative to the teacher. The second component is the \emph{teacher error}, which captures the error introduced by using $p$ as a proxy for $y$. We now bound each term.
\begin{align*}
    \mathcal{L}^{0-1}_{p}(h_\lambda)-\mathcal{L}^{0-1, \star}_p 
    &= \E_{x \sim \mathcal{D}}\left[ \max_y p(y|x) - p(h_\lambda(x)|x) \right] \\
    &\le \E_{x \sim \mathcal{D}}\left[ \|p(\cdot|x) - q_\lambda(\cdot|x)\|_1 \right] \quad \text{(by Lemma~\ref{lem:excess01_l1})} \\
    &\le \E_{x \sim \mathcal{D}}\left[ \sqrt{2 \KL(p(\cdot|x) \parallel q_\lambda(\cdot|x))} \right] \quad \text{(by definition of TV and Lemma~\ref{lem:pinsker})} \\
    &\le \sqrt{2 \E_{x \sim \mathcal{D}}\left[\KL(p(\cdot|x) \parallel q_\lambda(\cdot|x))\right]} \quad \text{(by Jensen's inequality)}
\end{align*}

\begin{align*}
    \left(\mathcal{L}^{0-1}_{y}(h_\lambda)-\mathcal{L}^{0-1}_{p}(h_\lambda)\right) &+ \left(\mathcal{L}^{0-1, \star}_p-\mathcal{L}^{0-1, \star}_r\right) 
    \le \left|\mathcal{L}^{0-1}_{y}(h_\lambda)-\mathcal{L}^{0-1}_{p}(h_\lambda)\right| + \left|\mathcal{L}^{0-1, \star}_p-\mathcal{L}^{0-1, \star}_r\right| \\
    &\le 2 \E_{x \sim \mathcal{D}}[\mathrm{TV}(y(\cdot|x),p(\cdot|x))] \quad \text{(by Lemma~\ref{lem:risk_shift_tv})} \\
    &\le \E_{x \sim \mathcal{D}}\left[\sqrt{2 \KL(y(\cdot|x) \parallel p(\cdot|x))}\right]\quad \text{(by Lemma~\ref{lem:pinsker})} \\
    &\le \sqrt{2 \E_{x \sim \mathcal{D}}\left[\KL(y(\cdot|x) \parallel p(\cdot|x))\right]}\quad \text{(by Jensen's inequality)}
\end{align*}
Combining the two bounds yields the desired result.
\end{proof}

\subsection{Proof of Theorem~\ref{thm:mtl_excess_bound}}
\begin{proof}
\label{proof:mtl_excess_bound}
We begin with the single-task excess risk bound from Proposition~\ref{lem:single_task_bound}, established for each task $t \in [T]$. We take the weighted average of this inequality across all tasks using the evaluation weights $\bm \alpha\in \Delta^{T-1}$: (Note that $\mathcal{L}^{0-1, \star}_{y_t} = 0$ by its definition)
\begin{align*}
\sum_{t=1}^T \alpha_t \mathcal{L}^{0-1}_{y_t}(h_\lambda)  
&\le \sum_{t=1}^T \alpha_t \left( \sqrt{2\,\E_{x \sim \mathcal{D}_t}\,\KL\big(p_t(\cdot|x) \parallel q_\lambda(\cdot|x)\big)} + \sqrt{2\,\E_{x \sim \mathcal{D}_t}\,\KL\big(y_t(\cdot|x) \parallel p_t(\cdot|x)\big)} \right) \\
&= \sum_{t=1}^T \alpha_t \sqrt{2\,\E_{x \sim \mathcal{D}_t}\,\KL\big(p_t(\cdot|x) \parallel q_\lambda(\cdot|x)\big)} + \sum_{t=1}^T \alpha_t \sqrt{2\,\E_{x \sim \mathcal{D}_t}\,\KL\big(y_t(\cdot|x) \parallel p_t(\cdot|x)\big)}.
\end{align*}

Given Jensen's inequality, we have:
\begin{align*}
\sum_{t=1}^T \alpha_t \sqrt{2\,\E_{x \sim \mathcal{D}_t}\,\KL\big(p_t(\cdot|x) \parallel q_\lambda(\cdot|x)\big)} 
&\le \sqrt{\sum_{t=1}^T \alpha_t \left( 2\,\E_{x \sim \mathcal{D}_t}\,\KL\big(p_t(\cdot|x) \parallel q_\lambda(\cdot|x)\big) \right)} \\
&= \sqrt{2 \sum_{t=1}^T \alpha_t \E_{x \sim \mathcal{D}_t} \KL\big(p_t(\cdot|x) \parallel q_\lambda(\cdot|x)\big)}
\end{align*}

and
\begin{align*}
\sum_{t=1}^T \alpha_t \sqrt{2\,\E_{x \sim \mathcal{D}_t}\,\KL\big(y_t(\cdot|x) \parallel p_t(\cdot|x)\big)}
&\le \sqrt{\sum_{t=1}^T \alpha_t \left( 2\,\E_{x \sim \mathcal{D}_t}\,\KL\big(y_t(\cdot|x) \parallel p_t(\cdot|x)\big) \right)} \\
&=\sqrt{2 \sum_{t=1}^T \alpha_t \E_{x \sim \mathcal{D}_t} \KL\big(y_t(\cdot|x) \parallel p_t(\cdot|x)\big)}.
\end{align*}
Combining these two inequalities gives the final multi-task bound as stated in the theorem.
\end{proof}

\section{Pseduo-code of \SAMerging}

\label{apx:samerging-alg}
Algorithm~\ref{alg:samerging} contains the pseudo-code of \SAMerging.

\begin{algorithm}
\small
\caption{\SAMerging: Sharpness-aware Model Merging}
\label{alg:samerging}
\begin{algorithmic}[1]
\Statex \textbf{Inputs:}
\State \quad $\theta_0$: Pretrained model parameters.
\State \quad $\{\theta_t\}_{t=1}^T$: Set of $T$ fine-tuned model parameters.
\State \quad $\{\mathcal{B}_t\}_{t=1}^T$: Set of $T$ unlabeled calibration datasets for each task.
\State \quad $\rho$: Neighborhood size for SAM.
\State \quad $\eta$: Learning rate for optimizer.
\State \quad $E$: Total number of epochs.
\State \quad $\{\alpha_t\}_{t=1}^T$: Task loss weights (typically $\alpha_t = \frac{1}{T})$.

\Statex
\Statex \textbf{Output:}
\State \quad $\lambda = \{\lambda_t^l\}$: Layer-wise merging coefficients.

\Statex
\Statex \textbf{Procedure:}
\Statex
\Function{ConstructMergedModel}{$\lambda, \theta_0, \{\tau_t\}_{t=1}^T$}
    \State Initialize $\theta_{\lambda} \leftarrow \theta_0$
    \For{each layer $l$}
        \State $\theta_{\lambda}^l \leftarrow \theta_0^l + \sum_{t=1}^{T} \lambda_t^l \tau_t^l$
    \EndFor
    \State \textbf{return} $\theta_{\lambda}$
\EndFunction

\Statex
\Function{KnowledgeDistillationLoss}{$\theta_{\mathrm{merge}}, \{\theta_t\}_{t=1}^T, \{\mathcal{B}_t\}_{t=1}^T, \{\alpha_t\}_{t=1}^T$}
    \State $\mathcal{L}_{\mathrm{total}} \leftarrow 0$
    \For{$t=1$ to $T$}
        \State $p_t(\cdot|x) \leftarrow \mathrm{Softmax}(f_{\theta_t}(x))$ \Comment{Teacher (fine-tuned model) distribution}
        \State $q_{\lambda}(\cdot|x) \leftarrow \mathrm{Softmax}(f_{\theta_{\mathrm{merge}}}(x))$ \Comment{Student (merged) distribution}
        \State $\mathcal{L}_{\mathrm{task}} \leftarrow \E_{x \in \mathcal{B}_t} \left[ \KL(p_t(\cdot|x) \, \Vert \, q_{\lambda}(\cdot|x)) \right]$
        \State $\mathcal{L}_{\mathrm{total}} \leftarrow \mathcal{L}_{\mathrm{total}} + \alpha_t \cdot \mathcal{L}_{\mathrm{task}}$
    \EndFor
    \State \textbf{return} $\mathcal{L}_{\mathrm{total}}$
\EndFunction

\Statex
\Statex \textbf{Initialization:}
\For{$t=1$ to $T$}
    \State $\tau_t \leftarrow \theta_t - \theta_0$ \Comment{Calculate task vectors}
\EndFor
\State $\lambda \leftarrow \mathbf{0}$ \Comment{Initialize merging coefficients}

\Statex
\Statex \textbf{Optimization Loop:}
\For{epoch = 1 to $E$}
    \State $\theta_{\lambda} \leftarrow \mathrm{ConstructMergedModel}(\lambda, \theta_0, \{\tau_t\})$
    \State
    \State \Comment{SAM Ascent Step: Find worst-case perturbation}
    \State $\mathcal{L}_{KD}(\lambda) \leftarrow \mathrm{KnowledgeDistillationLoss}(\theta_{\lambda}, \{\theta_t\}, \{\mathcal{B}_t\}, \{\alpha_t\})$
    \State $g(\lambda) \leftarrow \nabla_{\lambda} \mathcal{L}_{KD}(\lambda)$
    \State $\epsilon \leftarrow \rho \frac{g(\lambda)}{\|g(\lambda)\|_2}$ \Comment{Normalize gradient to find ascent direction}
    \State
    \State \Comment{SAM Descent Step: Update on perturbed parameters}
    \State $\theta_{\lambda+\epsilon} \leftarrow \mathrm{ConstructMergedModel}(\lambda + \epsilon, \theta_0, \{\tau_t\})$
    \State $\mathcal{L}_{KD}(\lambda+\epsilon) \leftarrow \mathrm{KnowledgeDistillationLoss}(\theta_{\lambda+\epsilon}, \{\theta_t\}, \{\mathcal{B}_t\}, \{\alpha_t\})$
    \State $g_{\mathrm{SAM}}(\lambda) \leftarrow \nabla_{\lambda} \mathcal{L}_{KD}(\lambda+\epsilon)$
    \State
    \State $\lambda \leftarrow \lambda - \eta \cdot g_{\mathrm{SAM}}(\lambda)$ \Comment{Update coefficients (e.g., via SGD or Adam)}
\EndFor

\Statex
\State \textbf{return} $\lambda^*$ \Comment{Return the final optimized coefficients}
\end{algorithmic}
\end{algorithm}

\section{Experiments Results and Ablation}
\label{apx:exp}
\subsection{Full data and baseline setup}
\label{apx:exp-baseline}
\subsubsection{Tasks and data}
We evaluate generalization across increasing interference regimes on four suites following \citet{Ilharco_Ribeiro_Wortsman_Gururangan_Schmidt_Hajishirzi_Farhadi_2023, Wang_Dimitriadis_Ortiz-Jimenez_Fleuret_Frossard_2024}: (i) \textbf{TA-8} (8 image classification tasks: Cars \citep{Krause_Stark_Deng_Fei-Fei_2013}, DTD \citep{Cimpoi_Maji_Kokkinos_Mohamed_Vedaldi_2014}, EuroSAT \citep{Helber_Bischke_Dengel_Borth_2019}, GTSRB \citep{Stallkamp_Schlipsing_Salmen_Igel_2011}, MNIST \citep{LeCun_1998}, RESISC45 \citep{Cheng_Han_Lu_2017}, SUN397 \citep{Xiao_Ehinger_Hays_Torralba_Oliva_2016}, SVHN \citep{Netzer_Wang_Coates_Bissacco_Wu_Ng_others_2011}), (ii) \textbf{TALL-14} \citep{Wang_Dimitriadis_Ortiz-Jimenez_Fleuret_Frossard_2024} (TA-8 + six: Oxford-102 Flowers \citep{Nilsback_Zisserman_2008}, CIFAR-100 \citep{Krizhevsky_Hinton_others_2009}, PCAM \citep{Veeling_Linmans_Winkens_Cohen_Welling_2018}, STL-10 \citep{Coates_Ng_Lee_2011}, Oxford-IIIT Pet \citep{Parkhi_Vedaldi_Zisserman_Jawahar_2012}, FER2013 \citep{Goodfellow_Erhan_Carrier_Courville_Mirza_Hamner_Cukierski_Tang_Thaler_Lee}), (iii) \textbf{TALL-20} \citep{Wang_Dimitriadis_Ortiz-Jimenez_Fleuret_Frossard_2024} (TALL-14 + six: EMNIST \citep{Cohen_Afshar_Tapson_VanSchaik_2017}, CIFAR-10 \citep{Krizhevsky_Hinton_others_2009}, Food-101 \citep{Bossard_Guillaumin_VanGool_2014}, Fashion-MNIST \citep{Xiao_Ehinger_Hays_Torralba_Oliva_2016}, KMNIST \citep{Clanuwat_Bober-Irizar_Kitamoto_Lamb_Yamamoto_Ha_2018}, RenderedSST2 \citep{Socher_Perelygin_Wu_Chuang_Manning_Ng_Potts_2013}), and (iv) \textbf{GLUE} \citep{Wang_Singh_Michael_Hill_Levy_Bowman_2019} (7 NLP tasks: CoLA \citep{Warstadt_Singh_Bowman_2019}, SST-2 \citep{Socher_Perelygin_Wu_Chuang_Manning_Ng_Potts_2013}, MRPC \citep{Dolan_Brockett_2005}, QQP, MNLI \citep{Williams_Nangia_Bowman_2017}, RTE, QNLI \citep{Rajpurkar_Zhang_Lopyrev_Liang_2016}). Vision backbones are CLIP ViT-B/32 and ViT-L/14; for GLUE, we use GPT-2 fine-tuned per task to obtain task vectors, mirroring the setup in \citet{Wang_Dimitriadis_Ortiz-Jimenez_Fleuret_Frossard_2024}.

\subsubsection{Baselines}
We use the following baselines for comparison:
\begin{itemize}
    \item Simple Averaging \citep{Wortsman_Ilharco_Gadre_others_2022}: this method averages the fine-tuned models' parameters to achieve the merged model.
    \item Task Arithmetic \citep{Ilharco_Ribeiro_Wortsman_Gururangan_Schmidt_Hajishirzi_Farhadi_2023}: This method treats the difference of each fine-tuned model as a ``task vector'', then scales and adds these task vectors to make the merged model.
    \item TIES-Merging \citep{Yadav_Tam_Choshen_Raffel_Bansal_2023}: This method prunes small-magnitude updates, resolves sign conflicts across fine-tuned models, and merges only weights that agree in sign to reduce interference.
    \item Isotropic Merging \citep{Marczak_Magistri_Cygert_Twardowski_Bagdanov_Weijer_2025}: This method derives a shared subspace from the combined updates (SVD), makes it isotropic, then adds and orthogonalizes each fine-tuned model's residual directions before the same isotropic scaling—reducing interference while preserving specialization.
    \item Fisher Merging \citep{Matena_Raffel_2022}: parameter-wise weighted averaging where each weight is scaled by its Fisher information.
    \item PCB-Merging:
    \item RegMean \citep{Jin_Ren_Preotiuc-Pietro_Cheng_2025}: closed-form layerwise regression on unlabeled activations to match fine-tuned model/ensemble logits; solve linear layers by least squares, average the remaining parameters.
    \item RegMean++ \citep{Nguyen_Huu-Tien_Suzuki_Nguyen_2025}: RegMean's closed-form layerwise regression, but compute Gram stats from activations of the partially merged model (not each fine-tuned model), capturing cross-layer dependencies.
    \item AdaMerging \citep{Yang_Wang_Shen_Liu_Guo_Wang_Tao_2023}: adaptively learns task-/layer-wise merge coefficients on unlabeled data by minimizing prediction entropy.
\end{itemize}

{ 
\subsection{Experiments Setup}
\label{apx:exp-setup}

Here we explain the setup in detail for each baseline and also \SAMerging. Note that all experiments are conducted through the Fusion Bench benchmarking \cite{tang2024fusionbench}:

\begin{itemize}
    \item \SAMerging: For training \SAMerging, we set the learning rate to $0.001$; we use the SAM optimizer with Adam as its base optimizer with momentum $0.99$ and weight decay $5\times 10^{-4}$; the perturbation radius is $\rho=0.07$; batch size is $16$; and weights are tied.
    \item MTL: We fine-tune CLIP with learning rate $1\times 10^{-5}$, weight decay $0$, seed $42$; batch size $128$ for TALL-14 over $4{,}000$ steps and batch size $64$ for TALL-20 over $8{,}000$ steps (LoRA disabled).
    \item Simple Averaging: This method does not have any hyperparameters. It is an unweighted average of task models.
    \item Task Arithmetic:
    \begin{itemize}
        \item TA-8: we set the scaling factor to $0.2$ according to Table~\ref{tab:lambda_sensitivity}.
        \item TALL-14:  we set the scaling factor to $0.1$ according to Table~\ref{tab:tall_sensitivity}.
        \item TALL-20:  we set the scaling factor to $0.05$ according to Table~\ref{tab:tall_sensitivity}.
    \end{itemize}
    \item TIES-Merging: \begin{itemize}
        \item TA-8: we set the scaling factor to $0.4$ according to Table~\ref{tab:lambda_sensitivity} and top-k threshold to $20$.
        \item TALL-14:  we set the scaling factor to $0.15$ according to Table~\ref{tab:tall_sensitivity} and top-k threshold to $20$..
        \item TALL-20:  we set the scaling factor to $0.15$ according to Table~\ref{tab:tall_sensitivity} and top-k threshold to $20$.
    \end{itemize}
    \item Isotropic Merging: We use ISO-CTS variant with a scaling factor $1.0$; we set the common-space fraction is $0.8$.
    \item PBC-Merging: We set the parameter competition balancing ratio to $0.05$.
    \begin{itemize}
        \item TA-8: We set the scaling factor to $1.2$ (default value).
        \item TA-8: We set the scaling factor to $0.6$.
        \item TA-8: We set the scaling factor to $0.5$.
    \end{itemize}
    \item Fisher (k=1600): We compute Fisher weights with $k=1600$ examples, normalize Fisher weights, set minimal Fisher weight to $1\times 10^{-6}$, and use dataloader batch size $16$ with $4$ workers.
    \item RegMean: We set $k=1600$ examples and set the $\mathrm{reduce\_off\_diagonal}$ to $0.6$ (default value).
    \item RegMean++: We set $k=1600$ examples and set the $\mathrm{reduce\_off\_diagonal}$ to $0.95$ (default value)
    \item AdaMerging: We train AdaMerging with learning rate $0.001$ using Adam and batch size $16$; weights are tied and initialized to $0.2$–$0.3$ depending on the setting.
\end{itemize}

\begin{table*}[t]

\centering

\begin{threeparttable}

\scriptsize

{

\begin{tabular}{l c c c c c c}
\toprule

Method & 0.05 & 0.07 & 0.1 & 0.15 & 0.2 & 0.3 \\

\midrule

Task Arithmetic (TALL-14) & 64.0 & 65.3 & \textbf{66.5} & 66.2 & 63.3 & 52.8 \\
TIES-Merging (TALL-14) & 62.4 & 63.9 & 65.7 & 67.8 & \textbf{68.7} & 67.6 \\

\midrule

Task Arithmetic (TALL-20) & 61.1 & \textbf{61.3} & 60.6 & 56.1 & 49.6 & 36.3 \\
TIES-Merging (TALL-20) & 60.6 & 61.6 & 62.7 & \textbf{63.0} & 61.8 & 55.5 \\

\bottomrule

\end{tabular}

\caption{{ Average accuracy on ViT-B/32 for TALL-14 (14 tasks) and TALL-20 (20 tasks) model pools across different scaling factors $\lambda$.}}

\label{tab:tall_sensitivity}

}

\end{threeparttable}

\end{table*}

\begin{table*}[t]
\centering
\begin{threeparttable}
{
\scriptsize
\begin{tabular}{l c c c c c c c c c c}
\toprule
Method & 0.1 & 0.2 & 0.3 & 0.4 & 0.5 & 0.6 & 0.7 & 0.8 & 0.9 & 1.0 \\
\midrule
Task Arithmetic & 64.1 & \textbf{69.5} & 67.5 & 60.7 & 51.3 & 42.4 & 35.1 & 29.3 & 24.4 & 19.3 \\
TIES-Merging & 60.6 & 67.6 & 71.9 & \textbf{73.1} & 71.7 & 68.4 & 64.0 & 59.0 & 53.5 & 48.6 \\
\bottomrule
\end{tabular}
\caption{{ Average accuracy across TA-8 tasks on ViT-B/32 for different $\lambda$ values. Bold values indicate the best $\lambda$ for each method.}}
\label{tab:lambda_sensitivity}
}
\end{threeparttable}
\end{table*}

\begin{table*}[t]
\centering
{
\begin{threeparttable}
\scriptsize
\begin{tabular}{l c c c c c}
\toprule
Method & 0.02 & 0.05 & 0.1 & 0.2 & 1.0 \\
\midrule
Accuracy (\%) & 81.77 & 81.82 & 81.82 & 81.82 & 81.80 \\
CE Loss & 0.649 & 0.648 & 0.648 & 0.647 & 0.648 \\
\bottomrule
\end{tabular}
\caption{ Average accuracy and cross-entropy (CE) loss on ViT-B/32 for TA8 model pool across different $\rho$ values. }
\label{tab:rho_sensitivity}
\end{threeparttable}
}
\end{table*}

}
\subsection{Full experiments results}
\label{apx:exp-res}

Below are the full results of merging methods' performance on each task for different suites and backbones.

We report per-task accuracies (Acc., \%) for TA-8, TALL-14, and TALL-20 using CLIP ViT-B/32 and CLIP ViT-L/14. ``Avg.'' denotes the mean over tasks. 
For data-dependent methods, $k$ indicates the number of unlabeled samples per task used for adaptation.
See Table~\ref{tab:ta8-complete}, Table~\ref{tab:tall14-complete}, and Table~\ref{tab:tall20-complete}.
We also report the results for the GLUE benchmark, as shown in Table~\ref{tab:glue-complete}
\begin{table*}[!h]
\centering
\begin{threeparttable}
\setlength{\tabcolsep}{2.0pt}
\scriptsize
\resizebox{\textwidth}{!}{
\begin{tabular}{@{\extracolsep{\fill}} l l c c c c c c c c c}
\toprule
Backbone & Method & SUN397 & Cars & RESISC45 & EuroSAT & SVHN & GTSRB & MNIST & DTD & Avg. \\
\midrule
 ViT-B/32 & \multicolumn{10}{l}{\it Base} \\
  & MTL & 72.2 & 76.5 & 92.0 & 97.2 & 95.5 & 97.7 & 99.3 & 77.5 & 88.5 \\
  & \multicolumn{10}{l}{\it Data-free} \\
  & Simple Averaging & 65.4 & 62.4 & 70.6 & 75.7 & 64.5 & 55.0 & 86.3 & 50.6 & 66.3 \\
  & Task Arithmetic & 57.0 & 55.7 & 64.7 & 73.3 & 77.9 & 68.5 & 96.1 & 47.1 & 67.5 \\
  & TIES-Merging & 67.0 & 64.2 & 74.3 & 74.5 & 77.7 & 69.4 & 94.1 & 54.0 & 71.9 \\
  & Isotropic Merging & 71.6 & 73.6 & 84.1 & 87.1 & 73.0 & 80.9 & 95.3 & 65.0 & 78.8 \\
  & \multicolumn{10}{l}{\it Data-dependent} \\
  & Fisher (k=1600) & 67.5 & 68.0 & 70.2 & 75.4 & 81.9 & 54.9 & 90.3 & 56.0 & 70.5 \\
  & RegMean (k=1600) & 67.9 & 68.6 & 82.5 & 94.4 & 90.0 & 78.8 & 97.7 & 64.0 & 80.5 \\
  & RegMean++ & 69.2 & 69.7 & 87.1 & 95.8 & 94.4 & 89.8 & 99.0 & 68.6 & 84.2 \\
  & AdaMerging LW (k=1600) & 61.5 & 61.2 & 71.3 & 86.9 & 83.9 & 76.7 & 97.4 & 50.5 & 73.7 \\
  & AdaMerging LW (k=16000) & 68.0 & 71.3 & 83.7 & 92.0 & 87.5 & 93.3 & 98.2 & 67.2 & 82.6 \\
  & \multicolumn{10}{l}{\it Ours} \\
  & SAMerging (k=1600) & 71.1 & 75.0 & 91.3 & 96.6 & 92.4 & 96.8 & 98.1 & 75.6 & 87.1 \\
\addlinespace
 ViT-L/14 & \multicolumn{10}{l}{\it Base} \\
  & MTL & 79.0 & 89.3 & 94.4 & 98.3 & 96.4 & 98.1 & 99.4 & 83.7 & 92.3 \\
  & \multicolumn{10}{l}{\it Data-free} \\
  & Simple Averaging & 72.5 & 81.5 & 82.3 & 88.5 & 81.6 & 74.0 & 96.6 & 61.8 & 79.9 \\
  & Task Arithmetic & 73.3 & 81.4 & 84.1 & 89.6 & 86.6 & 81.7 & 97.6 & 62.3 & 82.1 \\
  & TIES-Merging & 74.8 & 83.2 & 86.5 & 89.7 & 89.7 & 85.2 & 97.8 & 63.9 & 83.8 \\
  & Isotropic Merging & 79.5 & 91.0 & 93.9 & 96.3 & 91.4 & 94.5 & 98.6 & 77.1 & 90.3 \\
  & \multicolumn{10}{l}{\it Data-dependent} \\
  & Fisher (k=1600) & 70.0 & 79.2 & 70.3 & 99.0 & 65.0 & 58.8 & 85.5 & 58.4 & 73.3 \\
  & RegMean (k=1600) & 75.4 & 88.2 & 91.0 & 96.7 & 95.8 & 92.6 & 98.5 & 73.6 & 89.0 \\
  & RegMean++ & 77.5 & 89.6 & 68.2 & 97.3 & 97.0 & 96.3 & 99.1 & 81.4 & 88.3 \\
  & AdaMerging LW (k=1600) & 74.5 & 83.5 & 86.6 & 92.4 & 90.9 & 90.7 & 98.2 & 63.7 & 85.1 \\
  & AdaMerging LW (k=16000) & 78.1 & 90.7 & 90.7 & 96.1 & 95.0 & 97.6 & 98.6 & 81.3 & 91.0 \\
  & \multicolumn{10}{l}{\it Ours} \\
  & SAMerging (k=1600) & 80.5 & 92.1 & 95.3 & 97.4 & 95.5 & 98.1 & 99.1 & 82.7 & 92.6 \\
\addlinespace
\bottomrule
\end{tabular}
}
\caption{TA-8 per-task accuracies (Acc., \%). Columns list the 8 TA-8 tasks and Avg. Acc. is the mean over them.}
\label{tab:ta8-complete}
\end{threeparttable}
\end{table*}
\begin{table*}[!h]
\centering
\begin{threeparttable}
\setlength{\tabcolsep}{1.8pt}
\scriptsize
\resizebox{\textwidth}{!}{
\begin{tabular}{@{\extracolsep{\fill}} l l c c c c c c c c c c c c c c c}
\toprule
Backbone & Method & SUN397 & Cars & RESISC45 & EuroSAT & SVHN & GTSRB & MNIST & DTD & Flowers & PCAM & FER2013 & Pet & STL10 & CIFAR100 & Avg. \\
\midrule
 ViT-B/32 & \multicolumn{16}{l}{\it Base} \\
  & MTL & 73.0 & 72.9 & 93.2 & 98.5 & 96.4 & 97.7 & 99.6 & 76.7 & 87.4 & 86.1 & 71.4 & 90.9 & 97.8 & 86.4 & 87.7 \\
  & \multicolumn{16}{l}{\it Data-free} \\
  & Simple Averaging & 64.8 & 60.4 & 67.1 & 67.0 & 50.7 & 45.6 & 76.6 & 46.9 & 67.4 & 65.2 & 51.6 & 84.2 & 97.2 & 70.4 & 65.4 \\
  & Task Arithmetic & 64.4 & 59.6 & 67.3 & 67.8 & 54.0 & 50.0 & 80.7 & 48.0 & 66.1 & 69.8 & 53.1 & 84.2 & 96.6 & 69.2 & 66.5 \\
  & TIES-Merging & 62.2 & 54.6 & 65.3 & 63.0 & 65.7 & 63.9 & 92.6 & 49.9 & 58.2 & 77.1 & 54.9 & 81.4 & 94.8 & 62.4 & 67.6 \\
  & Isotropic Merging & 70.6 & 68.8 & 81.6 & 85.1 & 73.5 & 81.2 & 96.4 & 61.9 & 75.3 & 80.7 & 66.5 & 88.9 & 97.5 & 75.3 & 78.8 \\
  & \multicolumn{16}{l}{\it Data-dependent} \\
  & Fisher (k=1600) & 65.7 & 64.3 & 66.9 & 65.1 & 61.4 & 46.3 & 79.1 & 49.9 & 70.1 & 63.1 & 52.0 & 87.3 & 97.2 & 71.0 & 67.1 \\
  & RegMean (k=1600) & 66.3 & 64.6 & 76.6 & 90.3 & 78.3 & 65.7 & 94.8 & 56.9 & 71.8 & 81.8 & 61.7 & 87.9 & 96.9 & 71.9 & 76.1 \\
  & RegMean++ & 67.3 & 66.6 & 81.9 & 94.4 & 91.3 & 80.3 & 98.1 & 61.9 & 74.9 & 76.0 & 64.2 & 90.1 & 97.5 & 73.4 & 79.8 \\
  & AdaMerging LW (k=1600) & 62.7 & 58.5 & 69.0 & 82.2 & 73.6 & 62.5 & 95.5 & 50.3 & 62.8 & 75.3 & 58.3 & 83.6 & 95.2 & 65.5 & 71.1 \\
  & AdaMerging LW (k=16000) & 66.5 & 69.4 & 82.4 & 92.6 & 85.6 & 89.9 & 97.8 & 61.0 & 73.8 & 51.4 & 64.7 & 87.5 & 96.7 & 68.6 & 77.7 \\
  & \multicolumn{16}{l}{\it Ours} \\
  & SAMerging (k=1600) & 68.9 & 72.2 & 90.6 & 93.8 & 89.1 & 94.0 & 98.5 & 72.4 & 83.4 & 77.6 & 67.6 & 90.2 & 97.2 & 76.6 & 83.7 \\
\addlinespace
 ViT-L/14 & \multicolumn{16}{l}{\it Base} \\
  & MTL & 79.2 & 88.9 & 94.8 & 98.0 & 96.1 & 97.5 & 99.3 & 83.8 & 97.5 & 90.8 & 72.9 & 96.1 & 99.4 & 88.4 & 91.6 \\
  & \multicolumn{16}{l}{\it Data-free} \\
  & Simple Averaging & 71.2 & 79.0 & 78.7 & 80.4 & 71.3 & 64.6 & 94.3 & 58.7 & 81.9 & 74.2 & 54.8 & 94.6 & 99.3 & 82.4 & 77.5 \\
  & Task Arithmetic & 71.6 & 78.4 & 79.3 & 80.3 & 72.4 & 67.9 & 95.3 & 59.8 & 81.9 & 71.1 & 56.1 & 94.8 & 99.0 & 82.3 & 77.9 \\
  & TIES-Merging & 72.0 & 75.6 & 76.5 & 69.7 & 77.2 & 75.1 & 96.7 & 57.8 & 79.6 & 78.2 & 59.9 & 94.7 & 98.4 & 77.7 & 77.8 \\
  & Isotropic Merging & 79.1 & 90.5 & 94.2 & 95.8 & 91.1 & 94.6 & 98.8 & 76.2 & 96.9 & 84.4 & 71.8 & 96.6 & 99.6 & 88.2 & 89.8 \\
  & \multicolumn{16}{l}{\it Data-dependent} \\
  & Fisher (k=1600) & 70.0 & 77.4 & 75.9 & 97.1 & 60.4 & 57.7 & 86.6 & 58.1 & 84.2 & 59.8 & 52.8 & 94.7 & 99.4 & 81.0 & 75.4 \\
  & RegMean (k=1600) & 72.9 & 84.7 & 87.0 & 95.2 & 92.9 & 86.3 & 98.1 & 66.9 & 92.0 & 86.6 & 66.1 & 96.1 & 99.3 & 83.5 & 86.3 \\
  & RegMean++ & 74.1 & 86.7 & 89.7 & 96.7 & 95.7 & 91.7 & 98.9 & 71.1 & 94.2 & 80.2 & 70.0 & 96.1 & 99.3 & 85.7 & 87.9 \\
  & AdaMerging LW (k=1600) & 73.7 & 80.5 & 84.6 & 88.7 & 84.6 & 83.3 & 97.5 & 62.4 & 83.9 & 69.2 & 61.7 & 95.4 & 98.9 & 81.4 & 81.9 \\
  & AdaMerging LW (k=16000) & 77.5 & 90.0 & 91.2 & 96.1 & 94.3 & 96.2 & 98.5 & 77.0 & 95.3 & 51.3 & 74.0 & 95.9 & 99.4 & 83.3 & 87.2 \\
  & \multicolumn{16}{l}{\it Ours} \\
  & SAMerging (k=1600) & 78.5 & 89.5 & 94.4 & 97.6 & 94.5 & 97.2 & 98.9 & 81.6 & 97.2 & 86.3 & 72.2 & 95.6 & 99.1 & 86.5 & 90.7 \\
\addlinespace
\bottomrule
\end{tabular}
}
\caption{TALL-14 per-task accuracies (Acc., \%). Columns list the 14 TALL-14 tasks and Avg. Acc. is the mean over them.}
\label{tab:tall14-complete}
\end{threeparttable}
\end{table*}
\begin{table*}[!h]
\centering
\begin{threeparttable}
\setlength{\tabcolsep}{1.2pt}
\tiny
\resizebox{\textwidth}{!}{
\begin{tabular}{@{\extracolsep{\fill}} l l c c c c c c c c c c c c c c c c c c c c c}
\toprule
Backbone & Method & SUN397 & Cars & RESISC45 & EuroSAT & SVHN & GTSRB & MNIST & DTD & Flowers & PCAM & FER2013 & Pet & STL10 & CIFAR100 & CIFAR10 & Food101 & Fashion & EMNIST & KMNIST & SST2 & Avg. \\
\midrule
 ViT-B/32 & \multicolumn{22}{l}{\it Base} \\
  & MTL & 73.5 & 74.9 & 93.8 & 99.0 & 96.6 & 97.6 & 99.5 & 78.1 & 87.9 & 87.4 & 71.6 & 90.9 & 98.0 & 87.2 & 97.4 & 86.4 & 94.3 & 95.4 & 97.6 & 71.7 & 88.9 \\
  & \multicolumn{22}{l}{\it Data-free} \\
  & Simple Averaging & 64.2 & 59.6 & 64.8 & 60.9 & 47.3 & 43.1 & 71.8 & 46.4 & 66.5 & 63.9 & 50.2 & 84.1 & 97.0 & 69.8 & 92.7 & 79.7 & 71.3 & 15.0 & 11.4 & 61.8 & 61.1 \\
  & Task Arithmetic & 64.2 & 59.6 & 64.8 & 60.9 & 47.3 & 43.1 & 71.8 & 46.4 & 66.5 & 63.9 & 50.2 & 84.1 & 97.0 & 69.8 & 92.7 & 79.7 & 71.3 & 15.0 & 11.4 & 61.8 & 61.1 \\
  & TIES-Merging & 65.0 & 59.7 & 66.3 & 60.7 & 52.4 & 49.1 & 79.3 & 48.4 & 66.5 & 66.7 & 51.7 & 84.1 & 97.0 & 70.4 & 93.3 & 79.5 & 72.4 & 17.3 & 12.2 & 61.9 & 62.7 \\
  & Isotropic Merging & 68.0 & 59.2 & 76.9 & 81.8 & 73.9 & 80.8 & 96.5 & 58.5 & 72.7 & 83.5 & 64.6 & 86.3 & 96.9 & 73.9 & 95.0 & 75.9 & 83.2 & 35.3 & 37.2 & 69.8 & 73.5 \\
  & \multicolumn{22}{l}{\it Data-dependent} \\
  & Fisher (k=1600) & 65.0 & 62.8 & 64.4 & 59.1 & 53.8 & 43.3 & 71.5 & 48.3 & 68.6 & 62.1 & 50.1 & 86.2 & 97.1 & 70.4 & 93.7 & 80.4 & 70.8 & 16.8 & 13.1 & 67.4 & 62.2 \\
  & RegMean (k=1600) & 65.5 & 62.4 & 74.5 & 85.4 & 70.4 & 60.0 & 89.3 & 54.3 & 70.1 & 81.4 & 60.1 & 86.2 & 96.6 & 70.9 & 94.1 & 80.9 & 77.9 & 20.3 & 31.8 & 67.5 & 70.0 \\
  & RegMean++ & 66.2 & 64.5 & 79.3 & 92.6 & 87.3 & 73.3 & 93.8 & 57.7 & 72.4 & 73.6 & 63.4 & 88.4 & 97.1 & 72.5 & 94.9 & 83.1 & 82.7 & 28.7 & 40.4 & 67.1 & 74.0 \\
  & AdaMerging LW (k=1600) & 58.9 & 47.6 & 61.7 & 72.4 & 62.6 & 54.0 & 94.3 & 44.8 & 55.9 & 70.5 & 54.4 & 79.4 & 93.7 & 62.5 & 90.6 & 64.4 & 75.5 & 18.8 & 13.5 & 54.5 & 61.5 \\
  & AdaMerging LW (k=16000) & 66.6 & 67.4 & 81.9 & 91.7 & 80.7 & 87.9 & 92.7 & 60.3 & 73.3 & 52.2 & 64.9 & 85.8 & 96.9 & 69.6 & 91.2 & 78.3 & 70.8 & 15.8 & 10.0 & 50.0 & 69.4 \\
  & \multicolumn{22}{l}{\it Ours} \\
  & SAMerging (k=1600) & 66.1 & 67.8 & 86.6 & 94.7 & 81.8 & 90.1 & 95.2 & 68.6 & 78.9 & 72.2 & 63.7 & 89.2 & 96.0 & 71.2 & 92.0 & 78.8 & 87.8 & 81.2 & 88.8 & 71.1 & 81.1 \\
\addlinespace
 ViT-L/14 & \multicolumn{22}{l}{\it Base} \\
  & MTL & 79.2 & 89.2 & 95.1 & 98.2 & 96.1 & 97.7 & 99.3 & 82.9 & 98.0 & 90.7 & 72.7 & 95.7 & 99.5 & 87.9 & 98.5 & 92.3 & 92.1 & 93.0 & 91.2 & 85.7 & 91.8 \\
  & \multicolumn{22}{l}{\it Data-free} \\
  & Simple Averaging & 70.7 & 77.7 & 76.4 & 75.3 & 69.5 & 62.1 & 93.7 & 57.7 & 80.8 & 73.6 & 52.7 & 94.2 & 99.2 & 81.7 & 97.0 & 90.5 & 77.4 & 16.1 & 10.4 & 66.1 & 71.1 \\
  & Task Arithmetic & 70.7 & 77.7 & 76.4 & 75.3 & 69.5 & 62.1 & 93.7 & 57.7 & 80.8 & 73.6 & 52.7 & 94.2 & 99.2 & 81.7 & 97.0 & 90.5 & 77.4 & 16.1 & 10.4 & 66.1 & 71.1 \\
  & TIES-Merging & 71.7 & 77.9 & 78.1 & 75.8 & 73.8 & 66.6 & 95.4 & 59.1 & 81.4 & 72.4 & 55.2 & 94.7 & 99.1 & 82.1 & 97.4 & 90.5 & 80.8 & 18.4 & 10.8 & 65.2 & 72.3 \\
  & Isotropic Merging & 78.7 & 87.6 & 93.7 & 94.6 & 90.4 & 94.2 & 98.7 & 75.1 & 97.0 & 85.4 & 70.9 & 96.5 & 99.5 & 87.7 & 98.5 & 92.4 & 90.7 & 46.2 & 40.8 & 78.5 & 84.8 \\
  & \multicolumn{22}{l}{\it Data-dependent} \\
  & Fisher (k=1600) & 70.0 & 77.2 & 76.1 & 96.9 & 61.3 & 57.7 & 87.5 & 58.1 & 83.8 & 60.1 & 52.1 & 94.6 & 99.4 & 81.0 & 96.9 & 89.9 & 74.7 & 15.4 & 10.1 & 64.7 & 70.4 \\
  & RegMean (k=1600) & 71.7 & 82.6 & 84.6 & 94.1 & 89.0 & 80.0 & 97.1 & 64.1 & 90.0 & 84.6 & 63.1 & 95.6 & 99.3 & 82.5 & 97.6 & 91.5 & 86.9 & 32.9 & 19.9 & 68.3 & 78.8 \\
  & RegMean++ & 73.1 & 85.3 & 88.2 & 96.3 & 94.5 & 88.2 & 98.0 & 68.2 & 92.1 & 80.8 & 68.4 & 96.2 & 99.2 & 84.3 & 98.2 & 91.8 & 88.9 & 45.4 & 42.8 & 70.3 & 82.5 \\
  & AdaMerging LW (k=1600) & 71.8 & 73.3 & 76.1 & 71.1 & 76.0 & 72.6 & 97.0 & 57.4 & 78.9 & 68.8 & 59.3 & 94.5 & 98.1 & 76.7 & 96.2 & 83.4 & 81.5 & 18.3 & 11.7 & 66.4 & 71.5 \\
  & AdaMerging LW (k=16000) & 76.7 & 89.5 & 89.3 & 96.1 & 91.9 & 95.4 & 98.2 & 74.3 & 94.2 & 51.6 & 69.8 & 95.6 & 99.2 & 82.7 & 96.4 & 90.1 & 86.0 & 12.8 & 10.0 & 79.7 & 79.0 \\
  & \multicolumn{22}{l}{\it Ours} \\
  & SAMerging (k=1600) & 77.5 & 88.6 & 93.9 & 96.9 & 92.8 & 96.4 & 97.5 & 79.9 & 96.7 & 84.8 & 71.8 & 95.7 & 99.0 & 84.5 & 97.0 & 91.1 & 90.2 & 87.0 & 92.9 & 83.6 & 89.9 \\
\addlinespace
\bottomrule
\end{tabular}
}
\caption{TALL-20 per-task accuracies (Acc., \%). Columns list the 20 TALL-20 tasks and Avg. Acc. is the mean over them.}
\label{tab:tall20-complete}
\end{threeparttable}
\end{table*}
\begin{table*}[!h]
\centering
\begin{threeparttable}
\setlength{\tabcolsep}{2.8pt}
\scriptsize
\begin{tabular*}{\textwidth}{@{\extracolsep{\fill}} l l *{8}{c}}
\toprule
Backbone & Method & CoLA & MNLI & MRPC & QNLI & QQP & RTE & SST-2 & Avg. \\
\midrule
GPT-2 & \multicolumn{9}{l}{\it Reference Results} \\
 & Fine-tuned (STL) & 76.8 & 82.1 & 80.4 & 88.3 & 89.6 & 65.3 & 91.2 & 82.0 \\
 & \multicolumn{9}{l}{\it Model Merging} \\
 & Simple Average & 55.0 & 55.1 & 51.0 & 57.6 & 76.7 & 44.8 & 52.5 & 56.1 \\
 & Task Arithmetic ($\lambda{=}0.5$) & 68.7 & 68.6 & 69.6 & 70.5 & 81.8 & 47.3 & 83.6 & 70.0 \\
 & TIES-Merging ($\lambda{=}0.6$) & 68.4 & 71.4 & 68.4 & 69.6 & 82.4 & 47.7 & 81.8 & 70.0 \\
 & Fisher Merging & 54.8 & 58.0 & 39.5 & 63.3 & 81.5 & 49.1 & 64.7 & 58.7 \\
 & RegMean & 61.7 & 70.4 & 65.4 & 69.7 & 78.8 & 56.0 & 79.7 & 68.8 \\
 & AdaMerging & 67.8 & 59.2 & 70.6 & 63.4 & 80.6 & 47.3 & 74.0 & 68.8 \\
 & \multicolumn{9}{l}{\bf \it Ours} \\
 & \bf \SAMerging &  68.4 & 75.1 &  73.5 &  80.7 &  78.3 &  57.8 &  86.9 &  74.9 \\
\bottomrule
\end{tabular*}
\caption{Multi-task model merging methods using GPT-2 models on the GLUE benchmark.}
\label{tab:glue-complete}
\end{threeparttable}
\end{table*}

\clearpage
\section{Visualization}
\label{apx:vis}
Here we plot the loss landscape around the merged model on different pairs of tasks and benchmarks.

\begin{figure}[h!]
    \centering
    \includegraphics[width=0.65\linewidth]{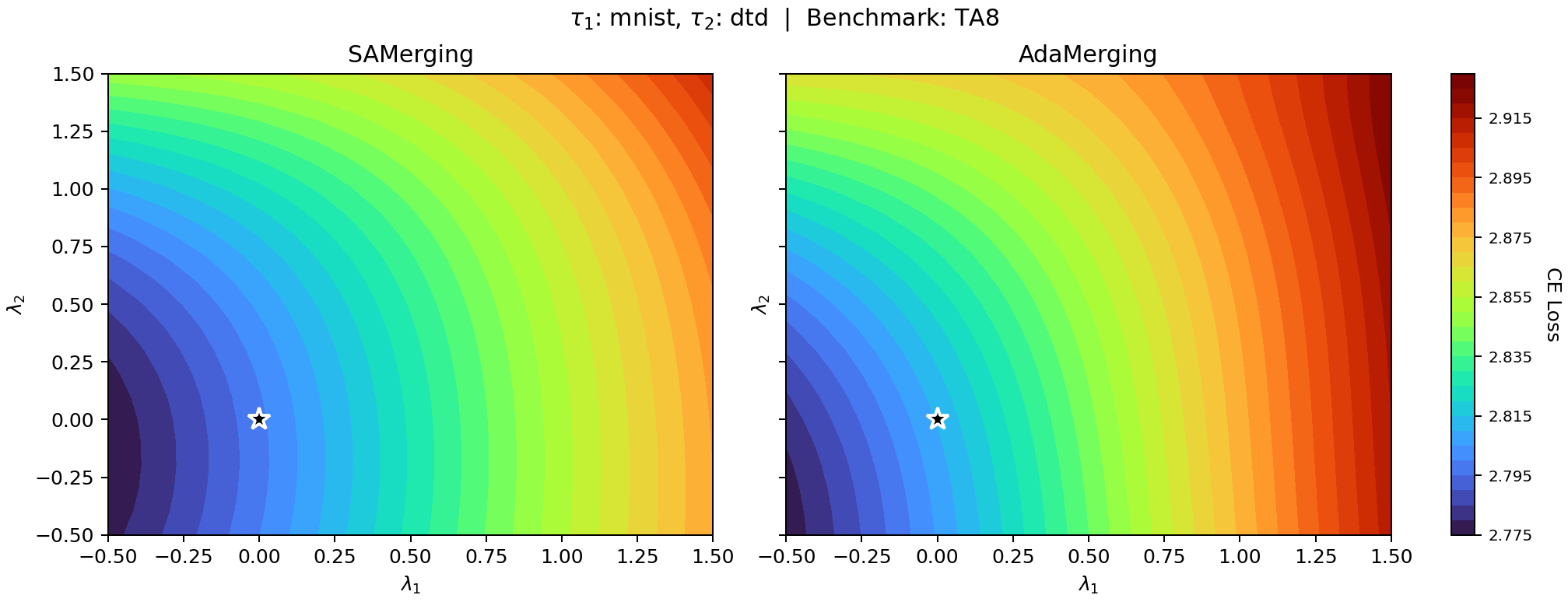}
    \caption{Loss surface for MNIST and DTD on TA-8.}
    \label{fig:mnist_dtd_8}
\end{figure}

\begin{figure}[h!]
    \centering
    \includegraphics[width=0.65\linewidth]{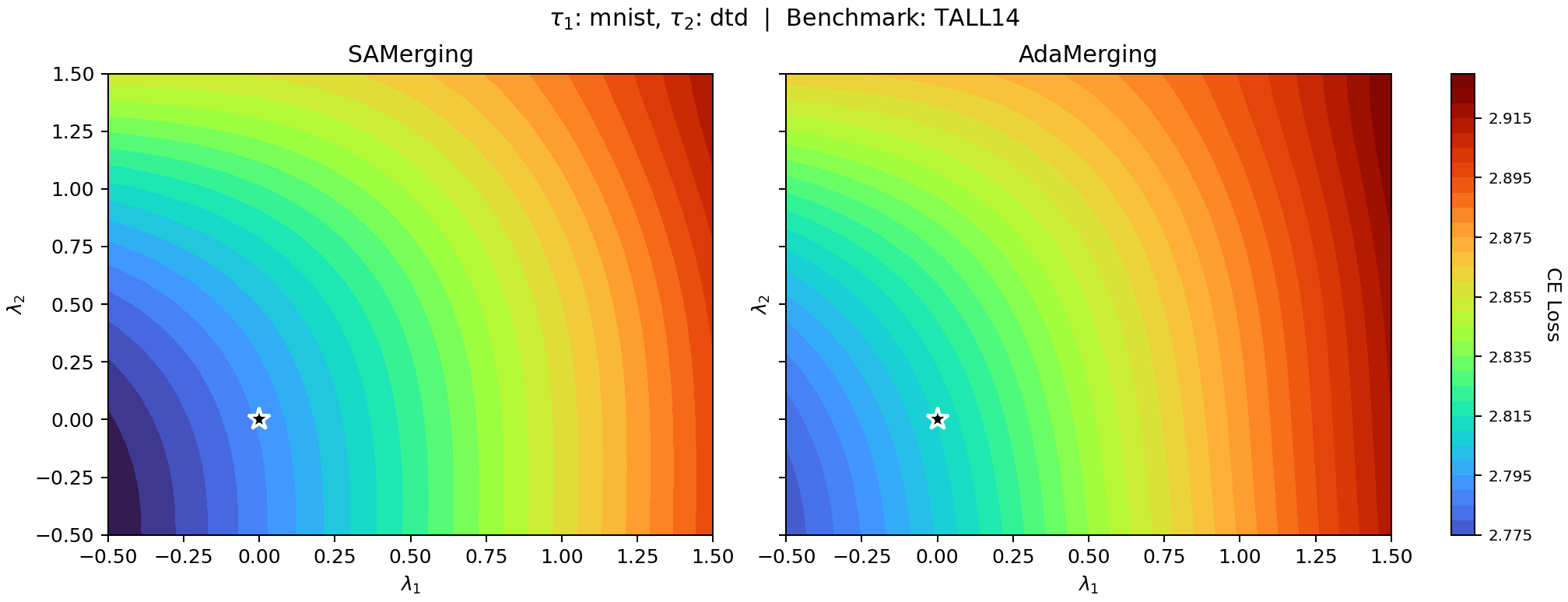}
    \caption{Loss surface for MNIST and DTD on TALL-14.}
    \label{fig:mnist_dtd_14}
\end{figure}

\begin{figure}[h!]
    \centering
    \includegraphics[width=0.65\linewidth]{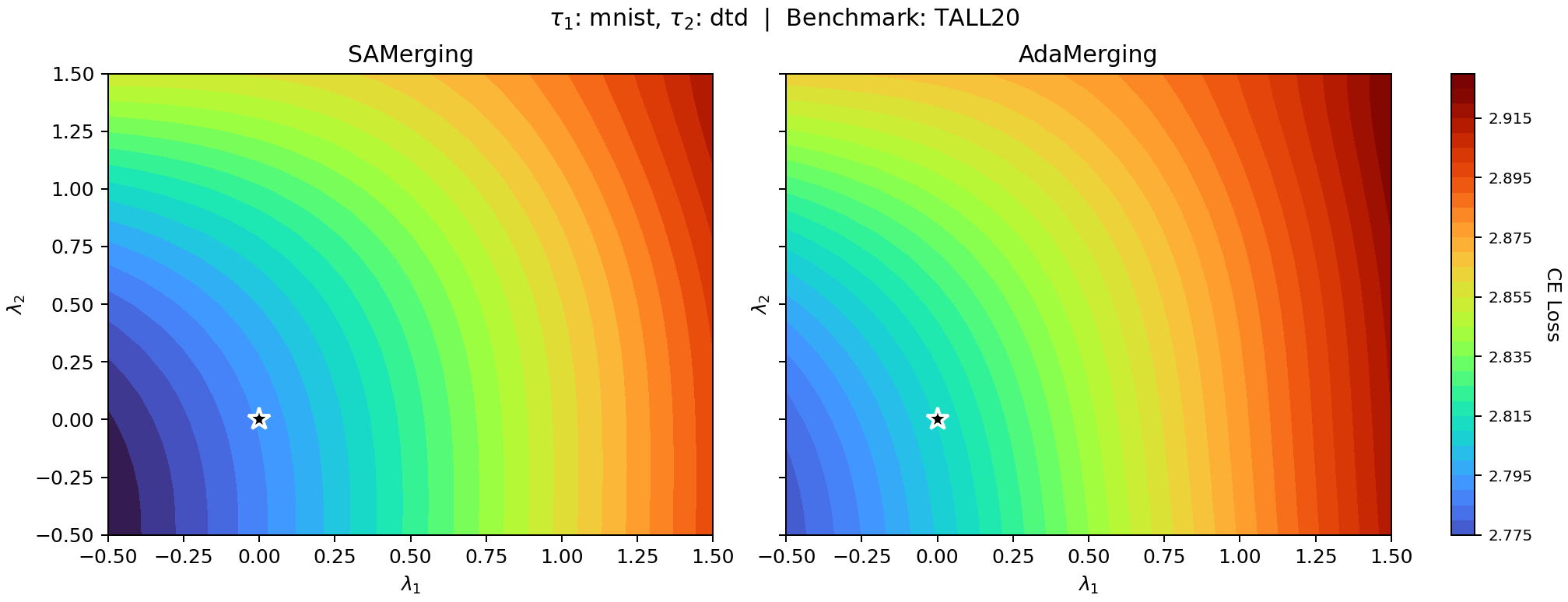}
    \caption{Loss surface for MNIST and DTD on TALL-20.}
    \label{fig:mnist_dtd_20}
\end{figure}

\begin{figure}[h!]
    \centering
    \includegraphics[width=0.65\linewidth]{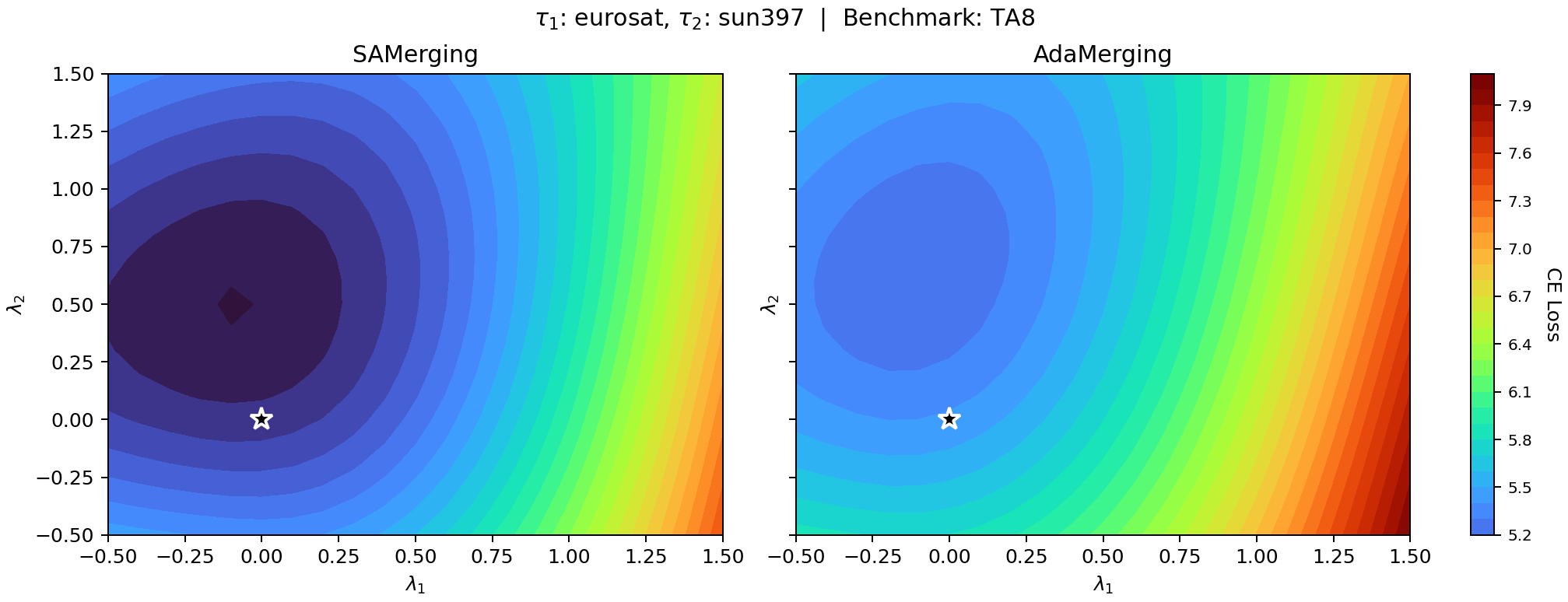}
    \caption{EuroSAT vs SUN397 (TA-8) loss landscape.}
    \label{fig:eurosat_sun_heatmap_appendix}
\end{figure}
\end{document}